\documentclass{article}
\usepackage{jfrExamplee}
\usepackage{amsmath,amsfonts}
\usepackage{algorithmic}
\usepackage{algorithm}
\usepackage{array}
\usepackage[caption=false,font=normalsize]{subfig}
\usepackage{textcomp}
\usepackage{stfloats}
\usepackage{url}
\usepackage{verbatim}
\usepackage{enumerate}
\usepackage[hidelinks]{hyperref}
\usepackage{graphicx}
\usepackage{cite}
\usepackage{multirow}
\usepackage[acronym]{glossaries}
\usepackage{comment}
\usepackage{cleveref}
\crefname{figure}{fig.}{figs.}
\Crefname{figure}{Fig.}{Figs.}
\usepackage{caption}
\usepackage{setspace}
\usepackage{placeins}
\usepackage{tabularx}
\usepackage{booktabs}
\usepackage{siunitx}
\usepackage[table]{xcolor}
\graphicspath{{figures/}}
\usepackage{hhline}
\usepackage[margin=1in, footskip=0.5in]{geometry}
\usepackage{bm}
\usepackage{tablefootnote}
\usepackage{tikz}
\usetikzlibrary{arrows.meta, positioning, calc}
\usepackage{mathtools}
\usepackage{soul}

\usepackage{listings}
\definecolor{codegreen}{rgb}{0,0.6,0}
\definecolor{codegray}{rgb}{0.5,0.5,0.5}
\definecolor{codepurple}{rgb}{0.58,0,0.82}
\definecolor{backcolour}{rgb}{0.95,0.95,0.92}

\lstdefinestyle{mystyle}{
    backgroundcolor=\color{backcolour},   
    commentstyle=\color{codegreen},
    keywordstyle=\color{magenta},
    numberstyle=\tiny\color{codegray},
    stringstyle=\color{codepurple},
    basicstyle=\ttfamily\footnotesize,
    breakatwhitespace=false,         
    breaklines=true,                 
    captionpos=b,                    
    keepspaces=true,                 
    numbers=left,                    
    numbersep=5pt,                  
    showspaces=false,                
    showstringspaces=false,
    showtabs=false,                  
    tabsize=2
}

\newcommand{\rev}[1]{\textcolor{black}{#1}}
\newcommand{\revdel}[1]{}
\usepackage{eso-pic}
\newcommand\AtPageUpperMyleft[1]{\AtPageUpperLeft{%
\put(\LenToUnit{1cm},\LenToUnit{-2cm}){#1}%
}}%
\AddToShipoutPictureBG*{%
  \AtPageUpperMyleft{\parbox[b][2cm][c]{\dimexpr\paperwidth-2cm\relax}{%
    \centering
    \fontsize{12}{12}\selectfont
    \color{gray!50}
    This paper has been accepted for publication in the Journal of Field Robotics (JFR) under DOI: 10.1002/rob.22429.
  }}%
}

\AddToShipoutPictureBG*{
  \AtPageLowerLeft{%
    \raisebox{25pt}{\makebox[\paperwidth]{\begin{minipage}{21cm}\centering
    \fontsize{10}{12}\selectfont
      \textcolor{gray!50}{ \copyright 2024 Wiley. This article may be used for non-commercial purposes in accordance with Wiley Terms and Conditions for Self-Archiving.
      }
    \end{minipage}}}%
  }
}

\begin{document}
\pagestyle{plain}
\newacronym{mpc}{MPC}{Model Predictive Control}
\newacronym{mpcc}{MPCC}{Model Predictive Contouring Controller}
\newacronym{pid}{PID}{Proportional-Integral-Derivative}
\newacronym{lidar}{LiDAR}{Light Detection and Ranging}
\newacronym{rl}{RL}{Reinforcement Learning}
\newacronym{mlp}{MLP}{Multilayer Perceptron}
\newacronym{ml}{ML}{Machine Learning}
\newacronym{sb3}{SB3}{Stable Baselines 3}
\newacronym{sac}{SAC}{Soft Actor Critic}
\newacronym{ppo}{PPO}{Proximal Policy Optimization}
\newacronym{ai}{AI}{Artificial Intelligence}
\newacronym{nn}{NN}{Neural Network}
\newacronym{sota}{SotA}{State-of-the-Art}
\newacronym{esc}{ESC}{Electronic Speed Controller}
\newacronym{ros}{ROS}{Robot Operating System}
\newacronym{imu}{IMU}{Inertial Measurement Unit}
\newacronym{slam}{SLAM}{Simultaneous Localization And Mapping}
\newacronym{sdc}{SDC}{Self Driving Cars}
\newacronym{obc}{OBC}{On Board Computer}
\newacronym{qp}{QP}{Quadratic Programming}
\newacronym{uav}{UAV}{Unmanned Aerial Vehicles}
\newacronym{cg}{CG}{Center of Gravity}
\newacronym{em}{EM}{Expectation Maximization}
\newacronym{rms}{RMS}{Root Mean Square}
\newacronym{map}{MAP}{Model- and Acceleration-based Pursuit}
\newacronym{pd}{PD}{Proportional-Derivative}
\newacronym{lut}{LUT}{Lookup Table}
\newacronym{RGB}{RGB}{Red Green Blue}
\newacronym{HYP}{HYP}{Hyper Spectral}
\newacronym{shm}{SHM}{Structural Health Monitoring}
\newacronym{iot}{IoT}{Internet of Things}
\newacronym{mcc}{MCC}{Matthews Correlation Coefficient}
\newacronym{pca}{PCA}{Principal Component Analysis}
\newacronym{sne}{SNE}{Stochastic Neighbor Embedding}
\newacronym{umap}{UMAP}{Uniform Manifold Approximation and Projection}
\newacronym{led}{LED}{Light Emitting Diode}
\newacronym{ftg}{FTG}{Follow The Gap}
\newacronym{gps}{GPS}{Global Positioning System}
\newacronym{vesc}{VESC}{Vedder Electronic Speed Controller}
\newacronym{poe}{PoE}{Power Over Ethernet}
\newacronym{tcs}{TCS}{Traction-Control System}
\newacronym{mcl}{MCL}{Monte Carlo Localization}
\newacronym{pf}{PF}{Particle Filter}
\newacronym{amcl}{AMCL}{Adaptive Monte-Carlo Localization}
\newacronym{pgo}{PGO}{Pose Graph Optimization}
\newacronym{icp}{ICP}{Iterative Closest Point}
\newacronym{ndt}{NDT}{Normal Distributions Transform}
\newacronym{gpu}{GPU}{Graphics Processing Unit}
\newacronym{cpu}{CPU}{Central Processing Unit}
\newacronym{fsd}{FSD}{Formula Student Driverless}
\newacronym{iac}{IAC}{Indy Autonomous Challenge}
\newacronym{cots}{CotS}{Commercial off-the-Shelf}
\newacronym{lipo}{LiPo}{Lithium Polymer}
\newacronym{nuc}{NUC}{Next Unit of Computing}
\newacronym{tof}{ToF}{Time-of-Flight}
\newacronym{dof}{DoF}{Degrees of freedom}
\newacronym{kf}{KF}{Kalman Filter}
\newacronym{ekf}{EKF}{Extended Kalman Filter}
\newacronym{ukf}{UKF}{Unscented Kalman Filter}
\newacronym{yolo}{YOLO}{You Only Look Once}
\newacronym{amz}{AMZ}{Academic Motorsports Club Zurich}
\newacronym{cv}{CV}{Computer Vision}
\newacronym{rmse}{RMSE}{Root Mean Squared Error}
\newacronym{los}{LoS}{Line of Sight}
\newacronym{tpr}{TPR}{True Positive Rate}
\newacronym{fdr}{FDR}{False Discovery Rate}
\newacronym{tp}{TP}{True Positive}
\newacronym{tn}{TN}{True Negative}
\newacronym{fp}{FP}{False Positive}
\newacronym{fn}{FN}{False Negative}
\newacronym{fpr}{FPR}{False Positive Rate}
\newacronym[firstplural=Gaussian Processes (GPs)]{gp}{GP}{Gaussian Process}
\newacronym{lpv}{LPV}{Linear Parameter Varying}
\newacronym{erpm}{ERPM}{Electric Revolutions Per Minute}
\newacronym{rc}{RC}{Radio-Controlled}
\newacronym{mcu}{MCU}{Microcontroller Unit}
\newacronym{bo}{BO}{Bayesian Optimization}
\newacronym{ei}{EI}{Expected Improvement}
\newacronym{ucb}{UCB}{Upper Confidence Bound}
\newacronym{pp}{PP}{Pure Pursuit}
\newacronym{ram}{RAM}{Random Access Memory}
\newacronym{rep}{REP}{ROS Enhancement Proposal}
\newacronym{ids3c}{IDS3C}{Information and Decision Science Lab Scaled Smart City}
\newacronym{idsc}{IDSC}{Institute for Dynamic Systems and Control}
\newacronym{mave}{mAVE}{mean Average Velocity Error}
\newacronym{acc}{ACC}{Adaptive Cruise Control}
\newacronym{ids}{IDS}{Identity Switches}

\title{ForzaETH Race Stack - Scaled Autonomous Head-to-Head Racing on Fully Commercial off-the-Shelf Hardware}

\author{
Nicolas Baumann\thanks{Corresponding Author:\\ \text{\quad} Nicolas Baumann, ETH Zürich, ETF F110, Sternwartstrasse 7, 8092 Zürich, Email: \url{nicolas.baumann@pbl.ee.ethz.ch}} \thanks{Contributed Equally} \thanks{Affiliated with the Center for Project-Based Learning (PBL), ETH Zürich, Zürich, Switzerland} , Edoardo Ghignone$^{\dag \ddag}$, Jonas Kühne$^{\ddag}$, Niklas Bastuck$^{\ddag}$, Jonathan Becker$^{\ddag}$, \\
\textbf{Nadine Imholz$^{\ddag}$, Tobias Kränzlin$^{\ddag}$, Tian Yi Lim$^{\ddag}$, Michael Lötscher$^{\ddag}$, Luca Schwarzenbach$^{\ddag}$,}\\
\textbf{Luca Tognoni$^{\ddag}$, Christian Vogt$^{\ddag}$, Andrea Carron\thanks{Affiliated with the Institute for Dynamic Systems and Control (IDSC), ETH Zürich, Zürich, Switzerland} , and Michele Magno$^{\ddag}$} \\
ETH Zürich\\
Zurich, Switzerland \\
}

\maketitle

\vspace{-0.5cm}

\begin{abstract}
\vspace{-0.5cm}

Autonomous racing in robotics combines high-speed dynamics with the necessity for reliability and real-time decision-making. While such racing pushes software and hardware to their limits, many existing full-system solutions necessitate complex, custom hardware and software, and usually focus on Time-Trials rather than full unrestricted Head-to-Head racing, due to financial and safety constraints. This limits their reproducibility, making advancements and replication feasible mostly for well-resourced laboratories with comprehensive expertise in mechanical, electrical, and robotics fields. Researchers interested in the autonomy domain but with only partial experience in one of these fields, need to spend significant time with familiarization and integration. 
The ForzaETH Race Stack addresses this gap by providing an autonomous racing software platform designed for F1TENTH, a 1:10 scaled Head-to-Head autonomous racing competition, which simplifies replication by using commercial off-the-shelf hardware. This approach enhances the competitive aspect of autonomous racing and provides an accessible platform for research and development in the field. 
The ForzaETH Race Stack is designed with modularity and operational ease of use in mind, allowing customization and adaptability to various environmental conditions, such as track friction and layout, \rev{which is exemplified by the various modularly implemented state estimation and control systems.}
Capable of handling both Time-Trials and Head-to-Head racing, the stack has demonstrated its effectiveness, robustness, and adaptability in the field by winning the official F1TENTH international competition multiple times.
\rev{Furthermore, the stack demonstrated its reliability and performance at unprecedented scales, up to over \SI{10}{\metre \per \second} on tracks up to \SI{150}{\metre} in length.}

\end{abstract}

\textbf{Keywords---} Autonomous driving, Autonomous racing, Motion control, Robotic perception, Path planning, State estimation, Open source software

\section*{Supplementary Material}
Open-source code of the proposed system: \url{https://github.com/ForzaETH/race_stack}\\
%Supplementary video material: \texttt{work in progress} 

\newpage

\FloatBarrier
\section{Introduction}
%Time and time again, from the invention of disc brakes to \gls{tcs}, 
Motorsport has historically demonstrated its capability to be a catalyst for introducing cutting-edge technologies to the broader automotive industry \cite{catalyst0, catalyst1}. Autonomous racing provides a valuable context to investigate some of the critical scenarios of general self-driving which necessitate operation at the friction limit, such as in high-speed or low-friction environments like icy or dusty roads. Autonomous racing inherently demands operation at these boundaries, compelling the vehicle to its physical, computational, and algorithmic limits, thus serving as a stress testing platform for self-driving \cite{amz_fullsystem, tum_fullsystem, betz_ar_survey, betz_weneed_ar}.

Drawing inspiration from human-driven motorsport, e.g. \emph{Formula 1}, autonomous racing competitions are structured in a two-step process: 
first come the \emph{Time-Trials}, a \emph{Qualifying} phase where autonomous agents aim to clock the fastest lap times, and then the competition culminates in the \emph{Grand Prix} where up to twenty cars ideally engage in \emph{Head-to-Head} racing.
Prominent autonomous racing leagues such as \gls{fsd} and \gls{iac} predominantly focus on the \emph{Qualifying} aspect \cite{tum_fullsystem, amz_fullsystem, kaist_fullsystem, raji2023erautopilot}, as the challenges of full-scale autonomous racing, including high costs, safety considerations, and significant engineering overhead, often necessitate restrictions in the racing scenarios. 
Conversely, small-scale autonomous racing, exemplified by \emph{F1TENTH}, offers an opportunity to fully embrace \emph{Head-to-Head} racing dynamics in a more accessible environment \cite{f110_wins, OKelly2019F110AO}. \emph{F1TENTH}'s unrestricted \emph{Head-to-Head} racing and the therefore full robotic autonomy stack required to compete, present complex algorithmic challenges, especially when further considering that the miniaturization intensifies algorithmic design challenges due to limited hardware resources. Full-scale autonomous racing solutions typically require complex, custom, or proprietary hardware and software \cite{amz_fullsystem, tum_fullsystem, kaist_fullsystem, raji2023erautopilot}, limiting reproducibility only to well-resourced research facilities. 

\begin{figure}[ht]
    \centering
    \includegraphics[height=5.3cm,trim={5cm 7cm 5cm 5cm},clip,keepaspectratio]{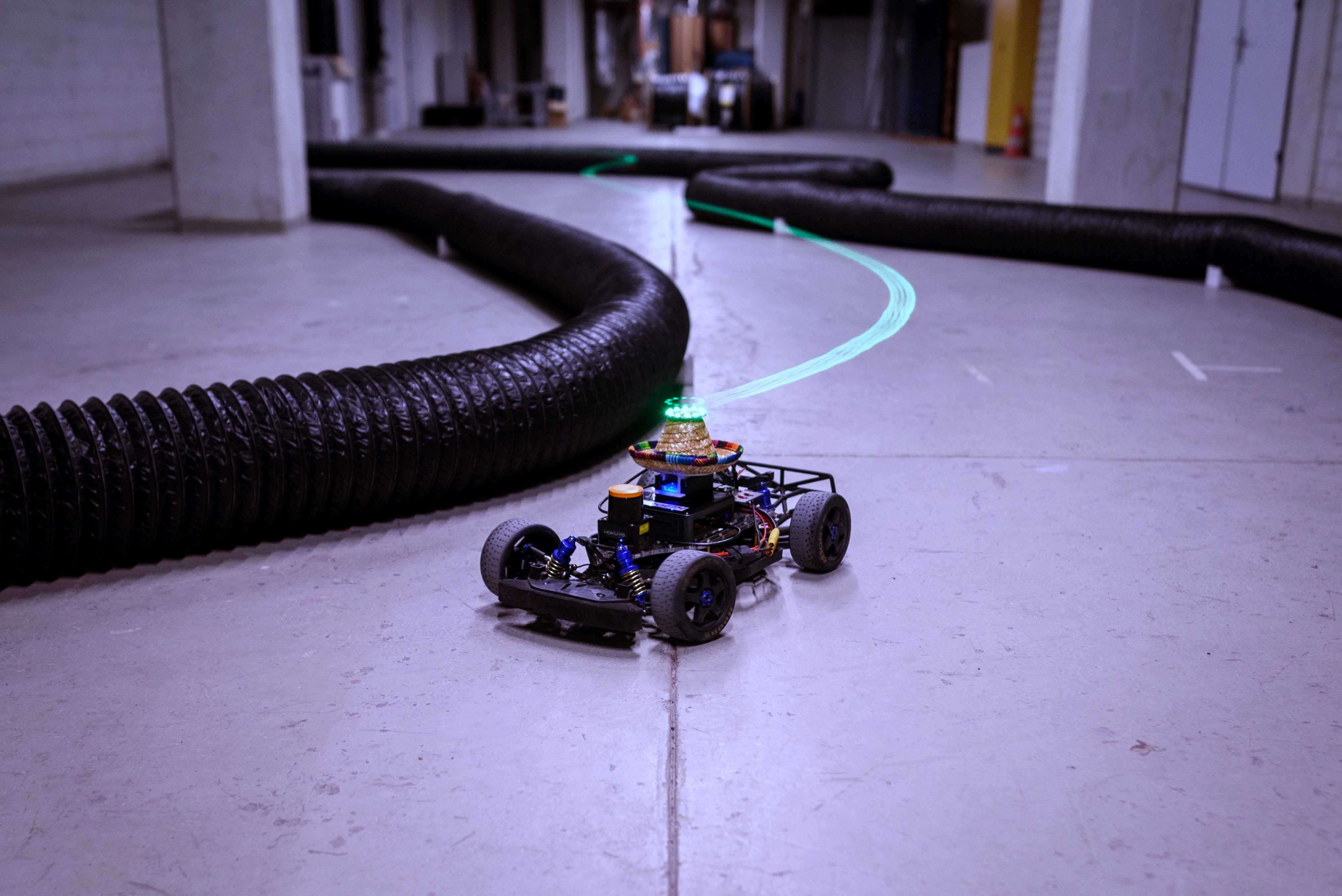}
    \hfill
    \includegraphics[height=5.3cm,trim={10cm 5cm 10cm 5cm},clip,keepaspectratio]{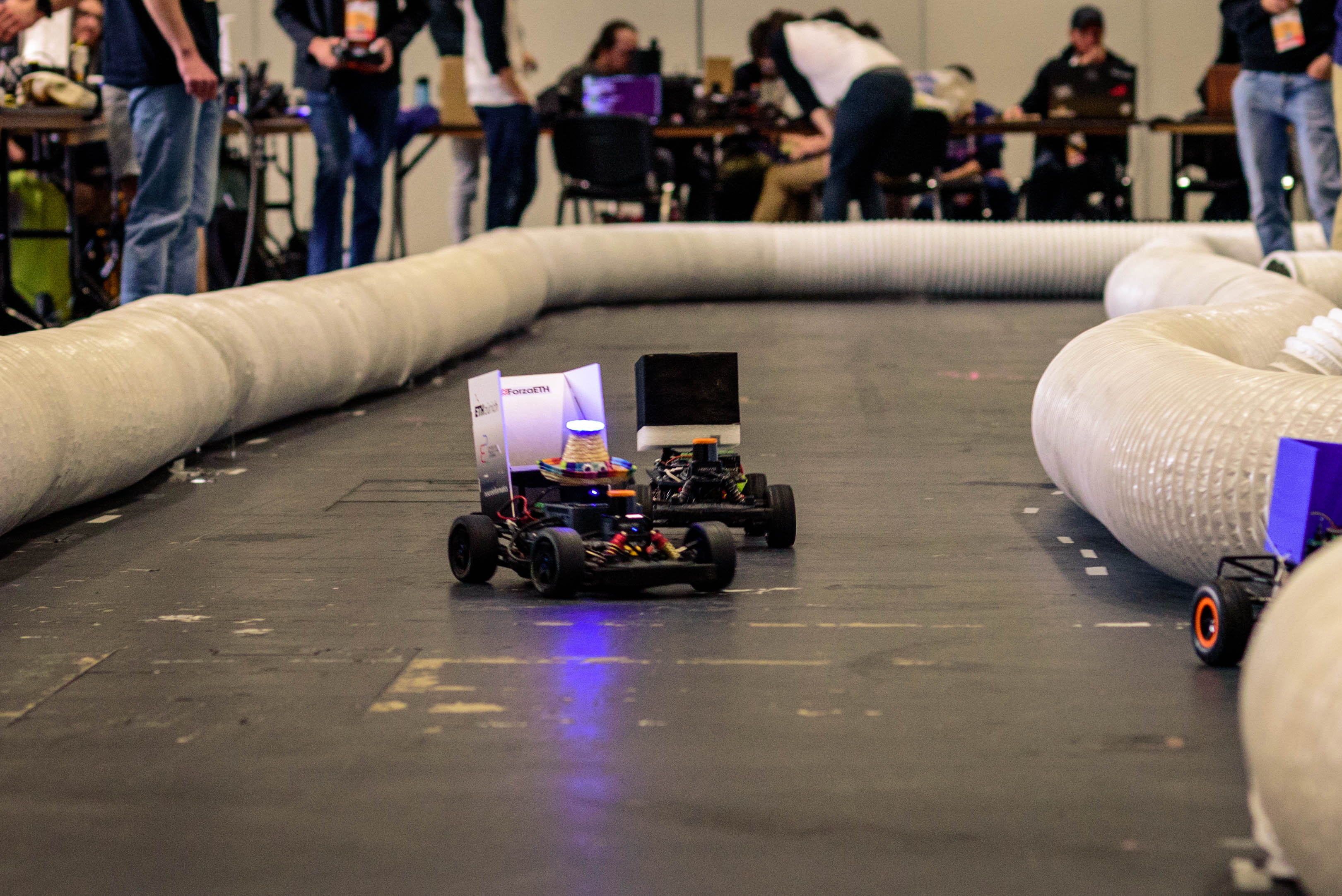}
    \hfill
    \caption{The physical \emph{ForzaETH} autonomous racecar running the proposed \emph{ForzaETH Race Stack}. On the right, an overtaking maneuver during the ICRA23 \emph{F1TENTH} race.}
    \label{fig:intro}
\end{figure}

\rev{To summarize, autonomous racing poses a promising environment to stress-test self-driving systems. The high cost of full-scale cars, which are often made from proprietary components restricts access to this research discipline to well-funded organizations and requires restrictions in the racing competitions to prevent expensive accidents.} \rev{To address these challenges}, the \emph{ForzaETH} team's \rev{1:10 scale} racecar, shown in \Cref{fig:intro}, is fully built on \gls{cots} hardware, facilitating accessibility \rev{when compared to other \gls{sota} autonomous racing platforms, that use custom components or expensive standardized hardware (see \Cref{tab:rw_platforms})}. \rev{Furthermore, this paper presents a complete autonomous racing software stack suitable to compete in a \emph{Head-to-Head} race setting, in order to facilitate participation in such a challenge to teams or research groups with only partial expertise or limited personnel.}

\emph{F1TENTH} racing competitions are typically held at robotics conferences, such as IROS and ICRA \cite{f110_wins, okelly2020f1tenth, OKelly2019F110AO}. Predominantly, university teams from all over the world compete to develop the fastest and most intelligent racecar. The race is typically structured in the following phases:
\begin{enumerate}[I]
    \item \textbf{Free Practice:} All teams can use the racetrack freely prior to the race. This time slot is typically used to map the track.
    \item \textbf{Time-Trials:} In this phase, each racecar navigates an empty track, aiming to complete as many uninterrupted laps as possible while minimizing lap time. Scores are calculated by considering both the number of consecutive crash-free laps and the best lap time.
    \item \textbf{Head-to-Head:} In \emph{F1TENTH}, the \emph{Time-Trials} establish the seeding for \emph{Head-to-Head} racing brackets. Thus, the first-place racer competes against the last, second against second last, and so forth, in a 1v1 knockout tournament. Each 1v1 battle consists of a best-of-three heat, with the victor being the first to complete 10 laps. Winners advance; losers are eliminated, continuing until a victor emerges.
\end{enumerate}

\emph{F1TENTH} imposes specific race constraints, including a 1:10 form factor, an electric drivetrain, infrastructure-less localization (no \gls{gps} or motion capture system), and most importantly, fully onboard computing. From a robotic standpoint, the competition demands a full autonomy stack for a mobile robot that pushes the robot to its limit in terms of physical traction, computational efficiency, and sensor processing. 

While the \emph{F1TENTH} competition has paved the way for research in \emph{Time-Trials} and \emph{Head-to-Head} racing, its format focuses on a 1v1 \emph{Head-to-Head} setting. The primary reason to limit this to a single opponent is the inherent complexity of the task. However, this challenge could be bridged and facilitated, by providing a fully open-sourced stack that allows for further development towards more complex challenges, like racing against multiple opponents. This work aims to do so, by offering a comprehensive and technical overview of the \emph{ForzaETH F1TENTH} autonomous racing stack and open-sourcing the full race stack, that was used when winning both the \emph{German Grand-Prix 2022} and the \emph{ICRA Grand-Prix 2023} \cite{f110_wins}. With this, we aim to contribute to the autonomous racing and research community, sharing the intricate technical and algorithmic lessons learned and providing the first complete, embedded, fully onboard, and real-time \emph{Head-to-Head} autonomous racing stack for \gls{cots} hardware. To summarise the contributions of this paper:
\begin{enumerate}[I]
    \item \textbf{Race Stack Architecture:} We describe and detail the complete \emph{Head-to-Head} capable race stack for scaled autonomous racing using \gls{cots} hardware, detailing technical and algorithmic intricacies of integrating and adapting the \gls{sota} algorithms in mobile robotics and embedded computing into a unified autonomous racing stack, adhering to the \emph{See-Think-Act} cycle.
    \item \textbf{Performance:} We provide comprehensive quantitative, and qualitative performance evaluations for each of the introduced subsystems and overall performance of the \emph{ForzaETH} race stack. This allows us to assess and compare robotic \gls{sota} algorithms in the context of autonomous racing and embedded computing.
    \item \textbf{Open-Source:} We provide the complete and race-winning robotic autonomy stack of \emph{ForzaETH} on \gls{cots} hardware, with reproducibility in mind. Hence, we enable the broader autonomous racing research community to build upon this stack, lowering the barrier of entry even more and allowing for further research in the challenging \emph{Head-to-Head} domain.
\end{enumerate}

\FloatBarrier
\section{Related Works}
Autonomous driving competitions go back to the 2005-2009 period, when the \emph{DARPA Grand Challenges} \cite{Thrun2007} and \emph{URBAN Challenge} \cite{Urmson2009} were held. The task for autonomous vehicles in such settings was to complete a predefined path on their own, either in a desert environment or in an urban one. More recently, the \emph{SAE autodrive challenge} \cite{zeus_SAE_autodrive} defines a competition where level 4 autonomous driving was the final target. 
While these competitions are of great importance for their fundamental work in sparking interest in autonomous vehicles, they are in rather different set-ups when compared to more recent autonomous racing competitions, such as \gls{fsd}, \gls{iac}, \emph{Roborace} and \emph{F1TENTH}.

Starting in 2017, one of the first examples of autonomous racing competition started to occur in the shape of the \gls{fsd} challenge \cite{amz_fullsystem}, in which the \emph{Formula Student} racing cars, previously only human-driven, needed to be adapted to be able to drive in different types of challenges. 
The competition sparked a parallel research interest, that was then manifested in multiple works detailing the high-performance algorithms, not only as a part of the system \cite{fsd_yolov3, Srinivasan_rnn_velocity, vazquez_2020_hierarchical, costa_online_2023}, but also in a broader view with the description of the full system, similarly to this work. The work of \cite{amz_fullsystem}, for example, details both the hardware and software architecture that brought the \gls{amz} team to multiple victories in their respective challenges. 
Comparably, the \emph{Roborace} \cite{Roborace} (2017-2021) and the \gls{iac} \cite{iac} (2019-now) associations organized competitions that attracted research interest both on the specific algorithms \cite{Heilmeier2020MinimumCar, christ_mintime, tum_tube_mpc_21} \cite{mixnet_2023, tum-pf, seong_23_indy}, and on full racing stack integrations \cite{rr_full_system_tum} \cite{kaist_fullsystem, tum_fullsystem, raji2022fs}. 

Compared to the full-size autonomous systems discussed, the specific platform utilized in this paper is much more accessible, as it is made up of \gls{cots} hardware, comes smaller at a 1:10 scale, and requires generally lower monetary and infrastructural investments (as can be seen in the comparison in \Cref{tab:rw_platforms}).
The forthcoming sections delve into the tangible benefits of this accessibility, as the platform turned out to be a validation platform for different research works, inspiring parallel 1:10 autonomous driving projects and attracting interest through the recurrent championships organized throughout the years.

\emph{F1TENTH} competitions have been organized since 2016 \cite{F1TENTH}, and since 2018 races have been held at least twice a year in conjunction with scientific conferences, therefore appealing to a broader and worldwide audience.
Furthermore, many other research platforms were developed simultaneously, highlighting the high flexibility and adaptability of research projects on downscaled platforms.
Such projects came out both on a similar 1:10 scale, such as the Berkeley Autonomous Racecar \cite{BARC}, the MIT RACECAR \cite{mit_racecar}, the RoSCAR \cite{roscar}, and the MuSHR racecar \cite{srinivasa2019mushr}, 
and at different smaller scales, such as Chronos \cite{chronos}, IDS3C\cite{ids3c}, Robotarium \cite{robotarium}, and the Cambridge Minicar \cite{cambridge_minicar}.
However, while there are papers detailing the specifications that define an \emph{F1TENTH} car \cite{okelly2020f1tenth, OKelly2019F110AO} and papers that describe high-level optimization toolchains for the platform \cite{tunercar_icra_2020}, there is, to the best of the authors' knowledge, no work that, similarly to \cite{tum_fullsystem, amz_fullsystem}, describes a full algorithmic architecture. 
This work aims to fill this gap by presenting the \emph{ForzaETH Race Stack}, a reproducible, autonomous robotic system deployed on \gls{cots} hardware during official competitions that achieved competitive results.

\definecolor{darkgray}{gray}{0.7}
\definecolor{lightgray}{gray}{0.92}
\begin{table}[t]
    \centering
    \begin{tabular}{cp{0.15\textwidth}p{0.15\textwidth}p{0.15\textwidth}p{0.15\textwidth}}
    \rowcolor{darkgray}
        & \textbf{F1TENTH} & \textbf{\gls{fsd}}  & \textbf{\gls{iac}}  & \textbf{Roborace}  \\
        Scale  & 1:10 & 1:1.5 & 1:1 & 1:1 \\
        \rowcolor{lightgray}
        Maximum Speed  & $\sim$\SI{15}{\metre \per \second} & $>$\SI{28}{\metre \per \second} & \SI{75}{\metre \per \second} & $\sim$ \SI{84}{\metre \per \second}\\
        Computation Units & Intel NUC or NVIDIA Jetson Xavier NX & PIP39 + NVIDIA Jetson TX 2 and RTX 1050Ti &  ADLink AVA‐3501 & NVIDIA Drive PX2  + Speedgoat Mobile Target Machine  \\
        \rowcolor{lightgray}
        Hardware Availability & Off-the-shelf components  & Fully custom platform & Platform provided by the competition & Platform provided by the competition \\
        Competition Format & \emph{Time-Trials} \newline \emph{Head-to-Head} & \emph{Time-Trials} \textdagger \newline 
        Efficiency, Business, Cost, Design & \emph{Time-Trials} \newline Overtaking Competition & \emph{Time-Trials} \\
        \rowcolor{lightgray} 
        Cost & $\sim$\SI{5000}{USD} & $>$\SI{100000}{USD} & \SI{350000}{USD} & \SI{1000000}{USD} \textdaggerdbl \\
    \end{tabular}
    \caption{Comparison of sizes, maximum speed, used computation platforms, and hardware availability of various autonomous racing platforms. The data for \gls{fsd}, \gls{iac} and \emph{Roborace} are taken, respectively, from \cite{amz_fullsystem}, \cite{tum_fullsystem}, \cite{rr_full_system_tum}. \textdagger: the \gls{fsd} competition is composed of multiple different disciplines where a single car is tested. These disciplines are here grouped under the \emph{Time-Trials} name. \textdaggerdbl: from \url{https://thearsenale.com/products/robocar}}
    \label{tab:rw_platforms}
\end{table}

In the next paragraphs, we will detail \gls{sota} works related to the specific subsystems of the stack that were previously deployed on either the \emph{F1TENTH} platform or similar autonomous racing machines, therefore being subject to the sensor and computational constraints considered in this work. An overarching comparison is then presented in \Cref{tab:rw_comparison}.

\textbf{Localization and State Estimation:} 
Low-latency localization and state estimation are needed to ensure correct knowledge of the robot's position, velocity, yaw rate, and acceleration.
The first step needed in an autonomous racing stack is obtaining the position of the ego robot in the space using localization algorithms, and, in the case of \emph{F1TENTH} autonomous racing, this typically happens in a predetermined environment. 
Localization is usually done with pose-graph based \gls{slam} techniques, such as \emph{Cartographer} \cite{loc_cartographer} and \emph{Slam Toolbox} \cite{SLAMToolbox}, or \gls{mcl} based techniques \cite{amcl}, \cite{MIT_PF}, \cite{synpf}, \cite{tum-pf}. 
To then obtain a filtered state estimate of the car, different filtering techniques are used, such as an \gls{ekf} or a \gls{ukf} \cite{ukf}, \cite{robotlocalization}, \cite{f1tenth_minicity}.
In addition, filtering techniques have also been previously described in the literature in the context of a full-system description similar to this work. 
A first example is the one proposed in \cite{WISCHNEWSKI2019roborace} where a \gls{kf}, that employs a kinematic model, is used to fuse \gls{gps} and \gls{lidar} localization based on \cite{amcl}. 
Another example is presented in \cite{amz_fullsystem}, where a highly specialized, vision-based \gls{slam} algorithm fuses velocity estimation with cone detection to obtain precise localization estimates. 
While these two latter systems were deployed on more computationally powerful computers, to the best of the authors' knowledge, no complete state estimation pipeline was deployed on \gls{cots} hardware without \gls{gps} before this paper, and only partial pieces of a system were studied, e.g. cone-based localization, which was also deployed on 1:10 scaled cars in \cite{brunnbauer_but_cones} but in a very restricted space (maximum \SI{2}{\metre} by \SI{2}{\metre}) and at presumably low velocities (below \SI{5}{\metre \per \second}). Whereas this work has been tested at speeds up to \SI{11}{\metre \per \second} \cite{map}. 

\textbf{Detection and Tracking:} 
While the \gls{sota} in 3D detection and tracking is clearly achieved with \gls{ml} techniques both for camera-based \cite{hop2023} and \gls{lidar}-based \cite{lu2023link} settings, the platform considered in this paper has a limited set of sensors at its disposal (e.g. no 3D \gls{lidar}) and an even more limited computational capacity, that does not allow for the deployment of the \gls{sota} \gls{ml} models with a sufficient latency, requiring research on either smaller models or classical algorithms.

For camera-based \gls{ml} systems, \gls{fsd} was a driving force of research, as the detection of cones is crucial for this task. The most commonly employed architecture is the one named \gls{yolo}, which was deployed on the \emph{Formula Student} platform in its v2 \cite{fsd_yolov2}, v3 \cite{fsd_yolov3} and v5 versions \cite{fsd_yolov5}.
When considering non-\gls{ml} \gls{cv} techniques, there are fewer works in general, as the standard sensor setup of the \emph{F1TENTH} car uses the less common 2D \gls{lidar}. A work that explains how such a setup can be incorporated in the \emph{F1TENTH} platform is \cite{kk_2d_lidar}, showing how basic rectangle fitting techniques and global nearest neighbor can be exploited to obtain detections that can then be incorporated in a \gls{ukf}.
We incorporate similar techniques and the \emph{Adaptive Breakpoint} method from \cite{breakpoint} into the detection and tracking system presented in this work.

\textbf{Planning:} When considering global planning, since this task is usually performed offline, the F1TENTH platform does not specify any limitations, therefore any global planner may be used to compute a racing line. An example of such a planner is the one presented in \cite{Heilmeier2020MinimumCar}, which computes a minimum curvature path iteratively solving a \gls{qp} problem and then calculates a velocity profile for such a trajectory using the longitudinal and lateral limits of the car.
This work was deployed on the \emph{Roborace}, similarly to \cite{christ_mintime}, and instead proposes an optimization program that minimizes lap time and approximates the vehicles' behavior with a single track model. For ease of tuning and the higher versatility given by the smoother minimum curvature racing line, in this work, we deployed and tested the method of \cite{Heilmeier2020MinimumCar}, nevertheless having the minimum time implementation also integrated.

On top of a global planner, a local planner can then be employed to incorporate obstacle-avoidance capabilities. Due to the online nature of such algorithms, the constraints of the platform need to be considered, making \gls{sota} algorithms harder to deploy directly, such as the graph-based planner presented in \cite{stahl2019}, that describes how a lattice can be used to search for different behavioral strategies (overtake left/right, follow, go straight) in the context of \emph{Roborace}, or the \gls{mpc} presented in \cite{Wischnewski_tubempc_2023}, that incorporates obstacle avoidance by modifying the reference racing line and boundaries used by the controller, deployed in the context of the \gls{iac}.
In the context of \emph{F1TENTH}, \gls{mpc} solutions for obstacle avoidance were considered, for example in \cite{zhu2022gaussian}, where \glspl{gp} are used to predict the opponent trajectory and incorporated into the \gls{mpc} formulation, or also in \cite{bulsarea2020mpcobstacleavoid},  where an \gls{mpc} is used to follow a trajectory generated around an obstacle by an external planner. However, both methods differ greatly from the setup of this work, as both use very low velocities or testing spaces (\cite{zhu2022gaussian} uses a max velocity of \SI{2.8}{\metre \per \second} and \cite{bulsarea2020mpcobstacleavoid} uses a \SI{4.85}{m} by \SI{3.5}{m} space). Furthermore, the obstacle was either perceived through a motion caption system in the former case or static as in the latter, and computational requirements were evaluated offline. To ease the computation of \gls{mpc} controllers, also \cite{heetmeyer2023rpgd} proposes a small-batch parallel gradient descent optimization strategy, to handle the non-linear model and non-convex constraints of an \emph{F1TENTH} setting, which is still quite different in the final velocity and setting tested on the real platform (around \SI{2.4}{\metre \per \second}).
Planning for obstacle avoidance on \emph{F1TENTH} platforms can be implemented through more standard techniques, such as with \textit{Frenet} planners, as in \cite{raji2022fs} in the context of the \gls{iac}. An accelerated version for NVIDIA CUDA platforms was described in \cite{muzzini2023frenet} and this version can be deployed on an \emph{F1TENTH} platform, e.g. when the main computation board used is the NVIDIA Jetson Xavier NX. This work, however, demonstrates fully real-time collision avoidance and overtaking trajectory generation using 
 online detections and estimations of the opponent through onboard computing and sensing.

\textbf{Control:} Similarly to the considerations done for the local planners, the best performing \gls{mpc} algorithms used in high-performance autonomous racing (e.g. \cite{tum_tube_mpc_21, hierarchical_mpc}) yield lower performances when constrained by the computation limits of \emph{F1TENTH} platforms. 
For example, the work of \cite{JainRaceOpt2020} proposes to use an \gls{mpc} with car dynamics learned with the use of \glspl{gp}. However, the test is carried out only in the context of the \emph{F1TENTH} simulator, and the \gls{mpc} solve time yielded a $\sim\SI{4}{\hertz}$ frequency.
Another \gls{mpc} strategy is presented in \cite{wang2021koopman}, where the authors propose to learn the model dynamics via data-driven Deep Koopman Representations for Control. The resulting maximum speed is however not exceeding \SI{3.5}{\metre \per \second} while the computational platform to carry out such an algorithm is unclear from the source.
The work in \cite{ALCALA2020104270} deploys a \gls{lpv} \gls{mpc} on a 1:10 scaled platform, but the computation is carried out on a remote computer and the localization is done by means of an indoor positioning system.
To avoid the complex requirements in parameter identification and computation complexity of \gls{mpc}, researchers have often resorted to \gls{ml} techniques, often in the context of \gls{rl} \cite{Ghignone_2023, brunnbaueretal, tcdrivercopy}. These methods however suffer from the sim-to-real gap, and the performance they achieve in simulation is either not tested on hardware (as in \cite{tcdrivercopy}) or tested at final speeds lower than those of simulation (as in \cite{brunnbaueretal, Ghignone_2023}).

The most performant techniques that have been deployed on \emph{F1TENTH} vehicles are closer to classical techniques than to \gls{mpc} or \gls{rl}. 
A first example is the one in \cite{tunercar_icra_2020}, where a 
\emph{Pure Pursuit} algorithm is deployed on tracks that reach up to \SI{7}{\metre \per \second} and achieves lap times up to 21\% faster when compared to expert solutions on real-world race tracks.
The controller that is tested on the highest velocities is the one presented in \cite{map}, where a dynamic model with \textit{Pacejka} tire formulas is used to extend the only geometric properties of \emph{Pure Pursuit}, reaching significantly faster laptimes at tested speeds up to \SI{11}{\metre \per \second}. This last controller is also the one used in this work, with only minor modifications.
A further controller used predominantly in \emph{F1TENTH} competitions is the \gls{ftg} controller \cite{ftg, ftg_orig}, a reactive method that processes directly the 2D-\gls{lidar} scans and steers the car towards the largest available cone of free space, without relying on neither \emph{State Estimation} nor \emph{Planning}. This type of controller, while avoiding obstacles directly as a result of its reactivity, is unfit for higher speeds, and can only drive in a circuit if a single closed trajectory is present.

\begin{table}[th]
    \centering
    \begin{tabular}{p{0.2\textwidth - 2\tabcolsep}p{0.2\textwidth - 2\tabcolsep}p{0.15\textwidth - 2\tabcolsep}p{0.16\textwidth - 2\tabcolsep}p{0.29\textwidth - 2\tabcolsep}}
        \rowcolor{darkgray}
        \textbf{Source} & Modules (Details) & Onboard
        
        Localization? & Onboard 
        
        Computation? & Testing Constraints \\
        \cite{brunnbauer_but_cones} & Localization, (Camera) & \textbf{Yes} & \textbf{Yes} & max size: \SI{2}{\metre} $\times$ \SI{2}{\metre} \\
        \rowcolor{lightgray}
        \cite{kk_2d_lidar} & Detection, Tracking & \textbf{Yes} & \textbf{Yes} & max size: \SI{4}{\metre} $\times$ \SI{4}{\metre}, max velocity: $<$\SI{3}{\metre \per \second}\\
        \cite{bulsarea2020mpcobstacleavoid} & Planning, Control (MPC) & No & No & max size: \SI{5}{\metre} $\times$ \SI{3.5}{\metre}, max velocity: N/D, constant \\
        \rowcolor{lightgray}
        \cite{zhu2022gaussian} & Local Planning, Control (\gls{mpc}) & No & No & max velocity: $<$\SI{3}{\metre \per \second}\\
        \cite{ALCALA2020104270} & Control (\gls{mpc}) & No & \textbf{Yes}\textdagger & max size: \SI{7}{\metre} $\times$ \SI{7}{\metre}, max velocity: $<$\SI{2.8}{\metre \per \second}\\
        \rowcolor{lightgray}
        \cite{brunnbaueretal} & Control (\gls{rl}) & \textbf{Yes} & \textbf{Yes} & max velocity: $<$\SI{5}{\metre \per \second}\\
        \cite{Ghignone_2023} & Control (\gls{rl}) & \textbf{Yes} & \textbf{Yes} & max velocity: $<$\SI{3}{\metre \per \second}\\
        \rowcolor{lightgray}
        \cite{tunercar_icra_2020} & Planning, Control (\emph{Pure Pursuit}) & \textbf{Yes} & \textbf{Yes} & max velocity: $<$\SI{7}{\metre \per \second}\\
        \cite{amz_fullsystem} & \textbf{Full Stack} & \textbf{Yes}\textdaggerdbl & \textbf{Yes}\textdaggerdbl & N/A\\
        \rowcolor{lightgray}
        \cite{tum_fullsystem} & \textbf{Full Stack} & \textbf{Yes}\textdaggerdbl & \textbf{Yes}\textdaggerdbl & N/A\\
        \textbf{Ours} & \textbf{Full Stack} & \textbf{Yes} & \textbf{Yes} & \textbf{max size: \SI{30}{\metre} $\times$ \SI{10}{\metre}}, \textbf{max velocity:} $<$\SI{11}{\metre \per \second}\\
    \end{tabular}
    \caption{Comparison of previous works in the context of \emph{F1TENTH}. Maximum size under the testing constraints is indicated to quantitatively assess the difference between the previous works test setups with the one of an official \emph{F1TENTH} competition (ca. \SI{30}{\metre} $\times$ \SI{10}{\metre}), as detailed for example at this link, available at the time of writing, 31st January 2024: \url{https://icra2023-race.f1tenth.org/orientation\_2.html}. N/D: No Data. \textdagger: Computation in \cite{ALCALA2020104270} is not carried out onboard, but the computing platform, using an Intel Core i7-8850U and no \gls{gpu}, is comparable in power to the \emph{F1TENTH} setup. \textdaggerdbl: \cite{amz_fullsystem}, \cite{tum_fullsystem} are not deployed in the context of \emph{F1TENTH}, but are included to highlight the previous works in the context of autonomous racing that propose full software stacks. N/A: Not Applicable. }
    \label{tab:rw_comparison}
\end{table}
%%%%%%%%%%%%%%%%%%%%%%%%%%%%%%%%%%%%System%%%%%%%%%%%%%%%%%%%%%%%%%%%%%%%%%%%%
\FloatBarrier
\section{System Overview} \label{sec:sys_overview}
This section introduces the hardware components of the racecar in \Cref{subsec:hw}. Subsequently, the design philosophy behind the proposed \emph{ForzaETH Race Stack} is presented in \Cref{subsec:sw}, emphasizing the interaction among individual autonomy modules and their collective contribution to the overall racecar architecture. Finally, the robotic conventions utilized within this work are detailed and defined in \Cref{subsec:frame_convention}.

\subsection{F1TENTH Hardware Architecture} \label{subsec:hw}

\begin{figure}[ht]
    \centering
    \includegraphics[width=0.95\columnwidth]{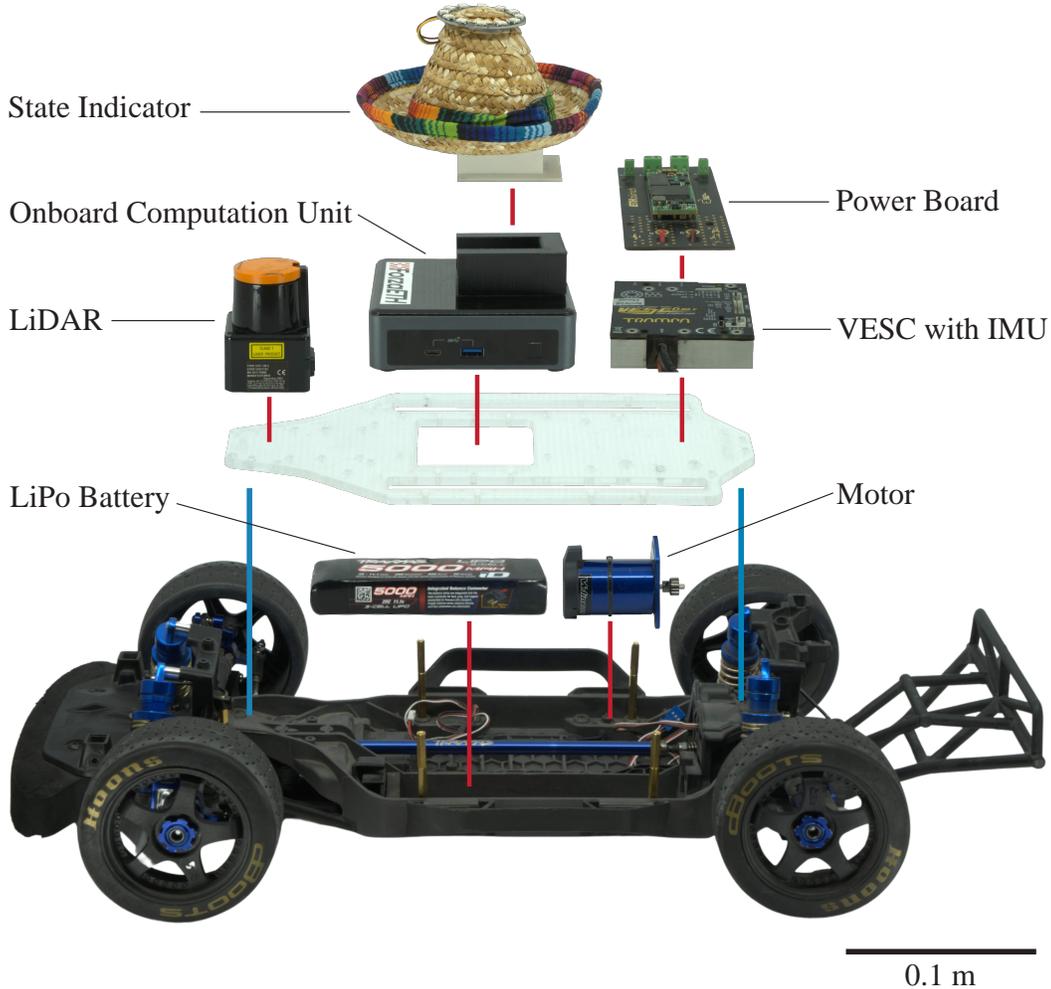}
            \caption{Comprehensive overview of the autonomous racing platform's hardware architecture. This figure presents an exploded view of the racecar, highlighting its key components as well as their integration.}
    \label{fig:hw_architecture}
\end{figure}

The foundation of the proposed robotic platform is the \emph{Traxxas TRAX68086-4FX} \gls{rc} racecar, as suggested by the official \emph{F1TENTH} bill of material. It is reduced to its core parts, namely the frame, the axles with tires and suspension, the actuator as well as front and rear bumpers. Over this foundation, an acrylic plate is mounted and acts as an even level for the mounting of the autonomy parts of the platform. Onto this platform, a \gls{lidar}, the \gls{obc}, the \gls{vesc}, acting as the motor controller, with its built-in \gls{imu} and the power distribution board are mounted. This setup can be seen in \Cref{fig:hw_architecture} while the hardware components are listed in \Cref{tab:hw_components}.

\definecolor{darkgray}{HTML}{A9A9A9} % eth 60 perc gray
\definecolor{lightgray}{HTML}{E2E2E2} % eth 20 perc gray

\begin{table}[th]
    \centering
    \begin{tabular}{lll}
        \rowcolor{darkgray}
        \textcolor{black}{\textbf{Component}} & \textcolor{black}{\textbf{Manufacturer}} & \textcolor{black}{\textbf{Model}} \\
        Tires & ARRMA & Dboots Hoons 42/100 2.9 Belted Tires Gold \\
        \rowcolor{lightgray}
        Suspension & Traxxas & Rustler VXL Aluminium Shocks \\
        \gls{lidar} & Hokuyo & UST-10LX \\
        \rowcolor{lightgray}
        \gls{vesc} & Trampa Boards Ltd. & VESC 6 MkIV \\
        \gls{imu} & Bosch & BMI160 \\
        \rowcolor{lightgray}
        Actuator & Velineon & 3500 Brushless Motor \\
        \gls{obc} Device & Intel & \gls{nuc} 10 \\
        \rowcolor{lightgray}
        \gls{obc} CPU & Intel & Core i5-10210U \\
        \gls{obc} RAM & Corsair & Vengeance \SI{32}{\giga\byte} \SI{3200}{\mega\hertz} \\
        \rowcolor{lightgray}
        Power Board & Murata Power Solutions & UWE-12/10-Q12N-C \\
        LiPo Battery &  Traxxas  & 2827X \\
        \rowcolor{lightgray}
        \gls{mcu} &  Arduino  & Micro \\
    \end{tabular}
    
    \caption{Detailed list of the key hardware components utilized in the \emph{F1TENTH} autonomous racecar. This table enumerates the type of the components, their respective manufacturers as well as the specific model of the component.}
    \label{tab:hw_components}
\end{table}

\DeclareSIUnit \mAh {mAh}
To power the racecar, a \gls{lipo} battery with a battery capacity of \SI{5000}{\mAh} is utilized and feeds power into the \gls{vesc} as well as the power board. The board regulates the battery voltage to a stable \qty{12}{\volt} and delivers power for the Intel \gls{nuc} \gls{obc} and the \gls{lidar} sensor.

Range sensing is achieved by a \emph{Hokuyo UST-10LX} \gls{lidar} system, which employs \gls{tof} technology to measure the interval between the emission of a laser beam and the reception of the reflected signal. \gls{lidar} is an active exteroceptive ranging modality, sweeping a laser beam \qty{270}{\degree} around the scene while ranging up to \qty{10}{\meter} at \qty{40}{\hertz} in case of the used model. The capabilities of the utilized \gls{lidar}, as well as comparisons with other range sensors in the context of autonomous racing are further explored in the work of \cite{Loetscher2023IllReflectivity}.

The racecar's orientation and motion are tracked by the \gls{imu}, analyzing rotational and linear movements using an accelerometer and a gyroscope to determine the 3D-orientation with 6 \gls{dof} at \qty{50}{\hertz}, as this is the same frequency at which the \gls{erpm} information is sampled from the \gls{vesc} motor controller.

The propulsion of the racecar is powered by a \emph{Traxxas Velineon 3500} brushless motor. This motor is characterized by a Kv rating of \qty{3500}{\text{RPM}\per\volt} and achieves maximum power at \qty{300}{\watt}. The golden \emph{DBoots Hoons Belted Tires} from \emph{ARRMA} are used. The challenge of slippage on various racetrack surfaces necessitates prioritizing tire traction rather than speed maximization. The golden version of the tire model is chosen for its pronounced siped tread pattern, promising the best traction among the models. The suspension is composed of the \emph{Traxxas Rustler VXL Aluminium Shocks} in the low \gls{cg} configuration. These shock absorbers are stiff, leading to enhanced stability of the racecar and consequently more predictable handling on smooth surfaces, which is the typical condition in \emph{F1TENTH} races. The stability of the racecar also simplifies the acquisition of sensor readings from the \gls{lidar} and \gls{imu} as no explicit roll and pitch compensation is needed.

To visualize the state or other telemetry data, a system using an \emph{Arduino} \gls{mcu} indicates the data on an \gls{led} ring, mounted on a sombrero hat. A magnetically detachable socket situated atop the \gls{obc} ensures secure placement of the components. 

\subsection{ForzaETH Race Stack Architecture} \label{subsec:sw}
\begin{figure}[th]
    \centering
    \includegraphics[width=\columnwidth]{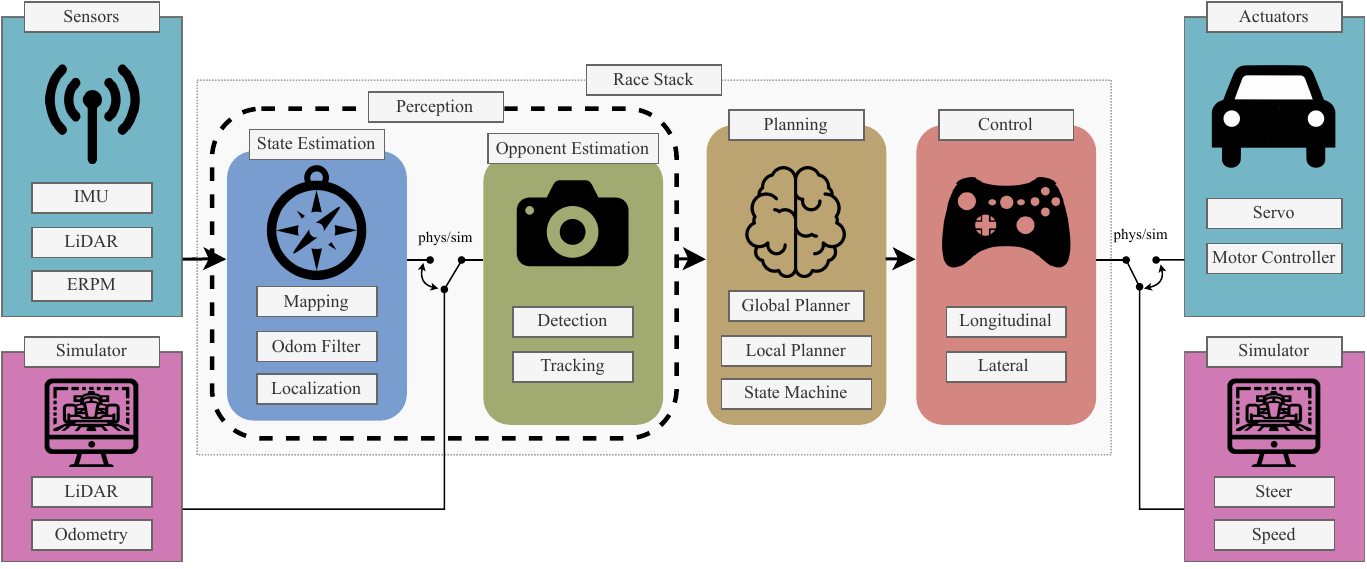}
    \caption{Architecture overview of the proposed \emph{ForzaETH Race Stack} following the \emph{See-Think-Act} paradigm. It highlights the interplay and interconnectivity of autonomy modules and hardware and illustrates how upstream tasks have knock-on effects that influence the subsequent autonomy modules. This depiction is inspired by \cite{betz_ar_survey}, but emphasizes the importance of \emph{State-Estimation} for autonomous racing, as a standalone autonomy module within \emph{Perception}. Further, it is depicted how the \emph{ForzaETH Race Stack} can switch seamlessly between the physical robot and the simulation environment.}
    \label{fig:sys_architecture}
\end{figure}

\Cref{fig:sys_architecture} illustrates the \emph{ForzaETH Race Stack}, a software architecture developed to implement the \emph{See-Think-Act} paradigm \cite{amr}, ensuring a structured and coherent approach to autonomous racing. 
\rev{In practice, such a paradigm is implemented in the shape of the \emph{Perception-Planning-Control} scheme, presented in \cite{betz_ar_survey} and successfully implemented in full-scale platforms such as in \cite{tum_fullsystem}.}
The architecture emphasizes the interaction and connectivity among various autonomy modules and hardware components, aiming to ensure efficient vehicle operation in a \emph{Head-to-Head} racing environment. \rev{The race stack software consists of the following modules:}
\begin{enumerate}[I]
    \item \rev{\textbf{Perception - State Estimation:} The state estimation module, further detailed in \Cref{chap:se}, is responsible for the fusion and filtering of multiple odometry sources for velocity estimation and localization. The state estimation module transmits the localization and velocity information obtained to all subsequent autonomy modules.}
    \item \rev{\textbf{Perception - Opponent Estimation:} In a \emph{Head-to-Head} race, the opponent estimation module, shown in \Cref{chap:perception}, is responsible for detecting and tracking an opponent racecar. The opponent estimation module utilizes the previously obtained localization information and the \gls{lidar} sensor readings to distinguish between static, and dynamic obstacles and then track said dynamic obstacles. The position and velocity of dynamic and static obstacles are forwarded to the subsequent modules.}
    \item \rev{\textbf{Planning:} The planning module consists of both a global and a local planner. Prior to a race, the global planner, detailed in \Cref{subsec:gb_optimizer}, calculates an optimized racing line for the given track in an offline setting. During the race, the local planner dynamically computes potential overtaking maneuvers based on real-time environmental changes, as described in \Cref{sec:loc_planning}. An integrated state machine within the planning module determines whether to follow the global racing line, trail an opponent, or execute an overtaking maneuver, according to the rule set outlined in \Cref{subsec:state-machine}. The resultant waypoints are then relayed to the final control module.}
    \item \rev{\textbf{Control:} The control module, is responsible for following the previously determined waypoints closely, keeping lateral deviations low, and tracking the desired velocity closely, as elaborated in \Cref{chap:control}. It computes the desired control action using state information such as velocity and position, planner-supplied tracking waypoints, and the distance to the opponent from the opponent estimation module.}
\end{enumerate} 

The architecture not only describes the functionalities of each module but also illustrates how upstream tasks can have cascading effects on subsequent autonomy modules, which is vital in scenarios demanding real-time decision-making and adaptability. For instance, even with an optimal controller, the robot will not operate effectively if its upstream state estimation task performs poorly. Thus, each part of the pipeline is meticulously adapted and configured to work holistically within the \emph{ForzaETH Race Stack}.

While the architecture of \Cref{fig:sys_architecture} draws inspiration from the work presented in \cite{betz_ar_survey, amr}, it emphasizes the criticality of state estimation as a standalone module, underscoring its significant influence on all downstream autonomy modules. 

The depicted autonomy modules of the race stack have been fully implemented in \gls{ros}1 \emph{Noetic} using \texttt{Python} and \texttt{C++}. \rev{While the results presented in this work were obtained using the \gls{ros}1 version, additional} \gls{ros}2 \rev{\emph{Humble} and \emph{Jazzy} versions have been implemented for future-proving and open-sourcing.} From a design philosophy point of view, the following is adhered to: \texttt{C++} where necessary, \texttt{Python} where possible. Latency critical nodes necessitate the speed and performance of \texttt{C++}, while the development simplicity and efficiency of \texttt{Python} can be leveraged otherwise. Using the \gls{ros} ecosystem, many open-source robotics tools, algorithms, and sensor-drivers can be purposed for the race stack.

\subsection{Simulation Environment} \label{subsec:sim}
A minimalistic and lightweight \gls{ros} simulation environment is utilized within the development of the proposed race stack. The simulator is slightly modified from the original \emph{F1TENTH} simulation environment \cite{OKelly2019F110AO}. The interfacing between the simulator and the physical system has been designed to be identical, such that seamless switching between the simulation and the physical system is enabled. The simulation model corresponds to a dynamic bicycle model, which can be selected to use either linear or \emph{Pacejka} tire dynamics, as in \cite{commonroad}. When using the race stack in the simulation, the architectural overview of \Cref{fig:sys_architecture} is nearly identical, with the exception that \emph{State-Estimation} is replaced with the ground-truth state forwarding from the simulator, as well as the sensors and actuators being provided by the simulator as well. This simulation environment allows for testing, verifying logic, and executability throughout the development process. Yet, due to the simulator's lightweight and simplicity, a considerable \emph{Sim-to-Real} gap exists, \rev{not fully capturing the actual car dynamics and not simulating varying track conditions as present in reality. More precisely, all the imperfections, discrete disturbances, and variability that come from the complex interaction of the tires with the floor are not modeled, and, similarly, tire deformation is only considered up to a limited point given the chosen \emph{Pacejka} formulas.} Hence the extrapolation of racing performance to the physical domain is not advisable, and the simulator is mostly advised for debugging purposes \cite{VD_survey}.

\subsection{Robotic Conventions} \label{subsec:frame_convention}

\begin{table}[th]
\centering
\begin{tabular}{p{0.16\columnwidth - 2\tabcolsep - 6\arrayrulewidth}p{0.40\columnwidth - 2\tabcolsep}p{0.24\columnwidth - 2\tabcolsep}p{0.20\columnwidth - 2\tabcolsep}}
    \rowcolor{darkgray}
    \textbf{Symbols} & \textbf{Description} & \textbf{Contained Symbols} & \textbf{Corresponding \gls{ros} message}\\

    \texttt{scan} & Array of 2D range data from the \gls{lidar} in the \texttt{laser} frame. & \texttt{ranges []} & \texttt{LaserScan.msg}\\ 
    
    \rowcolor{lightgray}
    \texttt{imu} & Orientation, angular velocities \texttt{va}, and linear acceleration \texttt{al} in \texttt{imu} frame. & \texttt{qx, qy, qz, qw, vax, vay, vaz, alx, aly, alz} & \texttt{Imu.msg} \\ 
    \texttt{pose} & Position and orientation \texttt{map} frame, using quaternion notation. & \texttt{x, y, z, qx, qy, qz, qw} & \texttt{Pose.msg} \\ 
    
    \rowcolor{lightgray}
    \texttt{odom} & Position, orientation in \texttt{map} frame, using quaternion notation and velocities in \texttt{base\_link} frame. & \texttt{x, y, z, qx, qy, qz, qw, vx, vy, vz} & \texttt{Odometry.msg} \\
\end{tabular}
\caption{Robotic naming conventions used within this work. Positions are indicated with \texttt{x, y, z} and quaternions are indicated with \texttt{qx, qy, qz, qw}.}
\label{tab:robotic_naming_conventions}
\end{table}

\begin{figure}[th]
    \centering
    % \includegraphics[width=0.3\columnwidth]{figures/frame/frames.png}
    % \hfillc
    \subfloat[Reference Frames \label{fig:frames}]{
        \includegraphics[width=0.29\columnwidth,trim={0cm 0cm 1cm 0.5cm}, clip]{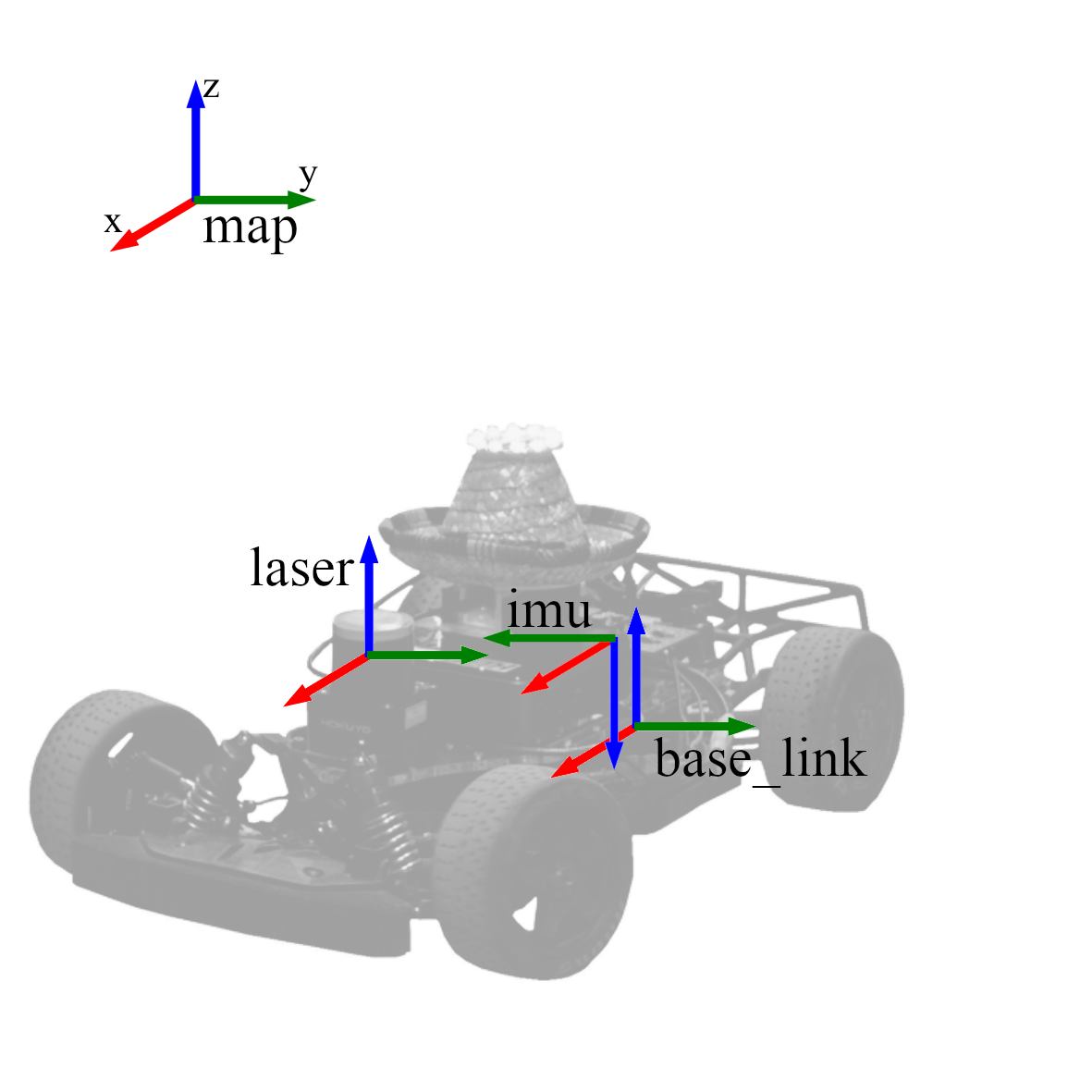}
    }
    % \vspace{1in}
    \subfloat[\emph{Cartesian} Coordinates \label{subfig:cart_coord}]{
        \includegraphics[width=0.34\columnwidth,trim={0.5cm 0.2cm 0.5cm 0.5cm}]{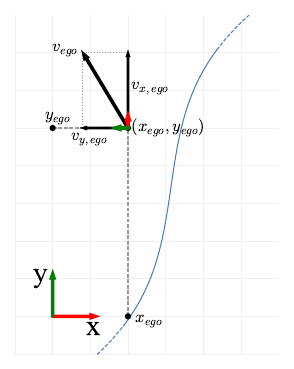}
    }
    \subfloat[\emph{Frenet} Coordinates \label{subfig:frenet_coord}]{
        \includegraphics[width=0.34\columnwidth,trim={0.5cm 0.2cm 0.5cm 0.5cm}]{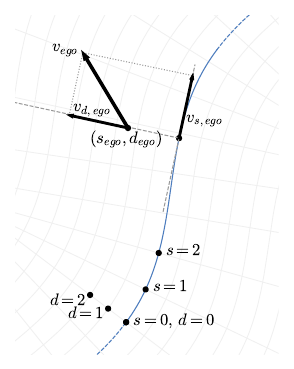}
    }
    \caption{
In \Cref{fig:frames}, the frames of reference used for the \emph{ForzaETH Race Stack} are shown. The inertial \texttt{map} frame is the reference frame for the global \emph{Cartesian} coordinates used across this work. The other frames are rigidly attached to the car, with the body frame \texttt{base\_link} at the center of the rear axle, the two sensor frames \texttt{laser} and \texttt{imu} at the center of the respective sensors.
A representation of the two used coordinate systems is further shown in \Cref{subfig:cart_coord} and \Cref{subfig:frenet_coord}, with a reference trajectory in blue. In \Cref{subfig:cart_coord}, the origin is represented by the red and green arrows, as it corresponds to the inertial \texttt{map} frame. The body frame \texttt{base\_link} is further represented with a pair of red-green arrows, centered in the position of the car. The car's velocity is further represented in both the body frame, in \Cref{subfig:cart_coord}, and in \emph{Frenet} coordinates in \Cref{subfig:frenet_coord}. \Cref{subfig:frenet_coord} further shows the \emph{Frenet} coordinate system's axis of the \emph{Cartesian} system, with the origin and a few example points on the reference axes shown in black to facilitate the reader. }
    \label{fig:coordinate_systems}
\end{figure}

Within this work, the \gls{ros} right-hand-rule coordinate convention is utilized, as in \gls{rep}-103, following convention and units as in \cite{ros_rep103}. The coordinate frames, as depicted in \Cref{fig:frames}, are \texttt{map}, \texttt{base\_link}, \texttt{imu}, and \texttt{laser}.
The body frame attached to the car is the \texttt{base\_link} frame, situated in the middle of the car's rear axle.
Two sensor frames are then also rigidly attached to the car, the \texttt{laser} frame, located at the laser sensor, and the \texttt{imu} frame, located at the \gls{vesc}'s position, where the \gls{imu} is located.
Rigid transformations link these frames to the body frame, and, whenever necessary, the sensor data is transformed to the \texttt{base\_link} frame before being used. 
The \texttt{map} frame is the inertial frame of reference and, in this frame, $(x, y)$ represent the positional and \emph{Cartesian} coordinates of the car, and $(s, d)$ represent the \emph{Frenet} coordinates, as explained in \Cref{subsec:frenet_frame}. 

The terms $v_x$ and $v_y$ denote the longitudinal and lateral velocities, respectively, in the \texttt{base\_link} frame, whereas $v_s$ and $v_d$, respectively denote the components of the velocity tangential and perpendicular to our reference trajectory, i.e. the velocity in \emph{Frenet} coordinates.
To distinguish between the ego vehicle and an opponent vehicle, the subscripts $ego$ and $opp$ are used respectively. For example, the tangential velocity of the ego and the opponent vehicle are $v_{s,\,ego}$ and $v_{s,\,opp}$. When this subscript is omitted, $ego$ is assumed.

Within this work robotic naming conventions such as \texttt{scan}, \texttt{pose}, and \texttt{odom} are used. These conventions adhere to \gls{ros} standards and are defined as in \Cref{tab:robotic_naming_conventions}.

\subsubsection{Frenet Frame Adoption} \label{subsec:frenet_frame}
A central feature of the proposed race stack is its extensive utilization of the curvilinear \emph{Frenet-Serret} frame, as detailed in \cite{frent_planner, hierarchical_mpc}. The \emph{Frenet} frame establishes a coordinate system relative to a designated reference path. In the context of our system, this reference path corresponds to the global racing line, \rev{obtained by the global planner} further discussed in \Cref{subsec:gb_optimizer}. Consequently, \emph{Cartesian} coordinates \texttt{(x,y)} can be mapped to \emph{Frenet} coordinates \texttt{(s,d)}, where \texttt{s} denotes progression along the \rev{global} racing line, and \texttt{d} signifies the orthogonal distance from the racing line. 
It is noteworthy that the \texttt{s} coordinate is cyclical, wrapping upon completing a lap, necessitating careful management of the path wrapping. 
The \texttt{d} coordinate is defined such that values to the right of the racing line are negative, while those to the left are positive.
A representation of the car's position in both \emph{Cartesian} and \emph{Frenet} coordinates is available in \Cref{fig:coordinate_systems} \rev{and further details on the conversion formulas are available in \Cref{app:frenet}}.

The adoption of the \emph{Frenet} frame offers several advantages in specific contexts. Tasks such as determining a point's position relative to the racetrack, describing motion models in relation to the \rev{global} racing line, or generating potential evasion waypoints that align with the racing line are considerably simplified within the \emph{Frenet} frame compared to the \emph{Cartesian} system. Succinctly, any computation involving spatial coordinates relative to the racing line benefits from the \emph{Frenet} frame transformation. As such, the \emph{ForzaETH Race Stack} places significant emphasis on simple and efficient transformations between \emph{Cartesian} and curvilinear coordinates, ensuring seamless and simplified operation.

\FloatBarrier
\section{State Estimation}\label{chap:se}

Accurate and robust state estimation is essential in autonomous mobile robotics, particularly in high-performance scenarios such as autonomous racing. Within the \emph{See-Think-Act} cycle illustrated in \Cref{fig:sys_architecture}, the \emph{State Estimation} module is the foundational element that precedes and influences crucial downstream tasks including \emph{Opponent Estimation}, \emph{Planning}, and \emph{Control}. The quality of state estimation data directly impacts the extent to which a racecar's performance can be optimized, pushing it to its physical limits. Therefore, precision in state estimation is not just a technical requirement; it is a critical factor that determines the racecar's overall performance and safety \cite{betz_ar_survey, synpf}.

In this chapter, we detail the methodologies and algorithms employed in the \emph{ForzaETH Race Stack} to extract accurate pose (position and orientation) and longitudinal velocity from the raw sensor readings. This involves sensor fusion of data from \gls{lidar}, \gls{imu}, and wheel-odometry (obtained through \gls{erpm} data) to achieve a reliable and high-fidelity representation of the vehicle's state, which is indispensable for executing complex racing maneuvers at the car's limit of friction. 

\subsection{Architecture}
\begin{figure}[ht]
    \centering
    \includegraphics[width=1.0\textwidth]{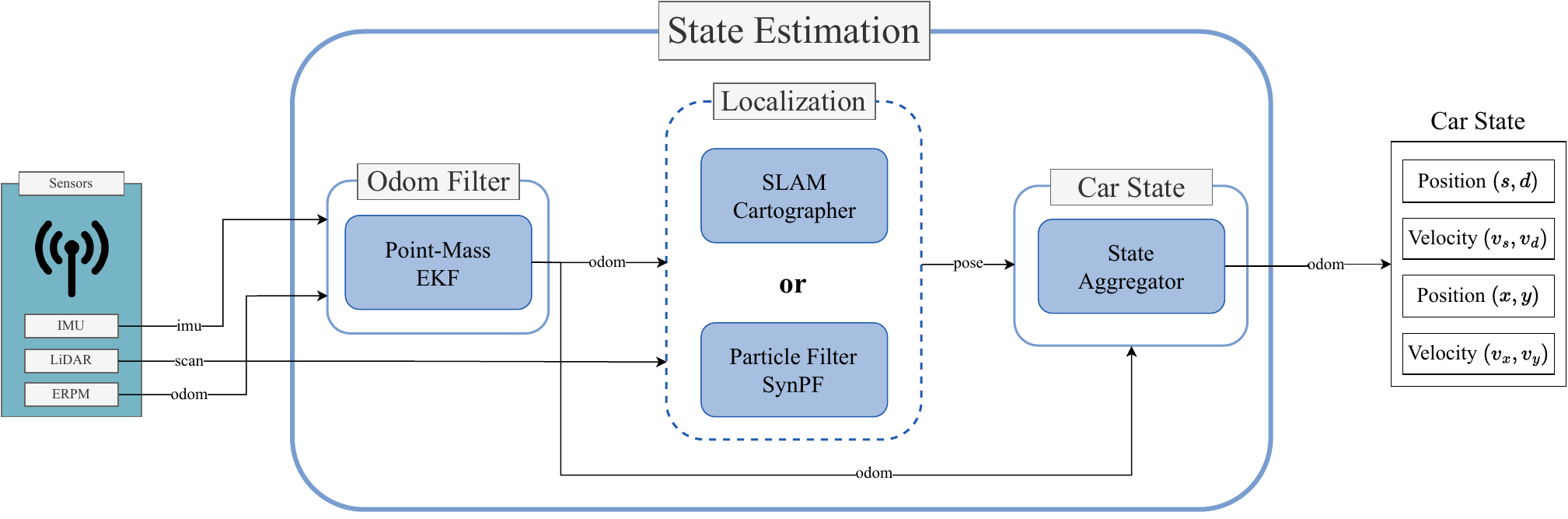}
    \caption{An overview of the proposed state estimation system architecture. The state estimation module incorporates velocity estimation and localization with respect to the pre-mapped racetrack and aggregates this information in a final car state odometry output both in Cartesian and in \emph{Frenet} coordinates.}
    \label{fig:se_loc_pipeline}
\end{figure}

The task of state estimation is divided into localization and velocity estimation. The pipeline is depicted in \Cref{fig:se_loc_pipeline}. The sensor inputs used are linear accelerations and angular velocities from the \gls{imu}, a 2D laser scan from the \gls{lidar}, and \gls{erpm} obtained wheel-odometry from the \gls{vesc} motor controller. The \gls{erpm} odometry is computed through the measured current and voltage within the \gls{vesc} motor controller and it is combined with the commanded steering angle to estimate the wheel odometry using the implementation of the \emph{F1TENTH} platform presented in \cite{okelly2020f1tenth}, which can be significantly affected by tire-slip.

Within the \emph{State Estimation} module, this odometry signal is then fused with \gls{imu} data in the \emph{Odom Filter} module (more details in \Cref{chap:ekf}).
This first filtering step is crucial, as not only are sufficiently accurate velocity estimates required for the control algorithm but \cite{synpf} has also shown that later localization modules are significantly sensitive to the accuracy of the odometry prior.
The filtered odometry signal is then fed to the localization algorithm, which is either the \gls{slam}-based \emph{Cartographer} \cite{loc_cartographer} or the \gls{mcl}-based \emph{SynPF} \cite{synpf}. The localization algorithms are mutually exclusive and both of them are described in \Cref{chap:localization} and evaluated later in \Cref{chap:SE_results}. The reason for having two mutually exclusive methods for localization is that both techniques have fundamentally different operation characteristics. The \emph{SynPF} \gls{mcl}-based localization tends to be more robust towards wheel-slippage than the \emph{Cartographer} \gls{slam}-based approach, yet \emph{Cartographer} performs smoother and more accurately under nominal conditions \cite{synpf}. Having both methods available underscores the strategic advantage of the \emph{ForzaETH Race Stack}, enabling adaptability and optimization of race strategy based on specific track conditions and requirements. Lastly, the final car state is aggregated merging the localization pose from \emph{Localization} and the velocity signal from \emph{Odom Filter}. Localization is carried out in a pre-mapped racetrack, and the mapping procedure is further described in \Cref{chap:mapping}.

\subsection{Odometry Filter - Extended Kalman Filter} \label{chap:ekf}
Velocity state estimates are generated using an approach based on the \gls{ekf} \rev{\cite{ekf} presented in the \texttt{robot\_localization} \gls{ros} package \cite{robotlocalization}. The use of such a filter enables the fusion of multiple noisy measurements, with customizable variances, in a nonlinear state estimator}.
It fuses the wheel odometry and the data provided by the \gls{imu} sensor. This aids the longitudinal and lateral velocity estimation for the robot in situations where tire slip occurs, as the accelerometer data from the \gls{imu} can be leveraged to compensate for this within the \gls{ekf}.\revdel{The EKF is integrated through the \texttt{robot\_localization} package [Moore and Stouch, 2014].}
The estimates are generated at a rate of \qty{50}{\hertz}, corresponding to the \gls{imu} and wheel-odometry update frequency.

The \gls{ekf} model consists of an omnidirectional, three-dimensional, point-mass motion model. The state $X$ and the discrete-time transfer function $f(X)$ used in the library are defined in the \Cref{sec:ekf_filter_definition}. To account for the fact that the \emph{F1TENTH} setup can be considered two-dimensional, the \texttt{two\_d\_mode} parameter is then set to \texttt{true}, which effectively enforces the measurements of the states $z,\,\phi,\,\theta,\,\dot{z},\,\dot{\phi},\,\dot{\theta},\,\Ddot{z}$ to zero, the respective covariances to $10^{-6}$ and the states to 1 (which have then to be neglected).

The remaining states that are then used by this approach are the longitudinal velocity, lateral velocity, and the yaw rate. However, if no lateral acceleration data is supplied to the motion model, the vehicle's lateral velocity is assumed to be \qty{0}{\metre\per\second}, which is equivalent to assuming that the vehicle experiences no lateral slip. While more vehicle-specific motion models have been investigated, this approach was observed to be more robust to significantly varying track conditions than the implemented vehicle models and performed sufficiently well under realistic driving conditions.

As outlined in the documentation of the \texttt{robot\_localization} package \cite{robotlocalization}, the a posteriori update step of the \gls{ekf} can be configured by selecting the specific measurement sources from each sensor or input to be considered. 
The fusion configuration is determined experimentally using ground-truth motion capture data and qualitative observations, as described in \Cref{chap:SE_results}. In the selected configuration, the \gls{ekf} fuses the \gls{imu} measurements of angular velocity, heading, and the full twist stemming from the wheel odometry. Note that linear accelerations observed by the \gls{imu} sensor are not considered (linear acceleration of the \gls{erpm}-based wheel-odometry is however used), as they were observed to experience significant noise due to shaking and may be sensitive to the exact positioning and orientation of the \gls{imu} sensor on the car. Nevertheless, the configuration of \gls{ekf} inputs can be changed quickly and easily to maximize the estimation accuracy given the external conditions.

The covariances associated with the individual measurements are determined at their respective sources and are used in forming the measurement update of the \gls{ekf}. In this implementation, the covariances of the \gls{imu} data and the control odometry are predetermined and set  to be static. Numerical values can be found in the \Cref{sec:app_SE}.

\subsection{Mapping} \label{chap:mapping}
Mapping is a fundamental part of autonomous racing, serving as the foundation for both localization and trajectory planning \cite{betz_ar_survey}. The proposed \emph{ForzaETH Race Stack} uses a pose-graph optimization \gls{slam} method, \emph{Cartographer} \cite{loc_cartographer} to create an initial map of the racetrack.

In a race setting, mapping is conducted during the free practice session. The occupancy grid produced is subsequently used for the computation of a global trajectory, detailed in section \Cref{subsec:gb_optimizer}. The most relevant tuning parameters for \emph{Cartographer} are listed in \Cref{sec:app_SE}.

\subsection{Localization} \label{chap:localization}

\begin{figure}[ht]
    \centering
    \includegraphics[width=\textwidth]{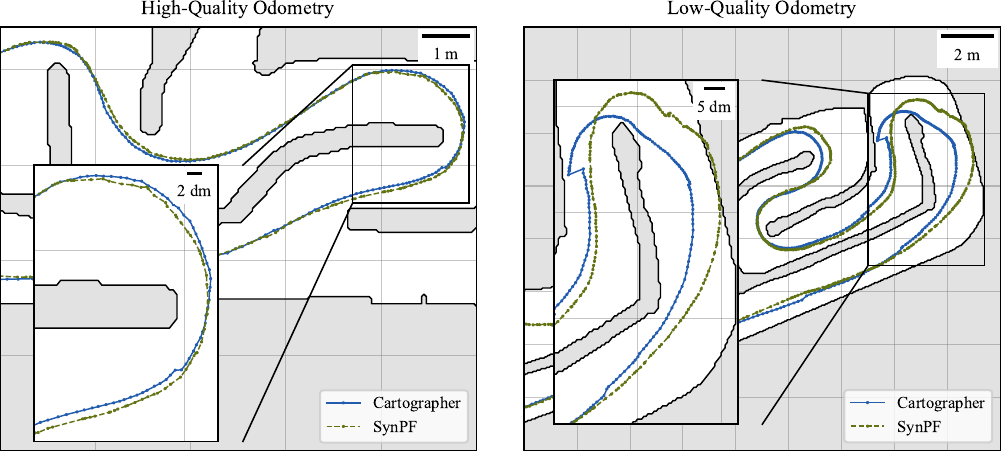}
    \caption{A comparison of the supported localization methods, \emph{Cartographer} and \emph{SynPF}. \textbf{Left:} Nominal conditions showing high-quality odometry. \emph{Cartographer} is the favorite method in this case, given the smoother pose and the higher operational frequency (\SI{200}{\hertz} versus \SI{50}{\hertz} for \emph{SynPF}).
    % \textit{Cartographer} publishes pose estimates at a higher frequency than \emph{SynPF}, shown by the more densely-spaced data points along its reported trajectory. 
    \textbf{Right:} Low-grip conditions with wheel spin. \textit{Cartographer} is unable to provide an accurate pose estimate, estimating that the racecar is in the middle of the track, while in reality, it has crashed into the boundaries, as evident from the \gls{lidar} scans in \Cref{fig:badodom_icra22}. However, \textit{SynPF} operates nominally, correctly estimating the racecar's actual position. 
    This comparison was made possible by retrospectively applying the \emph{SynPF} algorithm to telemetry data recorded during the car's operation with \emph{Cartographer}.}
    \label{fig:localization-method-pros-cons}
\end{figure}

\begin{figure}[ht]
    \centering
    \includegraphics[trim=0 0 0 0cm, clip, width=\textwidth]{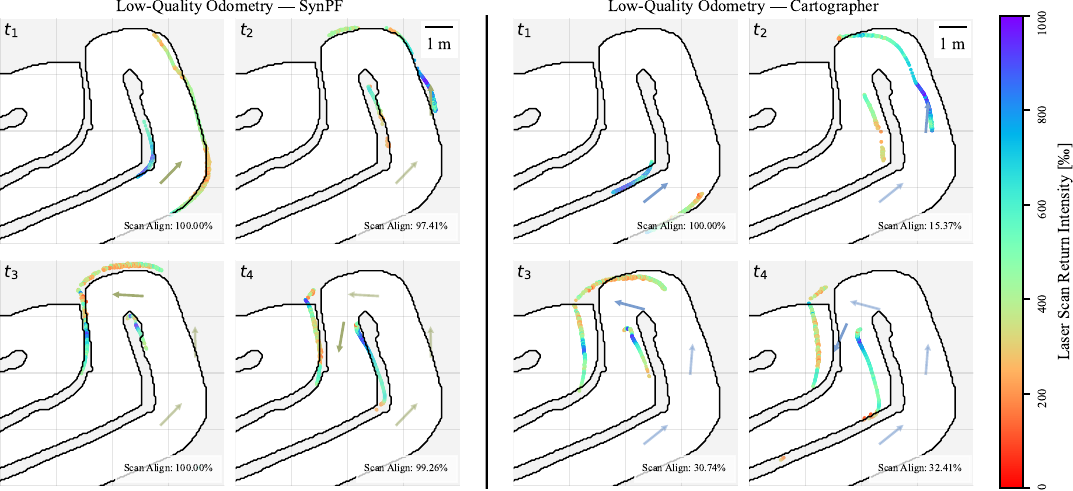}
    \caption{Low-quality odometry conditions corresponding to the right-hand side of \Cref{fig:localization-method-pros-cons} showcasing both \emph{Cartographer} and \emph{SynPF} under worst-case conditions in four sequential time steps, indicated with $t_i$. At $t_2$, \emph{Cartographer} is unable to align the \gls{lidar} scans with the track boundaries, while \emph{SynPF} maintains a visibly steady track alignment.}
    \label{fig:badodom_icra22}
\end{figure}

Accurate localization is crucial to the trickle-down effects on the rest of the \emph{ForzaETH Race Stack}. Therefore, the proposed race stack supports two mutually exclusive options for localization to suit varying conditions:
\begin{enumerate}[I]
    \item \textbf{Cartographer:} A pose-graph optimization \gls{slam} method \cite{loc_cartographer}. This localization method yields the most accurate and smoothest pose estimate, given high-quality odometry input data. This localization technique yields high accuracy localization up to \gls{rmse} of \SI{0.0535}{\metre}, as long as the racecar experiences relatively low levels of tire slip, such that the wheel odometry signal is sufficiently accurate, as further demonstrated in \cite{synpf}. 
    % It is a \gls{cpu} bound algorithm which is responsible for 67.5\,\% of \gls{cpu} utilization of the \emph{Race Stack}.

    \item \textbf{SynPF:} A high-performance \gls{mcl}-based \gls{pf} localization technique optimized for racing, specifically developed for the \emph{ForzaETH Race Stack} \cite{synpf}. As opposed to \emph{Cartographer}, this method yields slightly less accurate pose estimates with higher jitter, as can be seen in \Cref{fig:results_SE_1}. 
    % This lack of accuracy may be due to the approximations used. 
    However, this method is highly robust against low-quality odometry input. In a racing environment with high levels of wheel slip\rev{, due to for example a slippery floor}, \emph{SynPF} is a highly effective localization alternative. 
    % Furthermore, \emph{SynPF} is computationally lighter, responsible for 55.3\,\% of \gls{cpu} utilization of the \emph{Race Stack}. This is because it can be offloaded to memory by utilizing a \gls{lut}, or it can be computed in parallel on an available \gls{gpu}.
    Furthermore, \emph{SynPF} is computationally lighter, resulting in a \gls{cpu} utilization from 30\% to 50\% lower as compared to the \emph{Cartographer} counterpart, as from \Cref{ssubsec:tt_comp}, \Cref{ssubsec:h2h_comp}.
\end{enumerate}

Given the contrasting strengths of each localization method (pure performance versus robustness), human operators can decide which one to use given the conditions at each race event. A qualitative example to assess the different performance characteristics between \emph{Cartographer} and \emph{SynPF} given high- and low-quality odometry, i.e. wheel-slip, input is \rev{further} illustrated in \Cref{fig:localization-method-pros-cons} and \Cref{fig:badodom_icra22}. Specifically, \Cref{fig:badodom_icra22} demonstrates the catastrophic failure case of \emph{Cartographer} \gls{slam}, which could not handle the compromised odometry signal, leading to a collision against the track's right barrier, as evidenced by the \gls{lidar} scans. In contrast, \emph{SynPF} was able to accurately localize using the same dataset.
As in a racing scenario, one can not rely on a motion-capture system to be available, lap time is used as a proxy measurement for localization accuracy, as from a holistic viewpoint of the race stack, the improvement of localization yields overall better performance. Hence both \emph{Cartographer} and \emph{SynPF} algorithms were tuned to minimize the lap time in a \emph{Time-Trials} scenario over multiple maps. The utilized parameters are listed in \Cref{sec:app_SE}. 

%%%%%%%%%%%%%%%%%%%%%%%%%%%%%%%%%%%%%%%%%%%%%%%%%%%%%%%%%%%%%%%%%%%%%%%%%%%%%%%%%%%%%%%%%%%%%%%%%%%%%%%%%%%%
\subsection{State Estimation Results} \label{chap:SE_results}

\begin{figure}[ht]
    \centering
    \includegraphics[width=\textwidth]{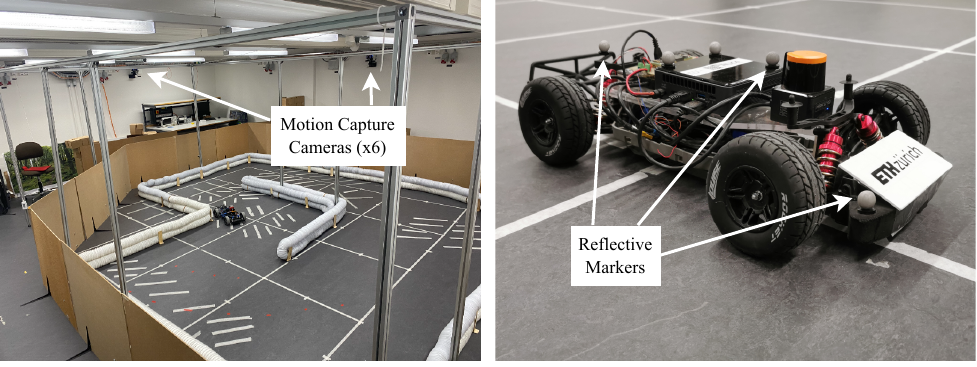}
    \caption{Experimental setup with the motion capture system and mounted reflective markers.}
    \label{fig:VICON}
\end{figure}

This section evaluates the accuracy of the entire state estimation framework against ground truth data, primarily using a motion-capture system for validation. The test vehicle was equipped with reflective markers and navigated autonomously on the track depicted in \Cref{fig:VICON}. Positional data was recorded from a motion-capture system composed of 6 \emph{Vicon Vero v2.2} cameras, recording position in a \SI{4}{\metre} by \SI{4}{\metre} space.
Furthermore, we also recorded the vehicle's sensor and control data, synchronizing measurements via the \gls{ros} timestamp. 
This comprehensive dataset enables offline analysis of different state estimation frameworks, including the two localization algorithms previously discussed. \Cref{fig:VICON} illustrates the experimental setup.
The nominal parameters described in \Cref{sec:app_SE} were used, and data was recorded over one minute of uninterrupted driving.

\subsubsection{Localization Accuracy} \label{chap:localization_results}
\Cref{fig:vicon-pf-slam-results} quantifiably shows that \emph{Cartographer} is capable of delivering accurate pose estimates, especially with the high-quality wheel odometry inputs in the given scenario, with an average positional \gls{rmse} of \SI{0.0535}{\meter}. However, the limited coverage area of the motion capture system necessitated a small racetrack setup. \emph{SynPF} struggles to give accurate localization in this scenario, with a higher \gls{rmse} of \SI{0.1998}{\meter}.

However, it should be stressed that the racetrack setup with the motion capture system is significantly smaller compared to a competition scenario. Furthermore, to assess the localization performance with respect to the quality of the wheel odometry signal, larger racetracks are needed. This is particularly necessary to capture wheel-slip effects at the edge of traction \cite{synpf}. \rev{Qualitative results of the localization accuracy on tracks that are similar in length to competition tracks, i.e. \SIrange{30}{60}{\meter}, can be found in \Cref{chap:res_tt}. An additional qualitative stress test on an \SI{150}{\meter} long race track is shown in \Cref{app:long_track}.}

\subsubsection{Velocity Estimation Accuracy}
\begin{figure}[ht]
    \centering
    \begin{minipage}{0.27\textwidth}
        \vspace{-1.1em}
        \centering
        \includegraphics[width=\textwidth]{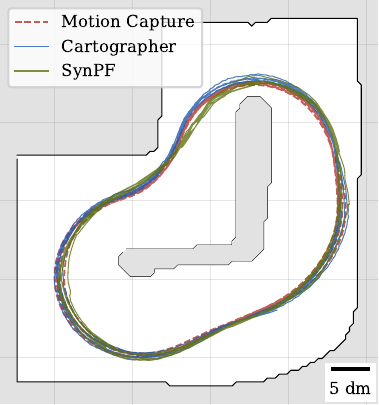}
        \caption{On the small racetrack set up with the motion capture system, \emph{Cartographer} is able to closely match the ground-truth trajectory, while \emph{SynPF} performs slightly worse.}
        \label{fig:vicon-pf-slam-results}
    \end{minipage}\hfill
    \begin{minipage}{0.70\textwidth}
        \centering
        \includegraphics[width=0.49\textwidth]{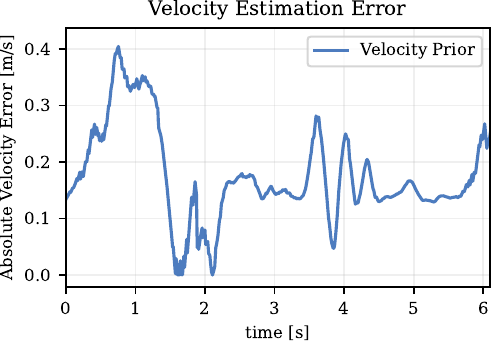}
        \includegraphics[width=0.49\textwidth]{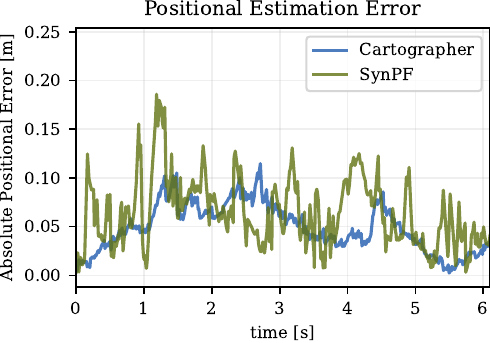}
        \caption{State estimation error curves for cartesian coordinates and longitudinal velocity corresponding to one lap around the test track. Where relevant, the two different localization techniques available are compared against ground-truth data, and the velocity and positional error are depicted for the duration of a single lap.}
        \label{fig:results_SE_1}
    \end{minipage}
\end{figure}

The longitudinal velocity state estimates produced by the two systems in question are depicted in \Cref{fig:results_SE_1}. The algorithm was executed for approximately one minute in order to calculate performance metrics, while the trajectories were plotted for a reduced time corresponding to two laps of the test track. The \gls{rmse} over the entire testing time was then computed for both localization frameworks, as summarized in \Cref{tab:SE_rmse}. The localization performance yields a low positional \gls{rmse} of $\sim$ \SI{0.08}{\metre}, on the other hand, the velocity estimation \gls{rmse} yields a rather high $\sim$ \SI{0.25}{\metre\per\second} and still leaves room for improvement. A possible reason for the worse performance of the velocity estimation is, that the velocity directly originates from the \emph{Odometry Filter} as in \Cref{chap:ekf} which fuses the information from \gls{erpm} wheel-odometry and \gls{imu} data, where the \gls{erpm} data is highly influenced by the friction of the track and can easily result in a low-quality odometry signal due to wheel-slip. \rev{Furthermore, the \gls{imu} sensor is susceptible to internal drift due to bias changes. An additional sensor that utilizes static reference points, such as a visual sensor (optical flow, visual odometry) could be used in the future to improve the velocity estimation.}

\begin{table}[ht]
    \centering
    \begin{tabular}{c|c|c|c|c}
         & \textbf{${v_{lon}}$} & \textbf{$s$} & \textbf{$d$} & \textbf{$\psi$} \\
        \emph{Configuration} & \gls{rmse} $\downarrow$ & \gls{rmse} $\downarrow$ & \gls{rmse} $\downarrow$ & \gls{rmse} $\downarrow$ \\
         & [\unit{\metre\per\second}] & [\unit{\meter}] & [\unit{\meter}] & [\unit{\radian}] \\
        \hline
        \hline
        \emph{Odom Filter} + \emph{Cartographer} & 0.1970 & \textbf{0.0729} & \textbf{0.0408} & \textbf{0.0284} \\
        \emph{Odom Filter} + \emph{SynPF} & \textbf{0.1893} & 0.0795 & 0.0725 & 0.0654 \\
    \end{tabular}
    \caption{State estimation accuracies of various odometry states while driving several laps around the test track as evaluated against motion capture data on the track setup outlined in \Cref{chap:SE_results}.}
    \label{tab:SE_rmse}
\end{table}

\FloatBarrier
\section{Opponent Estimation}\label{chap:perception}
The \textit{Opponent Estimation} module plays a pivotal role in consistently and precisely capturing moving objects, a capability imperative for successful planning and agile maneuvering throughout the race \cite{betz_ar_survey}. This module encompasses the processes of obstacle detection, classification, and tracking, extending to the computation of the opponent's position and velocity during the \emph{Head-to-Head} phase of the race. 

\subsection{Architecture}
The opponent estimation module requires the precise localization of the racecar in relation to the race track, as well as the raw sensor readings of its main exteroceptive sensor, the \gls{lidar}. Extensive trials and observations of the \emph{ForzaETH Race Stack} under racing conditions have underlined the necessity of this module for critical downstream robotic tasks, such as trailing closely behind an opponent, where it ensures an uninterrupted and precise opponent estimation. This becomes even more vital when the opponent is out of \gls{los}, potentially hidden behind a curve, as it prevents unnecessary braking or potential collisions due to misjudgment of the opponent's position. To meet these challenges, the opponent estimation architecture is designed with the following key objectives:

\begin{enumerate}[I]
\item \textbf{High Precision Detection:} The detection submodule is designed to achieve a high \gls{tpr}, ensuring consistent opponent detection whenever they are within \gls{los}. This has been evaluated to be 96.8\%, as in \Cref{subsec:od_acc}.
\item \textbf{Low False Detection Rate:} The detection submodule is optimized to minimize the number of \glspl{fdr}, enabling the racecar to maintain high speeds without unwarranted phantom breaking due to \glspl{fp}. This has been evaluated to be 1.6\%, as in \Cref{subsec:od_acc}.
\item \textbf{Continuous Opponent Estimation:} The tracking submodule is tasked with providing a continuous estimation of the opponent's position and velocity, even in the absence of \gls{los}. The position and velocity estimation of the opponent has been evaluated to yield an \gls{rmse} of \SI{0.17}{\metre} and \SI{0.49}{\metre\per\second} respectively, as in \Cref{fig:detection_res}.
\item \textbf{Low Latency:} The entire perception module is streamlined for efficiency, aiming to minimize latency and uphold reliable detection and tracking at high velocities. The latency has been evaluated to be \SI{6.39}{\milli\second} as later evaluated in \Cref{fig:h2h_compute}.
\end{enumerate}

\begin{figure}[ht]
    \centering
    \includegraphics[width=\columnwidth]{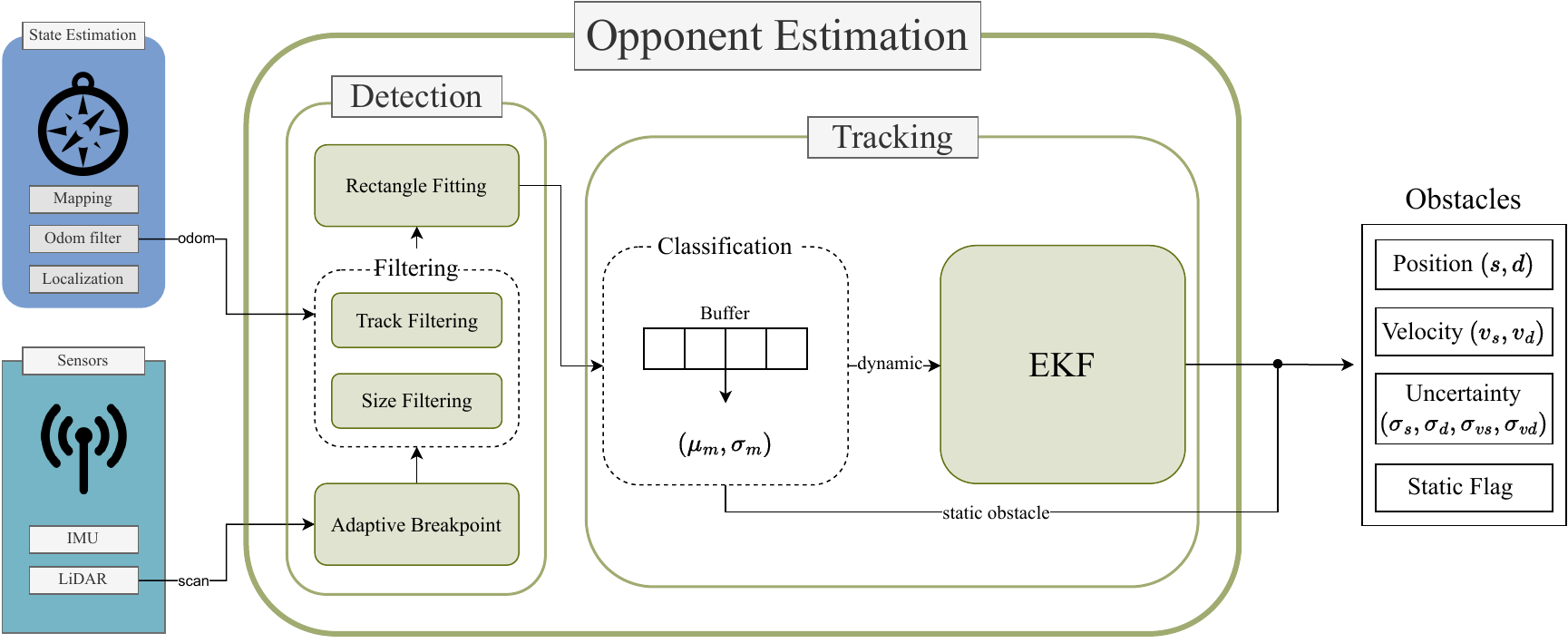}
    \caption{Overview of the proposed opponent estimation system architecture. The opponent estimation module incorporates the detection and tracking submodules and is integrated into the \emph{ForzaETH Race Stack}, as depicted in \Cref{fig:sys_architecture}.}
    \label{fig:perception_arch}
\end{figure}

\subsubsection{Opponent Detection}
The detection process \rev{is loosely inspired by the system presented in \cite{kk_2d_lidar} and} begins with acquiring the complete 2D \gls{lidar} scan data from the current \gls{lidar} sweep. The data is then segmented into smaller clusters representing potential obstacles using an \emph{Adaptive Breakpoint} method \cite{breakpoint}. This method is based on the premise that objects on the track are formed by consecutive \gls{lidar} points, and it segments the measured points of the \gls{lidar} by identifying gaps between objects that exceed a threshold distance.

    \revdel{Following segmentation,}\rev{However, this method may yield a high number of \glspl{fp}, hence to optimize the detection performance,} the point clusters are filtered to ensure a low \gls{fp} fraction. By utilizing the known positions of the track boundaries and the ego-localization from state estimation, the algorithm eliminates clusters that represent the track itself. This is achieved by inflating the track boundaries by a predefined distance, which is a parameter that can be adjusted to minimize \gls{fp}. This form of track filtering, represents a simple computation, as the curvilinear \emph{Frenet} transformation allows for thresholding of the $d$ coordinate. \rev{The level of boundary inflation is track-dependent, i.e. if the robotic operator identifies consistent \glspl{fp} on the track boundaries, then the boundary inflation can be increased as a simple measure to get rid of them.} Additionally, clusters with an insufficient number of \gls{lidar} points are discarded to further reduce the likelihood of \glspl{fp} caused by \gls{lidar} reflections. 

The remaining clusters are then approximated with rectangles, on the 2D map space, facilitating better feature extraction for subsequent processes. These features, including the size, center coordinates, and rotation of the rectangle, are crucial for the planning algorithm. A final filtering step based on the estimated size of the obstacles, i.e. removal of objects that are smaller than a given threshold, ensures a further reduction in \gls{fdr}, thus yielding an efficient detection with high \gls{tpr} (97\%) and low \gls{fdr} (2\%). \rev{Hence, in case opponents are significantly larger or smaller than the expected rectangle filtering parameters, then this should be adjusted accordingly. The complete set of utilized} parameters can be found in \Cref{sev:app_perception}.

\subsubsection{Opponent Tracking} \label{subsec:tracking}
Ensuring precise and continuous tracking of the opponent is crucial for downstream robotic tasks, both when closely following an opponent and also when the opponent is out of sight, such as when navigating around a corner, necessitating a reliable estimation to maintain safety and performance. To address this, a tracking submodule is incorporated, utilizing a racing line-based motion model, i.e. assuming that the opponent progresses along the racing-line potentially with a lateral offset. The model is implemented based on the fitted obstacle rectangles and propagated within an \gls{ekf} framework to offer consistent opponent estimation, regardless of \gls{los}.

The tracking sub-module classifies detected objects as static or dynamic based on their temporal movement patterns. A voting system, which leverages the standard deviation of recent positional data, aids in this classification process. For static obstacles, their position is deduced from the average of recent $(s_{opp},d_{opp})$ coordinate measurements. On the other hand, dynamic obstacles, such as the opponent's car, are tracked using an \gls{ekf}, as in \Cref{fig:perception_arch}. This filter processes the measured $(s_{opp},d_{opp})$ position and $(v_{s,\,opp},v_{d,\,opp})$ velocity, ensuring a stable estimation robust to measurement noise. As it is known that during the \textit{Head-to-Head} phase of a race, only one opponent will be present, only a single \gls{kf} instance is initialized, meaning the velocity is only estimated for a single opponent. This could however be expanded for future multi-opponent racing scenarios.

This \gls{ekf} operates similarly to a constant velocity \gls{kf}, but it incorporates a normalization in the residual function to accommodate the cyclical nature of the \emph{Frenet} coordinates. The state vector \( \mathbf{x_{opp}} = [s_{opp}, v_{s,\,opp}, d_{opp}, v_{d,\,opp}]^T \) and the measurement vector \( \mathbf{z_{opp}} = [z_{s,\,opp}, z_{v_s,\,opp}, z_{d,\,opp}, z_{v_d,\,opp}]^T \) are defined with their respective components representing position and velocity along the track, lateral displacement, and lateral velocity.

The behavior of the \gls{kf} adapts depending on the visibility of the opponent car. 
When the opponent is within \gls{los}, the behavior of the opponent is assumed to keep constant velocity along the \emph{Frenet} $s$ dimension, and the control input \( \mathbf{u_{LoS}} = [0, - d_{opp}, - v_{d,\,opp}]^T \) is updated accordingly. 
On the other hand, when the opponent is not in \gls{los}, the control input in the prediction step of the \gls{ekf} is updated in order to drive the unseen opponent to the target velocity of the racing line \( v_{s,\text{target}} \), i.e. \( \mathbf{u_{nonLoS}} = [(v_{s, \text{target}} - v_{s,\,opp}), - d_{opp}, - v_{d,\,opp}]^T \). 
When considering the $d$-axis dynamics, both cases are instead treated the same way, with the state of the opponent being driven to zero both positionally in $s$ and for the velocity in $v_d$.
This prediction model ensures a more stable estimation in scenarios where direct measurements are not available, guaranteeing a more realistic behavior in unseen track sections.
% The behavior of the \gls{kf} adapts depending on the visibility of the opponent car. When the opponent is visible, the measured speed and velocity are directly used for estimation, ensuring responsive and accurate tracking. However, when the opponent is out of sight, such as behind a corner, the system switches to a racing-line-based model. This model estimates the opponent's position and velocity based on the system's own racing line, as detailed in \Cref{sec:planning}, and a target velocity parameter. This approach ensures a more stable estimation in scenarios where direct measurements are not available, preventing erratic behavior in the estimation.

The model is described by:
\begin{align}
    \mathbf{x}[k+1] &= \mathbf{F} \mathbf{x}[k] + \mathbf{B} \mathbf{u}[k] + \mathbf{w}_x[k], 
    \quad \mathbf{F} = \begin{bmatrix} 1 & \Delta t & 0 & 0 \\ 0 & 1 & 0 & 0 \\ 0 & 0 & 1 & \Delta t \\ 0 & 0 & 0 & 1 \end{bmatrix},
    \quad \mathbf{B} = \begin{bmatrix} 0 & 0 & 0 \\ P_{v_s} & 0 & 0 \\ 0 & P_d & 0 \\ 0 & 0 & P_{v_d} \end{bmatrix}, \nonumber \\
    \mathbf{z}[k] &= \mathbf{H} \mathbf{x}[k] + \mathbf{w}_z[k], \quad \mathbf{H} = \begin{bmatrix} \sigma^2_{s} & 0 & 0 & 0 \\ 0 & \sigma^2_{v_{s}} & 0 & 0 \\ 0 & 0 & \sigma^2_{d} & 0 \\ 0 & 0 & 0 & \sigma^2_{v_{d}} \end{bmatrix}, \quad \mathbf{u[k]}=
    \begin{cases}
      \mathbf{u_{LoS}[k]}, & \text{if $opp$ in \gls{los}} \\
      \mathbf{u_{nonLoS}[k]}, & \text{else}
    \end{cases},
\end{align}
where \( P_{v_s} \), \( P_d \), and \( P_{v_d} \) are the proportional gains for the respective control inputs to the states \( v_s \), \( d \), and \( v_d \). Furthermore, \( \Delta t \) is the time between two updates, \( \mathbf{w}_x \sim \mathcal{N} (0, \mathbf{Q})\) is the process Gaussian noise where \( \mathbf{Q} \in \mathbb{R}^{4 \times 4}\) is the covariance matrix, and \( \mathbf{w}_z \sim \mathcal{N} (0, \mathbf{R}) \) is the input Gaussian noise where \( \mathbf{R} \in \mathbb{R}^{4 \times 4} \) is the input covariance matrix. \rev{While it has been empirically evaluated that the proposed system is capable of operating with high racing performance with the parameters described in \Cref{sev:app_perception}, a further point to counteract the effects of sensor noise would be to adapt and tune the measurement covariance matrix \( \mathbf{H} \) and adjust the $\sigma^2_{i}$ to the observed measurement variance of $\mathbf{x_{opp}}$. In addition, data-driven methods to estimate the covariance matrix, such as \cite{AKESSON2008769_ALS_tuning}, could further improve the model, which currently considers every measurement as an independent variable.}

%%%%%%%%%%%%%%%%%%%%%%%%%%%%%%%%%%%%%%%%%%%%%%%%%%%%%%%%%%%%%%%%%%%%%%%%%%%%%%%%%%%%%%%%%%%%%%%%%%%%
\subsection{Opponent Estimation Results}
In the experimental setup designed to assess the performance of the perception module, the ego-agent and an opponent racecar were positioned on a racetrack, each with the \emph{ForzaETH Race Stack} deployed. The opponent vehicle was set to maintain the same racing line as the ego-agent but at a slower pace. This configuration enables the ego-agent to trail at close proximity (using the trailing controller later described in \Cref{ssubsec:trailing_ctrl}) while running the perception module. As a result, the detection and estimation of the ego-agent's perception module and the localization information of the opponent car can be directly recorded to evaluate the accuracy of the detector and tracker. \rev{Conducting these experiments in a physical setting demonstrates the reliability of the proposed perception algorithms, particularly their ability to handle sensor noise and real-world imperfections encountered when deployed in a race scenario.}

\subsubsection{Opponent Detection Accuracy}\label{subsec:od_acc}
\begin{figure}[ht]
    \centering
    \includegraphics[width=\columnwidth]{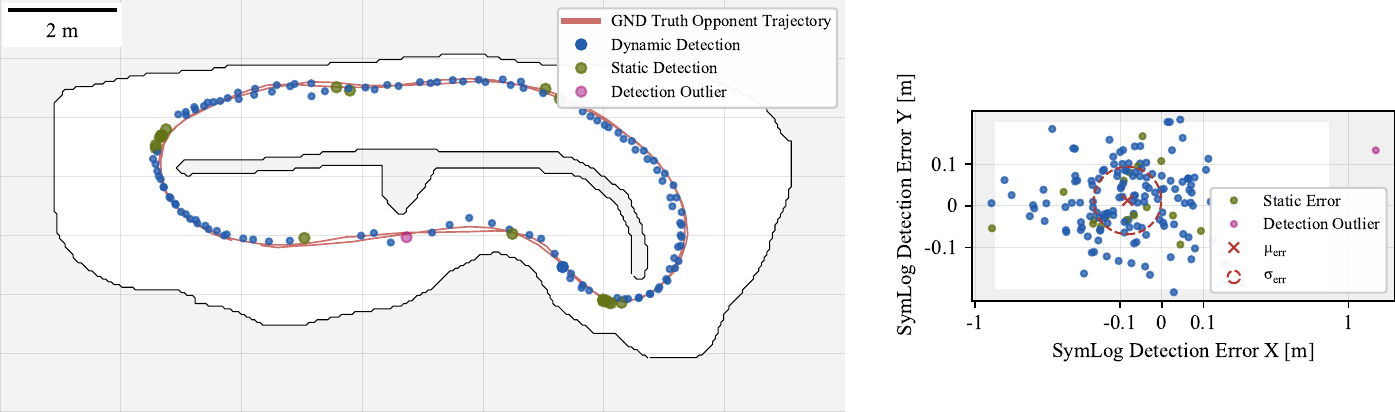}
    \caption{Evaluation of the spatio-temporal detection accuracy during multiple laps on a track, by trailing behind an opponent with the perception system active. Left: Spatial detections (blue), opponent trajectory (red), static detections (green), and detection outliers (purple). Right: Detection error in a symmetric linear-logarithmic plot; shaded regions indicate logarithmic scales and purple marks are outliers excluded from the \gls{rmse} calculation.}
    \label{fig:detection_res}
\end{figure}

\Cref{fig:detection_res} illustrates the performance of the detection submodule within the depicted racetrack environment. On the left side of the figure, spatial detections are mapped out across the racetrack, captured over several rounds of trailing behind a slower-moving autonomous opponent. The opponent's self-localization served as a reference for recording ground-truth positions over time, which are highlighted as red lines. The blue dots signify the classification of the detections, determined based on the mean and standard deviation of the buffer, as outlined in \Cref{fig:perception_arch}. The green dots indicate static detections; in this particular experiment, these instances are incorrectly classified since the opponent was consistently in motion. The purple dots, on the other hand, are labeled as detection anomalies, signifying that the spatial detections significantly deviated from the opponent's ground-truth position. For this experiment, a threshold of twice the measurement standard deviation, equivalent to \SI{0.17}{\metre}, was set to identify a detection outlier. 

It is worth mentioning, that even when the detection appears proximate to the opponent's ground-truth trajectory, the detection error over time-synchronized point-couples is taken into account and a longitudinal error is therefore present. On the right side of \Cref{fig:detection_res}, we present the quantitative analysis of the spatial and qualitative detections shown on the left. This plot illustrates the longitudinal and lateral detection error in meters with respect to the global coordinate frame. 
% To enhance visibility, the plot employs a symmetric linear-logarithmic scale, with the grey-shaded region representing the logarithmic domain. The red cross symbolizes the \gls{rmse} of the detections, calculated to be \SI{0.17}{\metre}, while the red dotted circle surrounding the mean represents the standard deviation of the \gls{rmse}, calculated to be \SI{0.08}{\metre}. 
To enhance visibility, the plot employs a symmetric linear-logarithmic scale, with the grey-shaded region representing the logarithmic domain. The red cross symbolizes respectively the average $X$ and $Y$ detection error $\mu_{err}=(\SI{-0.08}{\metre},\,\SI{0.01}{\metre})$, while the red dotted circle surrounding the mean represents the standard deviation $\sigma_{err}=\SI{0.08}{\metre}$. The resulting \gls{rmse} is calculated to be \SI{0.17}{\metre}.
% From this graphical representation, it is noticeable that the detections in this experiment exhibit a bias towards the left side of the ego-racecar. This bias is presumably attributed to the predominant right-turning nature of the racetrack, influencing the positional accuracy of the detections.

In this evaluation, the performance of the perception module's detection submodule was quantitatively assessed using the \gls{tpr} and the \gls{fdr}. The \gls{tpr}, calculated as \( \frac{TP}{TP + FN} \), where \( TP \) is the number of \gls{tp} and \( FN \) is the number of \gls{fn}, reflects the submodule's accuracy in correctly detecting the opponent car. The \gls{tpr} for the system was found to be 96.8\% out of 198 detections. On the other hand, the \gls{fdr}, calculated as \( \frac{FP}{TP + FP} \), where \( FP \) is the number of \gls{fp}, provides insight into the proportion of false alarms among all the positive detections made by the system. The computed \gls{fdr} for our system was 1.6\% out of 198 detections. Here, \gls{tp} are correct detections within \(2 \times \sigma_{RMSE}\) of the ground-truth, resulting in 182 \gls{tp} detections, while \gls{fp} are incorrect detections beyond this range, resulting in 6 \gls{fp} detections, of the total 198 detections. Lastly, \gls{fn} are missed detections within this range. \gls{tn} are not applicable in this context, as it would require defining true negative events, which are not present in this detection task.

\subsubsection{Opponent Estimation Accuracy}

\begin{figure}[ht]
    \centering
    \includegraphics[width=\columnwidth]{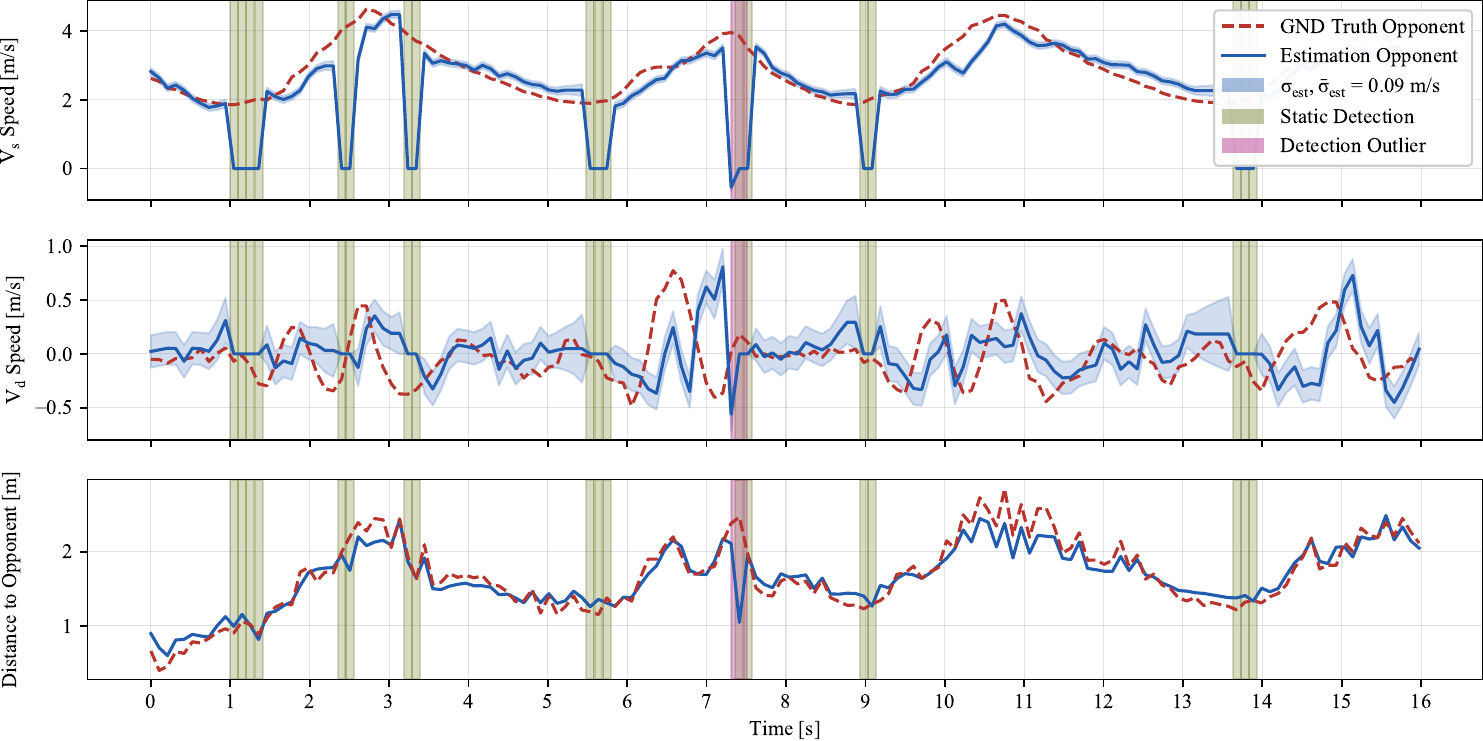}
    \caption{Opponent velocity estimation associated to the detections in \Cref{fig:detection_res}. Ground-truth measurements of the opponent states are depicted in red, while the velocity estimation of the opponent is highlighted in blue. Green-shaded regions are static classifications of the opponent. Purple-shaded regions depict detection outliers, i.e. exceeding an error of twice the measurement standard deviation.}
    \label{fig:tracker_res}
\end{figure}

\Cref{fig:tracker_res} highlights the performance of the velocity estimation capabilities, from the same experimental settings as the detection evaluations in \Cref{fig:detection_res}. Red depicts the ground-truth velocities in \((s, d)\) in \SI{}{\metre\per\second}, blue is the velocity estimation with the shaded blue estimation covariances $\bar\sigma_{est}$ both from the \gls{ekf} tracking submodule. Green-shaded represent the static misclassifications and purple-shaded the detection outliers.

The tracking submodule's estimation accuracy was evaluated through the computation of the \gls{rmse} for the longitudinal $v_{s, opp}$ and lateral $v_{d, opp}$ velocities. For non-static detections, the submodule exhibited a \gls{rmse} of \SI{0.49}{\metre\per\second} for \(v_{s, opp}\) and \SI{0.37}{\metre\per\second} for \(v_{d, opp}\), indicating a reliable accuracy in velocity estimation when the opponent car is in motion. However, when considering all detections, including static ones, the \gls{rmse} for \(v_{s, opp}\) increased significantly to \SI{1.00}{\metre\per\second}, while the \gls{rmse} for \(v_{d, opp}\) remained relatively stable at \SI{0.35}{\metre\per\second}. From this, it is visible, that the static misclassification significantly affects the longitudinal velocity estimation performance. Yet, it is to be mentioned, that during a race scenario within the context of \emph{F1TENTH}, static obstacles within a race can be neglected, thus allowing the tracker submodule to focus only on dynamic obstacles, which enables high performance.

\rev{Static misclassifications occur due to \gls{ids} where the tracker switches the identity of the object of interest briefly from a static to a dynamic identity. To mitigate this, future implementations could explore strategies such as dynamic thresholding or adaptive filtering and even consider the integration of \gls{sota} \gls{ml}-based perception methodologies used in large datasets like \emph{nuScenes} \cite{caesar2020nuscenes}. Such approaches involve \gls{kf}-tracklets to mitigate the \gls{ids}, as demonstrated in \cite{cr3dt, qd_track}. However, our focus remains on using \gls{cots} hardware suitable for embedded real-time applications, which typically is not the main focus of the aforementioned large scale \gls{ml} approaches.}

\rev{Despite the inherent challenges of direct comparisons due to differences in scale, sensor modalities, complexity difference of the perception objective, and the high computational requirements of \gls{sota} methods on \emph{nuScenes}, our system achieves a velocity estimation \gls{rmse} of approximately \SI{0.49}{\metre\per\second}. This performance is comparable to the \gls{mave} of \SI{0.47}{\metre\per\second} reported by \cite{cr3dt} on the \emph{nuScenes} validation set. Future research could incorporate full-scale \gls{ml}-based \gls{sota} detection methods like those found in \cite{centerpoint} and enhance multi-obstacle tracking capabilities using approaches from \cite{qd_track, cr3dt}, potentially improving performance and robustness, particularly in scenarios with multiple opponents. Such advancements could significantly elevate the accuracy and scalability of perception systems in autonomous racing, pushing the field toward more effective multi-opponent handling.}

\FloatBarrier
%%%%%%%%%%%%%%%%%%%%%%%%%%%%%%%%%%%%%%%%%%%%%%%%%%%%%%%%%%%%%%%%%%%%%%%%%%%%%%%%%%%%%%%%%%%%%%%%%%%%
\section{Planning} \label{sec:planning}
Planning is a core robotic task within the \emph{See-Think-Act} cycle \cite{amr}. The planning module leverages environmental data from the \emph{Perception} module and ego-information from the \emph{State Estimation} module to plan efficient and effective trajectories, adapting dynamically to environmental changes. These planned trajectories will then be supplied to the downstream \emph{Control} task for execution. 

For \emph{F1TENTH}, this translates into two primary objectives: \rev{\emph{Global Planning}, described in \Cref{subsec:gb_optimizer}, to obtain} an optimal racing line, \rev{that is} computed offline. \rev{Secondly, \emph{Local Planning}, further elaborated in \Cref{sec:loc_planning}, where the racecar reacts in real-time} to opponents during \emph{Head-to-Head} races. \rev{Whether to follow the optimal racing line or to take an evasive trajectory is decided by a state machine, as further described in \Cref{subsec:state-machine}}, to dynamically prioritize planning and decision-making between safety and performance.\revdel{This involves strategic planning for collision avoidance and executing overtaking maneuvers, ensuring both competitiveness and safety on the track.}

\subsection{Architecture}

\begin{figure}[ht]
    \centering
    \includegraphics[width=\columnwidth]{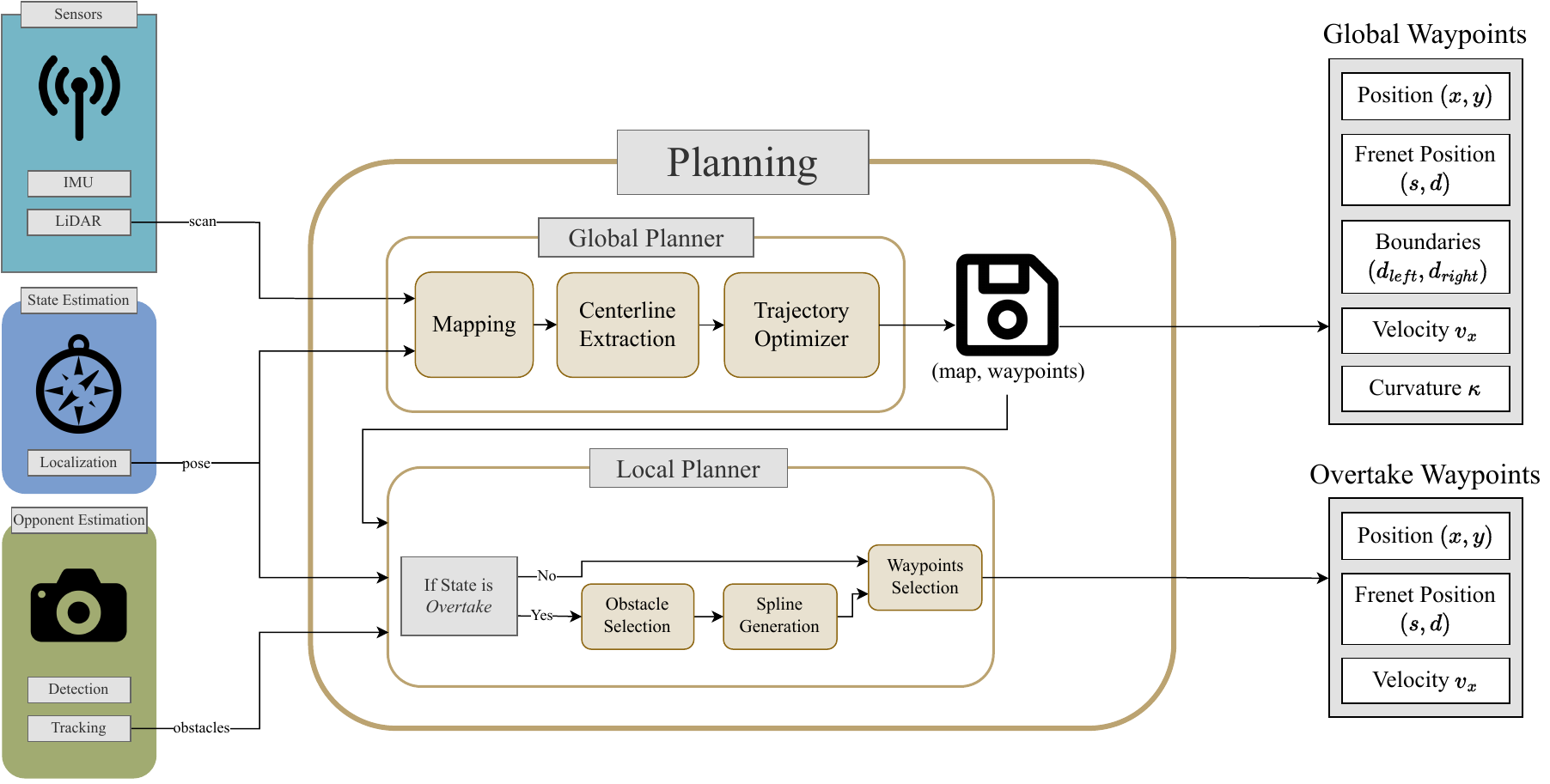}
    \caption{Overview of the proposed planning module. The system integrates a global planning submodule for computing an optimal racing line offline, alongside a local planner that dynamically responds to opponent estimations to compute overtaking trajectories.}
    \label{fig:planner_arch}
\end{figure}

\Cref{fig:planner_arch} depicts the architecture of the \emph{Planning} module within the proposed \emph{ForzaETH Race Stack}. The \emph{Planning} module is subdivided into two submodules: a global and a local planner. The global planner leverages the occupancy grid, generated during the free practice session, and yields a performant racing line for the given track. The global planner operates offline, solely attributed to the calculation of a global trajectory that is stored for later use in the race. On the other hand, the local planner employs the global trajectory as a reference racing line and, based on it, generates a local trajectory designed to avoid obstacles, and thus allows the car to overtake an opponent.

%%%%%%%%%%%%%%%%%%%%%%%%%%%%%%%%%%%%%%%%%%%%%%%%%%%%%%%%%%%%%%%%%%%%%%%%%%%%%%%%%%%%%%%%%%%%%%%%%%%%
\subsection{Global Planner} \label{subsec:gb_optimizer}
% \assign{Luca Schwarzenbach} here you talk briefly about how you extract the centerline from the map.
% \assign{Luca Schwarzenbach} here you talk about the iterative minimum curvature planner. Also talk about the params used in the tool, like in your report. Focus only on the most important ones.

The global planner is based on the work presented in \cite{Heilmeier2020MinimumCar}. Their work describes the planning of a minimum curvature trajectory using a quadratic optimization problem formulation. To optimize a global path around a racetrack, it is necessary to acquire the centerline and the corresponding track boundaries from the map.

\begin{figure}[ht]
    \centering
    \subfloat[Occupancy Grid \label{subfig:occ_grid}]{
        \includegraphics[width=0.49\textwidth]{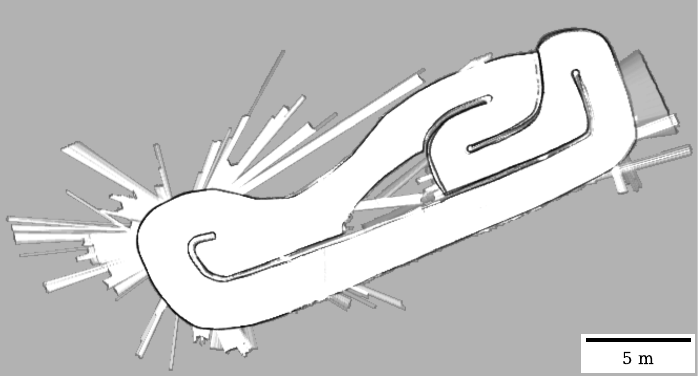}}
    \hfill
    \subfloat[Centerline and boundaries \label{subfig:centerline}]{
            \includegraphics[width=0.49\textwidth]{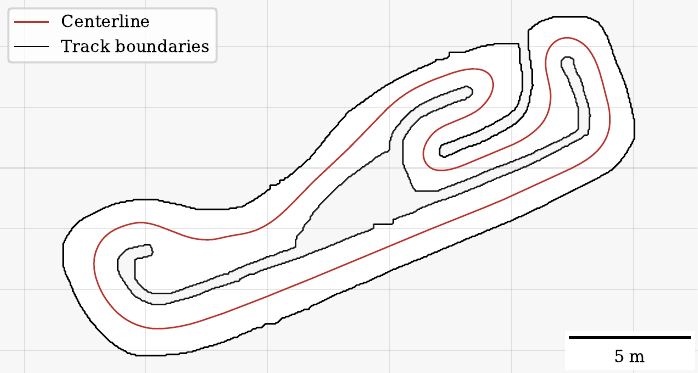}}
    \hfill
    \caption{The centerline extraction step of the global planner shown on the track of the \emph{ICRA Grand-Prix 2022}. \Cref{subfig:occ_grid}: Occupancy grid obtained by \gls{slam}. \Cref{subfig:centerline}: Extracted centerline (red) including track boundaries (black).}
    \label{fig:cent_extraction}
\end{figure}

The occupancy grid, generated by \gls{slam}, can be interpreted as an image by transforming each cell into a pixel as illustrated in \Cref{subfig:occ_grid}. In the first step, the occupancy grid is binarized. Subsequently, the binarized image is smoothed with a morphological open filter \cite{Bovik2009BasicProcessing} to mitigate most of the \gls{lidar} scans located outside of the track. Following this, the centerline is extracted from the filtered image using the morphological skeleton method \cite{Kong1996TopologicalProcessing}. The centerline, characterized by its angular shape, can pose challenges for path optimization due to abrupt directional changes. Therefore, the centerline is smoothed using a Savitzky-Golay filter \cite{Orfanidis1995IntroductionProcessing}, and the resulting centerline is depicted in \Cref{subfig:centerline}. Finally, the centerline, in combination with the \emph{Watershed} algorithm \cite{bertrand:hal-00622398} is utilized to derive the racetrack and, consequently, the distances along the normal vector to the track boundaries, crucial information that, being in \emph{Frenet} frame, allows us to efficiently compute the distance of a specific coordinate to the boundary. This is useful, for example, to evaluate if candidate trajectories are safely within the track boundaries or if detected obstacles are inside the racetrack. 

\begin{figure}[ht]
    \centering
    \subfloat[Sectors along the racetrack \label{fig:sectors_track}]{\includegraphics[width=0.49\textwidth]{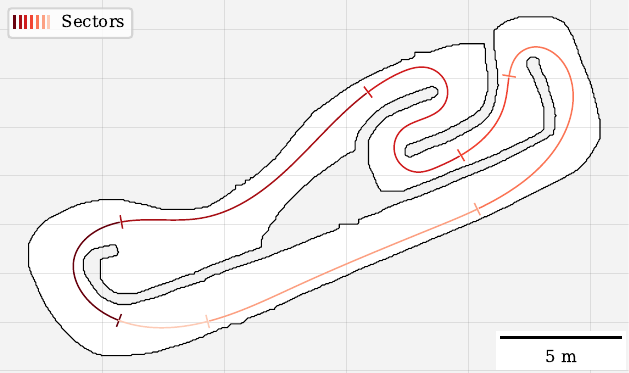}}
    \hfill
    \subfloat[Scaled velocity profile \label{fig:sector_vels}]{\includegraphics[width=0.49\textwidth]{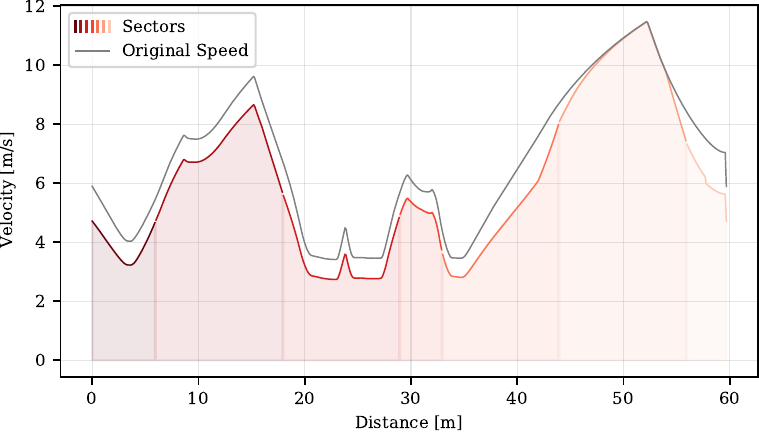}}
    \hfill
    \caption{Example \emph{sectors} defined on the \emph{Global Waypoints} for the \emph{ICRA Grand-Prix 2022} racetrack. The \emph{sectors} are shown with increasingly lighter hues in \Cref{fig:sectors_track}, with a segment indicating the beginning of each \emph{sector}. The effect of such \emph{sectors} on the racing line's speed profile is then displayed in \Cref{fig:sector_vels}, with the original velocity profile in gray. The 7 chosen \emph{sector scalers}, in this example, have as values \texttt{[0.8, 0.9, 0.8, 0.9, 0.8, 1.0, 0.8]}, corresponding to the \emph{sectors} ordered from the darkest to the lightest hue.}
    \label{fig:sectors}
\end{figure}

For the computation of the global trajectory, we employ the global trajectory optimization tool presented in \cite{Heilmeier2020MinimumCar}. The centerline points $[x_i, y_i]$ and their corresponding distances to the track boundaries $w_{tr,left,i}, w_{tr,right,i}$ serve as the input. The optimization method used is the iterative minimum curvature optimization. This approach iteratively applies the standard minimum curvature optimization to address linearization errors occurring in the conventional problem formulation. The objective of the minimum curvature optimization is to minimize the sum of the discrete squared curvature $\kappa_i^2$ along the racing line with $N$ points, which are interpolated through splines. Taking into account the vehicle width $w_{ego}$, the optimization problem can be formulated as
\begin{equation} \label{eq:opt1}
\begin{aligned}
   & \underset{[\alpha_1 \ldots \alpha_N]}{\text{minimize}} \quad & & \sum_{i=1}^{N} \kappa_i^2 = \sum_{i=1}^{N} \frac{x_i'^2 y_i''^2 - 2 x_i' x_i'' y_i' y_i'' + y_i'^2 x_i''^2}{(x_i'^2 + y_i'^2)^3} \\
   & \text{subject to} \quad & & \alpha_i \in [-w_{tr,left,i} + \frac{w_{ego}}{2}, w_{tr,right,i} - \frac{w_{ego}}{2}] \\
   &&& \forall \ 1 \leq i \leq N.\\
\end{aligned}
\end{equation}
Given the cubic spline formulation of the points with respect to the parameter $t$ as follows:
\begin{equation}
    x_i=a_i+b_it+c_it^2+d_it^3,
\end{equation}
the terms $x_i',\,x_i''$ are defined as $x_i'=\frac{dx_i}{dt},\,x_i''=\frac{dx_i'}{dt}$ (and similarly for the $y_i$ terms). For more information on how the problem is setup and solved as a \gls{qp}, refer to \cite{Heilmeier2020MinimumCar}.
In addition to path optimization, this tool also generates a velocity profile using a forward-backward solver, resulting in a global trajectory consisting of \emph{Global Waypoints} as illustrated in \Cref{fig:sectors}. 
The optimization tool offers a range of configurable parameters, such as vehicle specifications or optimization-specific attributes. The most relevant parameters are reported in \Cref{sec:app_planning}\rev{, such as the car width or the maximum curvature allowed, derived from the car's limited steering angle}.

\rev{The optimization objective in \cref{eq:opt1} only considers the geometrical track layout, assuming constant surface grip. However, in practice, friction limits vary, significantly impacting the car's performance, and limiting the achievable lap time.} To improve the tunability of the obtained racing line and handle mixed conditions such as floors with different friction levels, multiple \emph{sectors} along its length are sequentially defined, based on track-by-track heuristics. Every \(i\)-th  \emph{sector} is then linked to a scaling parameter \(\sigma^{sector}_{i}\in[0,1]\), which then multiplies the speed of the global racing line, in order to scale it down. To ensure smooth transitions between the \emph{sectors}, linear interpolation is applied for a range of \SI{1}{\metre} at the junction of the \emph{sectors}. \rev{By default, all \textit{sectors} are set to \(\sigma^{sector}_{i}=0.5\), as it has been empirically determined that these initial parameters result in very safe driving behavior.}
An example of such \emph{sectors} can be seen in \Cref{fig:sectors}. The \emph{sector scalers} $\sigma_i^{sector}$ can then be tuned, either by hand or automatically by means of \gls{bo}. 
% \gls{bo} is a technique that can be used to estimate and find the optimum of a function \(f\) when only realizations \(f(x)\) of such function are measured.
% The \gls{bo} setup is particularly fit for settings where the function is difficult to evaluate, such as the autonomous racing case, where an entire lap around the track is needed to evaluate the lap time. 
% To ease this critical point, specific acquisition functions can be used to find the selected evaluation points for the cost function in order to optimize the \emph{exploration-exploitation} tradeoff. 
% In other words, the algorithm optimally chooses the parameter settings to use for driving the car for a predefined number of laps.
In this case, the technique is used to minimize a cost function which is a linear combination of lap time, racing line deviation, and minimum distance from the boundaries. The goal of the \gls{bo} setup is therefore to find a mapping from the \emph{sector scalers} to this cost function. 
The implementation is carried out with the \emph{BayesOpt4ROS} package from \cite{Froehlich2021BayesOpt4ROS}, and the specific acquisition function used is \gls{ei} \cite{Jones1998EfficientGO}, which was selected over the other implemented option \gls{ucb} \cite{ucb2002} after empirical assessments deemed it more sample efficient for our task, achieving faster performance when given the same amount of sample laps. \rev{Overall, it has been observed that the usage of either hand-tuned or \gls{bo} obtained \textit{sector} scalers in a race-setting, significantly boosts lap-time and tracking performance, as demonstrated in \Cref{chap:res_tt}, while allowing for simple and flexible robot-operator interaction.}

\subsection{State Machine} \label{subsec:state-machine}

\begin{figure}[ht]
    \centering
    \includegraphics[width=\columnwidth, trim={0 0 0 0}, clip]{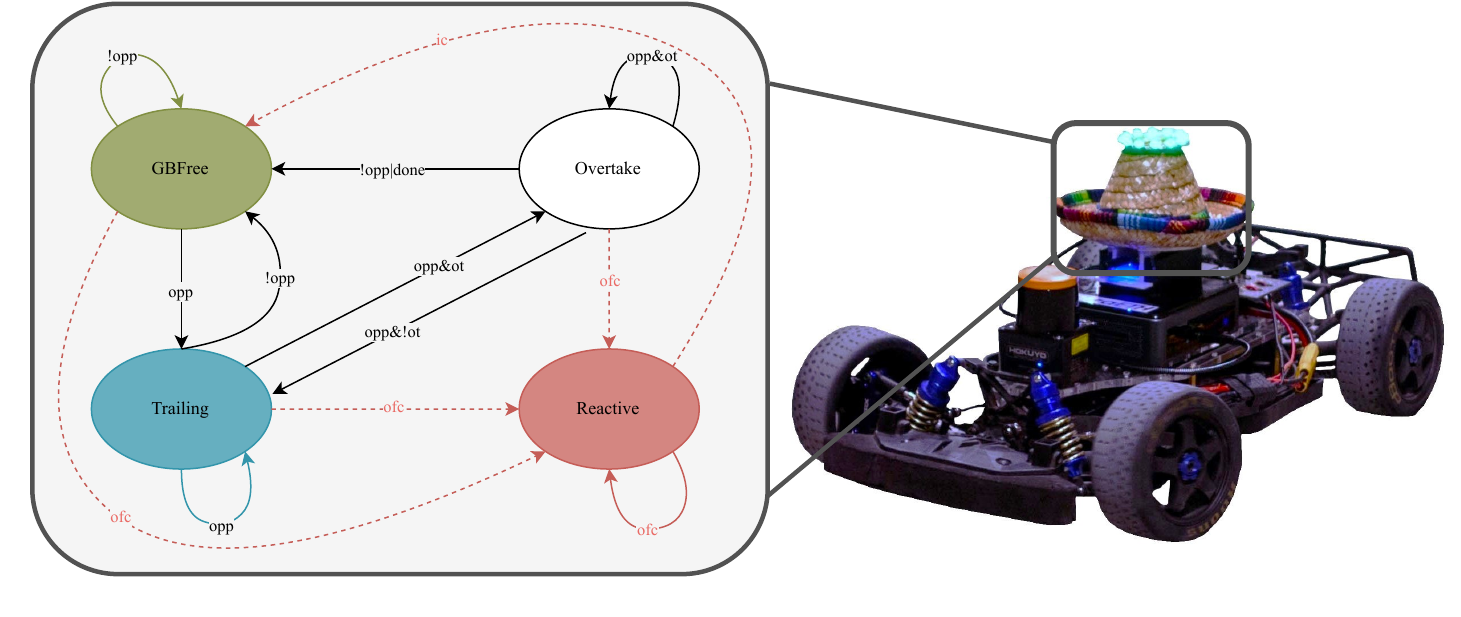}
    \caption{Visualization of the implemented state-machine with the corresponding state-transitions. The vehicle's \gls{led} indicators correspond to these states, providing an immediate visual cue of the real-time status. The transition between states is determined by a set of boolean conditions: \texttt{opp} denotes the necessity to account for an opponent; \texttt{ot} signifies the feasibility of an overtake; \texttt{done} confirms the completion of an overtake; \texttt{ofc} indicates out-of-control, and \texttt{ic} signals the vehicle is in-control. For ease of presentation, the figure presented here is a simplified version; the full state machine is available in the open-source codebase. }
    \label{fig:state_machine}
\end{figure}

\Cref{fig:state_machine} illustrates the devised state machine, with each state being physically represented on the vehicle through corresponding \gls{led} colors, thereby augmenting real-time visual feedback.
This state machine is responsible for orchestrating different high-level behaviors through the information obtained from the different autonomy modules and \rev{supplying} the \emph{Control} module with the correct waypoints \rev{obtained either from the static global planner or the dynamically updated local planner. To balance and prioritize the agent's behavior in this dynamic environment in real time, different sets of state transitions are employed.} A detailed explanation of the states, with conditions for switching formatted as \texttt{<cond>}, is provided below:

\begin{enumerate}[I]
    \item \textbf{GBFree:} This state denotes an unobstructed global racing line, facilitating unimpeded trajectory tracking. \emph{GBFree} denotes the Global racing line being free. 
    \item \textbf{Trailing:} This state denotes when an opponent, denoted as \texttt{opp}, is proximal to the forthcoming racing line, and the longitudinal controller is activated to keep a constant gap, ensuring safe and close trailing behind the opponent and positioning the robot well for potential overtaking scenarios.
    \item \textbf{Overtake:} This state is engaged when the robot, after trailing an opponent, is presented with a valid overtaking solution (\texttt{ot}) by the local planner. The robot adheres to the proposed overtaking waypoints unless the solution becomes invalid, in which case it reverts to the \emph{Trailing} state.
    \item \textbf{Reactive:} Serving as a safety net, this state is activated when poor state estimation jeopardizes safe operation (\texttt{ofc}), prompting the robot to employ a reactive scheme via the \gls{ftg} controller. The state can be exited as soon as control has been regained, i.e. is in control (\texttt{ic}).
\end{enumerate}

%%%%%%%%%%%%%%%%%%%%%%%%%%%%%%%%%%%%%%%%%%%%%%%%%%%%%%%%%%%%%%%%%%%%%%%%%%%%%%%%%%%%%%%%%%%%%%%%%%%%
\subsection{Local Planner} \label{sec:loc_planning}
As \emph{F1TENTH} features unrestricted \emph{Head-to-Head} competitions, dynamically avoiding obstacles is crucial, and local planning constitutes a fundamental step in this endeavor, by generating a feasible and performant trajectory in order for the controller to follow it. 
Aiming for simplicity and computational efficiency, the core algorithm revolves around generating a spline around the opponent that reconnects with the racing line.  

When no obstacle is within a predefined distance from the nominal trajectory, the local planner provides the controller described in \Cref{chap:control} with the global waypoints. This state is the one described as \emph{GBFree} in \Cref{subsec:state-machine}. When instead the detection of an opponent brings the car to the \emph{Overtake} state, a local trajectory is needed and a detection of a target opponent is assumed to be available in the space in front of the agent. As discussed in \Cref{subsec:tracking}, the provided opponent state is formulated in the \emph{Frenet} frame as \( \mathbf{x_{opp}} = [s_{opp}, v_{s,\,opp}, d_{opp}, v_{d,\,opp}]^T \). 

\begin{figure}[ht]
    \centering
    \includegraphics[height=2in, width=\columnwidth]{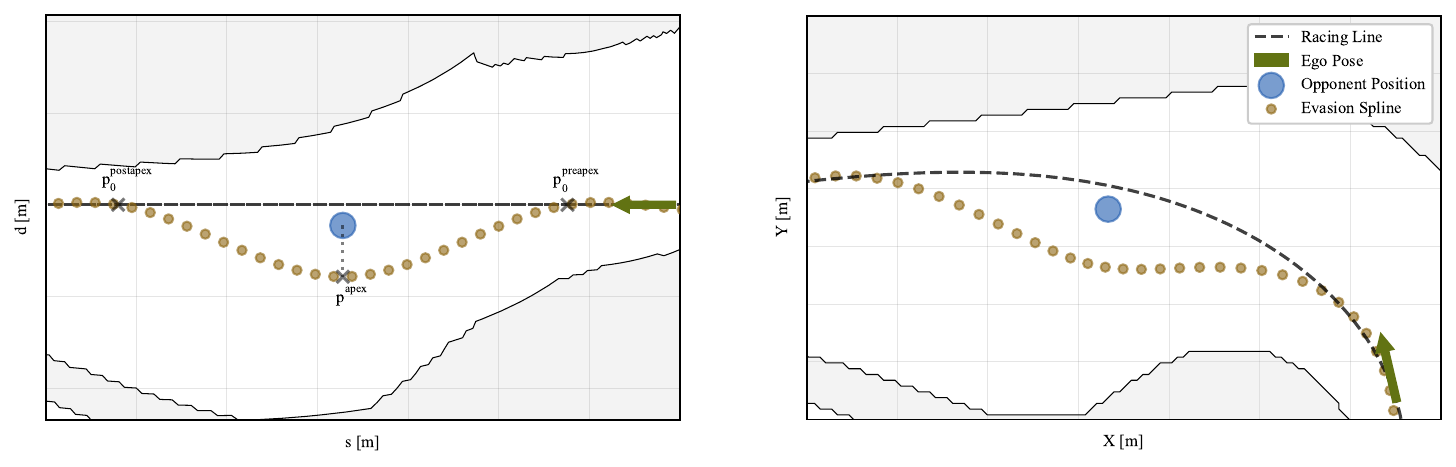}
    \caption{On the left, the planning setup in the \emph{Frenet} coordinate system is described. The ego car's pose is identified with a green arrow, the opponent's position is identified with a blue dot, and the final avoidance trajectory around the opponent is in bronze. The points used to construct the spline are marked with black crosses. For ease of visualization, only three example points are used in the figure: the first points before and after the apex are, respectively, $p^{preapex}_0,\,p^{postapex}_0$, while the point at the apex is $p^{apex}$. The safety distance used to find the lateral apex distance from the opponent's position is shown with a dotted gray line. On the right, the resulting overtaking trajectory is displayed in \emph{Cartesian} coordinates.}
    \label{fig:spliner}
\end{figure}

A set of points $\mathbf{p}^{spline}$ is chosen around the opponent and a third-order spline is then fit through these points, to eventually resample at the same discretization step of the global waypoints. An example setup, describing the set of points and the final spline can be seen in \Cref{fig:spliner}.
The first point obtained for $\mathbf{p}^{spline}$ is the \emph{overtaking apex}, which is the point closest to our opponent where we plan to overtake. This point is defined as \(p^{apex} = [s_{opp},\,d^{apex}]^\top\), where \(s_{opp}\) is the same as the target opponent and \(d^{apex}\) is defined to be either on the left or the right of the obstacle.
The side is decided by evaluating the space from the opponent's position to the track boundaries on the left and on the right. If one of the two sides is too small, the other one is selected for overtaking. If both have enough space, the side that yields an overtaking racing line closer to the global trajectory is chosen. 
The overtaking \(d^{apex}\) is then obtained in the following way:
\begin{align}
    \text{If the overtake is to the left: } d^{apex} &= d_{opp} + \left( \frac{1}{2}(w_{opp} + w_{ego}) + \delta^{apex} \right), \\
    \text{if the overtake is to the right: } d^{apex} &= d_{opp} - \left( \frac{1}{2}(w_{opp} + w_{ego}) + \delta^{apex} \right),
\end{align}

where \(w_{ego},\,w_{opp}\) indicate respectively the width of the ego-car and of the detected opponent, and \(\delta^{apex}\) indicates a tunable distance at which we want to overtake, to account for robustness.
Such a safety distance can be set based on the sum of our detection system's positional \gls{rmse} plus the measured average lateral distance of our car from the racing line. Practically, this sum only constitutes a lower bound, and usually a value at least twice this bound is chosen\rev{: generally this remains a crucial tuning parameter that can be tuned to account for unexpected errors from the detection system (e.g. in case of a faulty sensor), or in case of a particularly erratic opponent}.
To obtain the remaining points of $\mathbf{p}^{spline}$, \(n^{spline}\) points before and after the \emph{overtaking apex} are selected on the racing line (they are grouped in the two vectors \(\mathbf{p}^{{preapex}},\,\mathbf{p}^{{postapex}}\;\in\;\mathbb{R}^{2 \times n^{spline}}\)).
These points have all \(d = 0\) and are used to fit the spline in a way that reconnects to the racing line. The distances along the \(\textbf{s}\) dimension are then scaled with respect to the ego-car velocity to account for the decreased steering capacity with the increasing longitudinal velocity. The coordinates of $\mathbf{p}^{spline}$ are then defined as follows: 
\begin{align}
    \mathbf{p}^{{preapex}} &= \left[ 
    \begin{bmatrix}
    s_{opp} - \alpha_{v}\Delta_{0}^{{preapex}} \\ 0    
    \end{bmatrix},\,
    \begin{bmatrix}
    s_{opp} - \alpha_{v}\Delta_{1}^{{preapex}} \\ 0    
    \end{bmatrix},\,\hdots,\,
    \begin{bmatrix}
    s_{opp} - \alpha_{v}\Delta_{n^{spline}}^{{preapex}} \\ 0    
    \end{bmatrix}\right],
    \\
    p^{{apex}} &= \begin{bmatrix}
    s_{opp} \\ d^{{apex}}    
    \end{bmatrix},
    \\
    \mathbf{p}^{{postapex}} &= \left[ 
    \begin{bmatrix}
    s_{opp} + \alpha_{v}\Delta_{0}^{{postapex}} \\ 0    
    \end{bmatrix},\,
    \begin{bmatrix}
    s_{opp} + \alpha_{v}\Delta_{1}^{{postapex}} \\ 0    
    \end{bmatrix},\,\hdots,\,
    \begin{bmatrix}
    s_{opp} + \alpha_{v}\Delta_{n^{spline}}^{{postapex}} \\ 0    
    \end{bmatrix}\right],
    \\
    \mathbf{p}^{spline} &= \left[ p^{{preapex}}_{n^{spline}},\, \hdots,\, p^{{preapex}}_0,\, p^{{apex}},\, p^{{postapex}}_0,\, \hdots,\, p^{{postapex}}_{n^{spline}} \right],
\end{align}
where \(\alpha_{v}\) is the velocity scaler, scaling the length of the overtaking sector according to the current velocity, and \(\Delta^{\{{preapex},\,{postapex}\}}_i\) are the distances at which the spline points on the racing line are taken, defined as follows:
\begin{align*}
     \alpha_{v} &= 1 + \min\left(\frac{v_{s,\,ego}}{v_{max}}, 0.5\right), \\
     \bm{\Delta}^{\{{preapex},\,{postapex}\}} &= [\Delta^{\{{preapex},\,{postapex}\}}_1,\,\Delta^{\{{preapex},\,{postapex}\}}_2,\,\hdots,\,\Delta^{\{{preapex},\,{postapex}\}}_{n^{spline}}], \\
\Delta^{\{{preapex},\,{postapex}\}}_i &\in\mathbb{R}, \quad i\leq j \Rightarrow \Delta^{\{{preapex},\,{postapex}\}}_i \leq \Delta^{\{{preapex},\,{postapex}\}}_{j} \quad \forall i,\,j \in \{1,\,2,\,\hdots,\,n^{spline}\},
\end{align*}
where \(v_{s,\,ego}\) is the component of our velocity tangential to the racing line, and \(v_{max}\) is the maximum racing line velocity. 
% The \(\mathbf{\Delta}^{\{\text{preapex},\,\text{postapex}\}} = [\Delta^{\{\text{preapex},\,\text{postapex}\}}_1,\,\Delta^{\{\text{preapex},\,\text{postapex}\}}_2,\,\hdots,\,\Delta^{\{\text{preapex},\,\text{postapex}\}}_{n
% _{spline}}]\) with \(\Delta^{\{\text{preapex},\,\text{postapex}\}}_i \in\mathbb{R}\) are predefined distances with the only property of being ordered, i.e. \( i\leq j \Rightarrow \Delta^{\{\text{preapex},\,\text{postapex}\}}_i \leq \Delta^{\{\text{preapex},\,\text{postapex}\}}_{j} \quad \forall i,\,j \in \{1,\,2,\,\hdots,\,n^{spline}\}\).
Given this set of points, a spline is then fit in order to regress the \(\textbf{d}\) coordinate from the \(\textbf{s}\) coordinate. 
An evasion trajectory is then resampled from the obtained spline with the same discretization step used for the global racing line, and its speed profile is also applied to the evasion trajectory.
The avoidance trajectory is then substituted into the global racing line for \(s \in [s_{opp} - \alpha_{v}\Delta_{n^{spline}}^{{preapex}},\,s_{opp} + \alpha_{v}\Delta_{n^{spline}}^{{postapex}}]\) to provide waypoints all around the track. \rev{Lastly, a spatial threshold is defined to verify that the lateral component of all points on the computed trajectory, denoted as \(\mathbf{d}^{spline}\), remains within the track limits. Specifically, the condition \(\max(|\mathbf{d}^{spline}|) \leq d^{track}\) must be satisfied, where \(d^{track}\) represents the half-width of the racing track minus a safety margin parameter. This ensures that the entire evasion trajectory is safely contained within the boundaries of the racing track. If any part of the trajectory violates this constraint, the spline is deemed invalid and will not be considered.}
A diagram of the obtained trajectory is represented in both \emph{Frenet} and \emph{Cartesian} coordinates in \Cref{fig:spliner}.
% In this case, the numerical values were set to be \(n^{spline}=3,\,\bm{\Delta}^{\text{preapex}}=(2,\,3,\,4),\,\bm{\Delta}^{\text{postapex}}=(4.5,\,5,\,5.5)\).

%%%%%%%%%%%%%%%%%%%%%%%%%%%%%%%%%%%%%%%%%%%%%%%%%%%%%%%%%%%%%%%%%%%%%%%%%%%%%%%%%%%%%%%%%%%%%%%%%%%%
\subsection{Planning Results}

This section presents the results obtained from the implementation of our planning strategies, focusing on both global and local planning aspects. Initially, we present the results of our global planning strategy, displaying the obtained trajectory on different racetracks. This is followed by an examination of the local planning results, showcasing the local planner's ability to perform overtaking maneuvers. 

\subsubsection{Global Planning Results}

\begin{figure}[ht]
    \centering
    \subfloat[\emph{ICRA Grand-Prix 2022} track \label{fig:gb_icra}]{\includegraphics[width=0.49\textwidth]{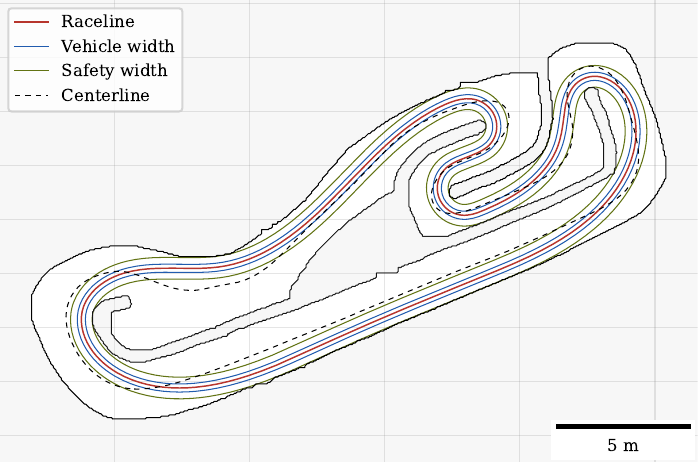}}
    \hfill
    \subfloat[Test track\label{fig:gb_rb}]{\includegraphics[width=0.49\textwidth]{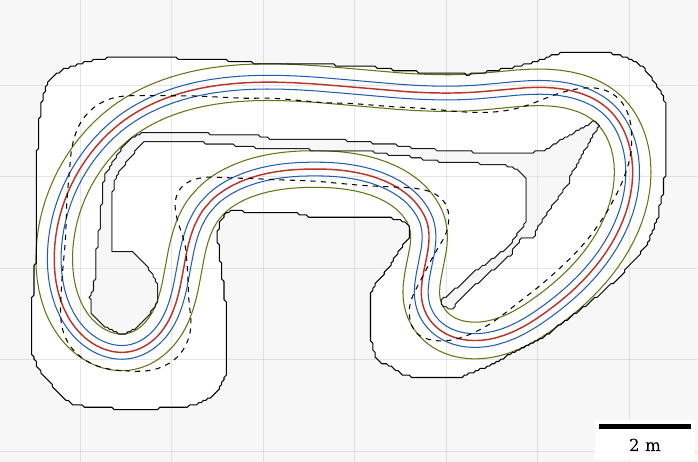}}
    \hfill
    \caption{The global trajectory optimized with iterative minimum curvature on two different tracks. Besides the racing line (red), the figure also shows the centerline (dashed black) and the vehicle width without (blue) and with a safety distance (green). The illustration shows that even with the conservative safety distance the selected optimizer finds a feasible trajectory for typical \emph{F1TENTH} competition tracks.}
    \label{fig:gb_trajectory}
\end{figure}

To showcase different trajectories from the global planner, we conducted qualitative assessments on various racetracks, as depicted in \Cref{fig:gb_trajectory}. \Cref{fig:gb_icra} illustrates the racetrack from the \emph{ICRA Grand-Prix 2022}, characterized by a combination of long straights and narrow turns. Conversely, \Cref{fig:gb_rb} presents a test track, notable for its narrowness and frequent sharp turns with minimal straight segments. Across both tracks, the planner consistently generated a global trajectory that adhered to predefined constraints, such as vehicle width including a safety margin, thereby ensuring the vehicle remained within the track bounds without corner-cutting.
As it can be seen from \Cref{fig:gb_trajectory}, the safety width is significantly larger than the vehicle width (\SI{0.3}{\metre} for the vehicle width and \SI{0.8}{\metre} for the safety width). This \rev{is needed mainly} for two reasons: \rev{firstly}, to account for the additional size needed by a car in case it is not perfectly aligned with the racing line, and \rev{secondly} to account for the limited tracking capabilities of the controller, as later described in \Cref{chap:control}. \rev{Hence, the imperfections of downstream autonomy modules can lead to a deviation of the racing line --- especially in sharp hairpin turns, which can be accounted for by increasing the safety width towards the boundaries.}

% \assign{Luca Schwarzenbach} show some cute trajectories on the actual maps, nothing more, only qualitative results.

\rev{Empirical evaluations demonstrate that the proposed global planning strategy effectively scales and generalizes across diverse racetrack layouts and dimensions, as illustrated in \Cref{fig:gb_trajectory}, \Cref{fig:map-vs-pp}, and \Cref{fig:winti_run}. Environmental variations, particularly in traction, are not explicitly factored into the velocity profiles generated by the global optimizer. To address this, the integrated velocity scalers have proven to be a straightforward yet effective method to adjust for different traction levels.}

\subsubsection{Local Planning Results}
\begin{figure}[ht]
\centering
\includegraphics[width=\textwidth]{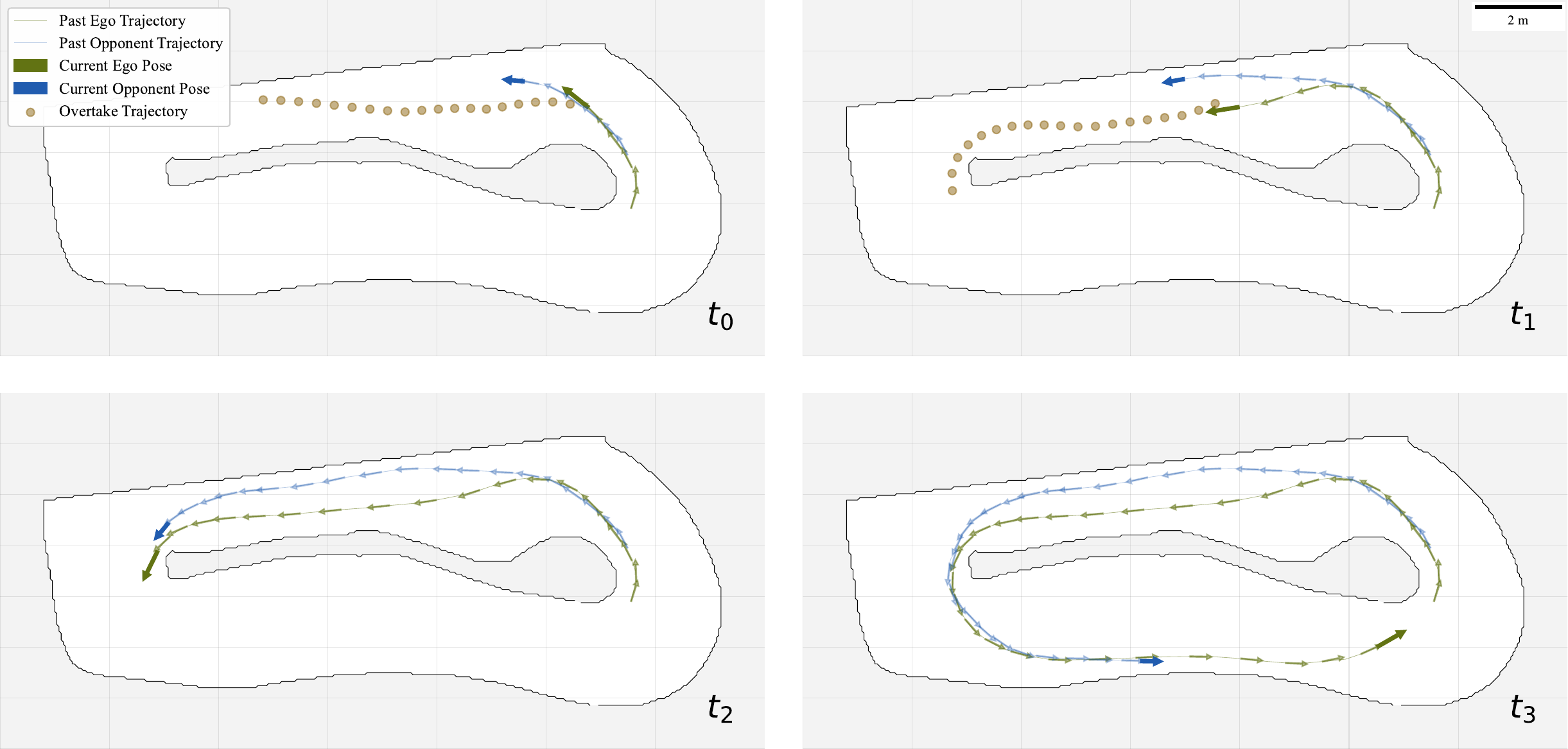}
\caption{Sequential frames of an overtaking maneuver. The blue trajectory represents the opponent, while the green trajectory is that of the ego-agent. The figure also shows the ego and opponent poses, and the planned overtaking waypoints for each timestep. It can be seen that in the frames after the overtake has happened ($t_2,\,t_3$) the planned waypoints are on the nominal racing line\rev{, hence they are no longer plotted}.}
\label{fig:spliner_quali_res}
\end{figure}

To qualitatively assess the overtaking capabilities of the proposed local planner, \revdel{race simulations on real platforms}\rev{real world overtaking experiments on the physical racecar} have been conducted with two agents on a track, as depicted in \Cref{fig:spliner_quali_res}. Both agents were equipped with the \emph{ForzaETH Race Stack}. The ego-agent (green trajectory) was tasked with overtaking a slower opponent (blue trajectory), set to complete the track at a lap time 66\% slower than the ego-agent. \rev{The ego-agent was perceiving the opponent through the aforementioned \emph{Opponent Estimation} module of \Cref{chap:perception}. This demonstrates that the proposed planner is capable of handling the real-world imperfections and disturbances of the entire autonomy pipeline (e.g. localization and opponent estimation inaccuracies).} \rev{To give insights into the decision-making process during the overtaking maneuver, the enumeration of timesteps $t_0-t_3$ provides information on the factors and constraints considered to successfully handle the highly dynamic environment:}

\begin{enumerate}[I]
\item \rev{\textbf{Timestep} \bm{$t_0$}:} The ego-agent catches up with the slower opponent. The local planner identifies a feasible path for overtaking \rev{, by considering if the computed overtaking trajectory satisfies the spatial boundary constraints: \(\max(|\mathbf{d}^{spline}|) \leq d^{track}\)}.
\item \rev{\textbf{Timestep} \bm{$t_1$}:} The ego-agent is actively executing the overtaking maneuver. \rev{The overtaking trajectory is recomputed with the updated position of the opponent and still satisfies the boundary constraints. If the constraints would have no longer been satisfied, the overtaking maneuver would be canceled and the state machine would have activated \textit{Trailing} mode instead.}
\item \rev{\textbf{Timestep} \bm{$t_2$}:} The overtaking is complete. The opponent is now behind the ego-agent, indicating a successful maneuver. \rev{The opponent is no longer visible, hence the \textit{Local Planner} does not propose an overtaking trajectory anymore.}
\item \rev{\textbf{Timestep} \bm{$t_3$}:} The ego-agent resumes \rev{nominal} racing operations.
\end{enumerate}

\rev{Overall, the local planner has demonstrated the ability to adapt to various track layouts. Similar to the global planner, the local planner does not inherently adjust for variations in racetrack friction levels, but adjusting the velocity profile scaling can effectively mitigate this issue. A notable limitation occurs with narrow racetracks where the spline can fail to find a feasible path; reducing $d^{apex}$ can yield a solution but at the cost of riskier overtaking maneuvers. Safety considerations suggest increasing $d^{apex}$ with speed, a point of interaction for the robotic operator. Further overtaking results are elaborated in \Cref{chap:res_h2h}.}

\FloatBarrier
\section{Control} \label{chap:control}

The goal of the \emph{Control} module within the \emph{ForzaETH Race Stack} is to enable performant, reliable, and consistent lap completions by accurately tracking a trajectory provided by the local planner. Therefore, key performance and safety metrics of the control algorithms are lap time, lateral deviation from the racing line, and deviation from the desired velocity.
The velocity and steering commands are managed, respectively, by the longitudinal and lateral controllers.
% The \emph{F1TENTH} vehicle responds to \emph{Ackermann} steering commands, which specify the desired speed and steering angle. The derivation of these control commands is computed by the longitudinal and lateral controllers, which manage velocity and steering, respectively.
\rev{Further, a comparison towards \gls{sota} control algorithms in scaled autonomous racing is performed later on in \Cref{chap:res_tt} when considering the full stack performance.}

\subsection{Architecture}
\begin{figure}[ht]
    \centering
    \includegraphics[width=1\linewidth]{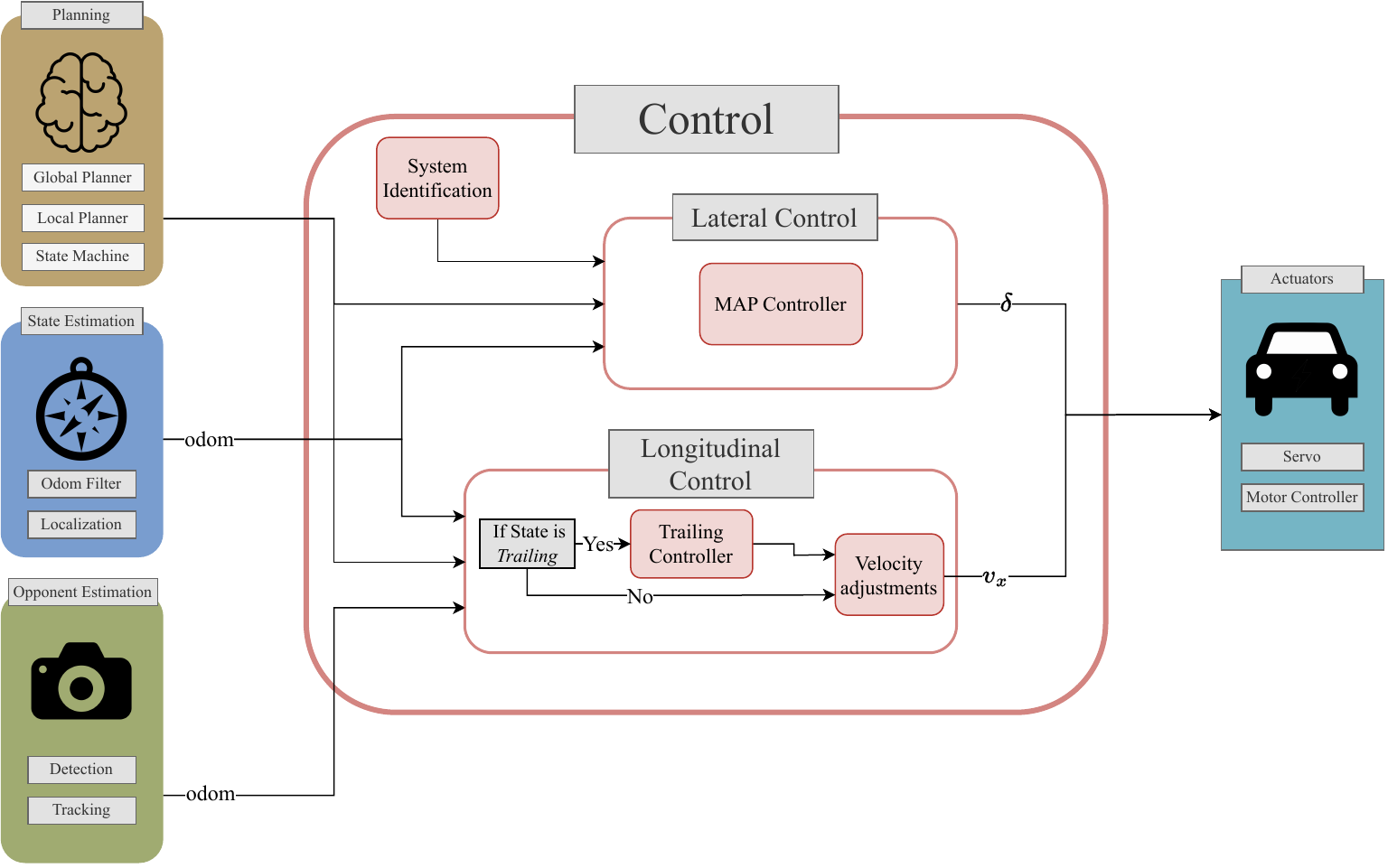}
    \caption{Overview of the proposed control module that depicts the separation into lateral and longitudinal control. The \emph{Control} module, positioned at the end of the \emph{See-Think-Act} sequence, utilizes upstream data from \emph{State Estimation}, \emph{Perception}, and \emph{Planning} to generate control commands to finally execute it on the hardware actuators.
    In case the state provided by the state machine is not \emph{Trailing}, the nominal velocity controller is used, described in \Cref{ssubsec:nom_long_ctrl}. On the other hand, if the state is \emph{Trailing}, the used controller is the trailing velocity controller, described in \Cref{ssubsec:trailing_ctrl}.}
    \label{fig:control_architecture}
\end{figure}

\Cref{fig:control_architecture} depicts the architecture of the \emph{Control} module. It relies on the upstream robotics modules: \emph{State Estimation}, \emph{Perception}, and \emph{Planning}, to compute the final control outputs. The local trajectory, state estimation data, and opponent information are then used to calculate the desired steering angle and velocity. Lateral and longitudinal controllers are separated to simplify development and testing.

\begin{enumerate}[I]
    \item \textbf{Longitudinal Controller:} This submodule computes the desired velocity of the racecar. This velocity is either obtained directly from the velocity of the local trajectory (nominal velocity controller, \Cref{ssubsec:nom_long_ctrl}) or computed by the trailing controller (trailing velocity controller, \Cref{ssubsec:trailing_ctrl}) if the car is trailing an opponent and therefore in the \emph{Trailing} state.
    
    \item \textbf{Lateral Control:} This computes the steering angle to enable accurate spatial trajectory tracking and is based on the \gls{map} controller \cite{map}. It finds the desired steering angle based on the desired lateral acceleration and by leveraging the tire model of the racecar.
\end{enumerate}
%%%%%%%%%%%%%%%%%%%%%%%%%%%%%%%%%%%%%%%%%%%%%%%%%%%%%%%%%%%%%%%%%%%%%%%%%%%%%%%%%%%%%%%%%%%%%%%%
\subsection{Longitudinal Controller} \label{sec:long_contr}

The longitudinal controller computes the velocity input to the \emph{F1TENTH} racecar. This input is then commanded to the \gls{vesc} actuator, which converts the desired velocity to the appropriate \gls{erpm} command with the help of the \gls{vesc} internal low-level \gls{pid} controller.

In both the \emph{GBFree} and \emph{Overtake} states, the racecar derives its reference velocity from the local trajectory generated by the \emph{Planning} module, this is referred to as the nominal velocity control, described in \Cref{ssubsec:nom_long_ctrl}. However, when an opponent obstructs the racing line and overtaking is not immediately viable, the vehicle transitions into the \emph{Trailing} mode, described in \Cref{ssubsec:trailing_ctrl}. In this state, it adjusts the velocity to maintain a safe following distance, thereby preventing potential collisions. This trailing strategy not only upholds safety by avoiding collision with the opponent car but also positions the ego vehicle close to seize an overtaking opportunity when it arises.

\subsubsection{Nominal Velocity Controller}\label{ssubsec:nom_long_ctrl}
When operating in the \emph{GBFree} or the \emph{Overtake} state, the car adopts its target velocity from a corresponding point along the local trajectory. 
The reference velocity is first obtained by querying the velocity profile \( v_{traj} \). 
As the global planner provides a discretized velocity profile, with a discretization step every \SI{0.1}{\metre}, the $s$ coordinate is first rounded, then used to obtain the required velocity. For ease of notation, the velocity lookup on the precomputed racing line for a predetermined $s$ is here noted with $v_{traj}(s)$.

% , originally obtained by the global planner (see \Cref{subsec:gb_optimizer}):
% \begin{align}
%     v_{traj}: [ 0,\,s_{max} ) &\longrightarrow \mathbb{R}\\
%     s &\longmapsto v_{traj}(s) \nonumber
% \end{align}

% $s_{max}$ is the length of the global trajectory and $v_{traj}(s)$ is then the velocity from the speed profile correspondent to a chosen $s$ value.
% This mapping is, in practice, implemented via a \gls{lut} with a discretization step of \SI{10}{\centi \metre} for the $s$ input variable.

To account for actuation and computation delay, the actual reference velocity is obtained from an $s$ coordinate forward-propagated in time. A lookahead distance $s_{la}$ is first calculated using a $t_{la}$ lookahead time and the velocity of the car in the \emph{Frenet} $s$ coordinate $v_s$, then $s_{la}$ is summed to the current car position in the $s$ coordinate $s_{ego}$ to evaluate the reference velocity $v_{ref}$:
\begin{equation}
    v_{ref} = v_{traj}(s_{ego} + s_{la}).
\end{equation}
We further apply the following heuristic, to scale down the reference velocity when the lateral deviation is not zero, using the term $\alpha_{lat\_dev}$ defined as follows:
\begin{align} \label{eq:v_des_nom}
 v_{des} &= \alpha_{lat\_dev} \cdot v_{ref} \\ 
 &= \left(  1 + \lambda_{lat}(-1 + e^{-  d_{norm} \cdot c_{norm}}) \right) \cdot v_{traj}(s_{ego} + s_{la}). \nonumber
\end{align}
Here $\lambda_{lat} \in [0,1]$ determines how much the lateral error is accounted for. A value of $0$ means no accounting for lateral error and a value of $1$ means maximal speed reduction on lateral error.  Both $d_{norm}$ and $c_{norm}$ are normalized values in $[0,1]$ representing the lateral deviation from the desired trajectory and the curvature of the trajectory at the current position.
This adjustment helps the car slow down when it is not on the racing line, in order to reconnect more smoothly to the trajectory.
The final value sent to the \gls{vesc} for low-level actuation is then $v_{des}$ defined as in \Cref{eq:v_des_nom}.

\subsubsection{Trailing Velocity Controller}\label{ssubsec:trailing_ctrl}

\begin{figure}[ht]
    \centering
    \includegraphics[width=\linewidth]{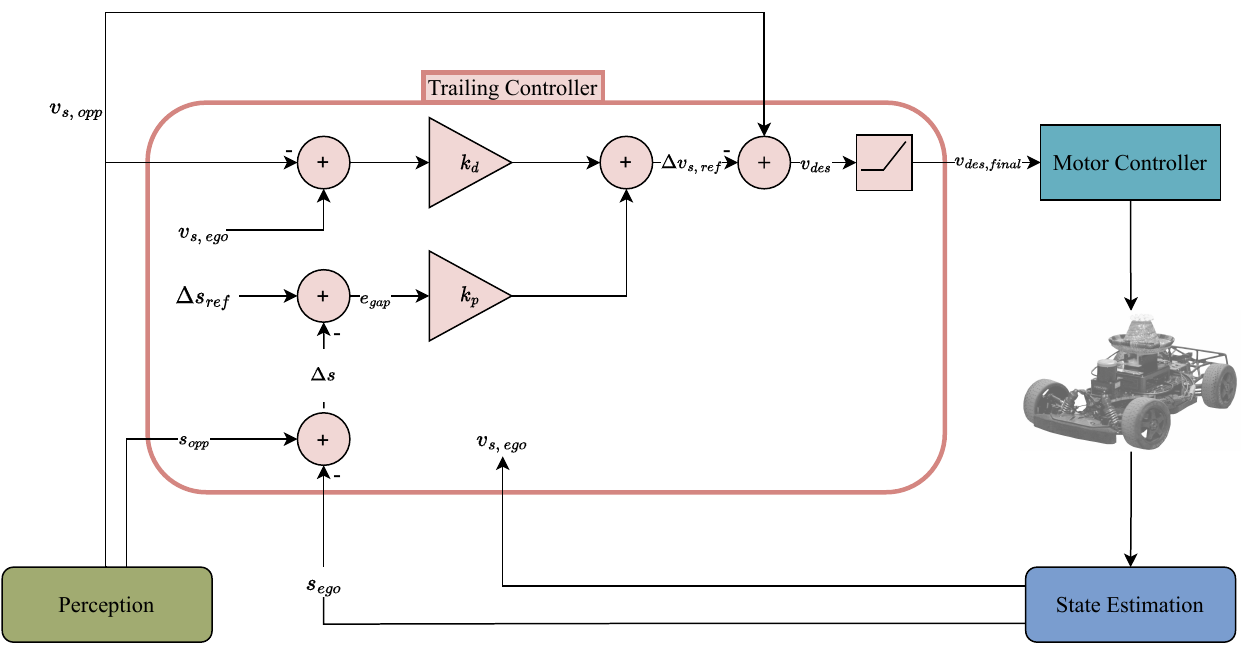}
    \caption{Diagram of the trailing control flow. The trailing controller represents a subset of the overall \emph{Control} module of \Cref{fig:control_architecture}. It is responsible for maintaining a close and steady distance towards the opponent while avoiding collisions. This allows the agent to set itself up well for a possible overtaking maneuver.}
    \label{fig:trailing_controller}
\end{figure}

If the car is in state \emph{Trailing}, the goal is to adjust the ego-car velocity to the one of the to-be-trailed opponent, such that a constant distance gap can be achieved. This is achieved with a \gls{pd} controller with a feedforward term of the opponent's velocity as displayed in \Cref{fig:trailing_controller} \rev{and is inspired by previous work on \gls{acc}, such as detailed in \cite{ACC_HE201921}}. 

The $v_{des}$ for the trailing controller is then defined as follows:
\begin{equation}\label{eq:v_trail_ctrl}
    v_{des} = v_{s,\,opp} - (k_{p} e_{gap} + k_d \Delta{v_s}),
\end{equation}
where $v_{s,\,opp}$ is the velocity of the opponent in the \emph{Frenet} s dimension. $k_p$ is the gain of the proportional error term $e_{gap}$, which is the difference between the fixed target gap $\Delta s_{ref}$ and the measured gap $\Delta s$. The gain $k_d$ is applied to the derivative term, which is obtained by subtracting the opponent's velocity along the \emph{Frenet} $s$ dimension $v_{s,\,opp}$ to the ego-car's current velocity along the \emph{Frenet} $s$ dimension $v_{s,\,ego}$. 
\begin{align} \label{eq:g_kurwa}
    e_{gap} &= \Delta{s_{ref}} - ((s_{ego} - s_{opp}) \,\%\, s_{max})\\
    \Delta v_s &= v_{s,\,ego} - v_{s,\,opp}.
\end{align}
In \Cref{eq:g_kurwa} $\%$ represents the modulo operator, to account for the $s$ coordinate wrapping over multiple laps. On top of the core part, the \gls{pd} controller, some considerations were made to improve the controller's performance in real-world scenarios:
\begin{enumerate}[i]
    \item Only the closest obstacle is considered
    \item If both static obstacles and dynamic obstacles are present, only dynamic obstacles are considered
    \item The velocity is clipped to a minimum velocity
\end{enumerate}

Given the current \emph{F1TENTH} \emph{Head-to-Head} racing rules, which only allow for one opponent on the track, the first consideration should not affect the car's behavior, however, it still remains necessary when multiple opponents can be detected in free practice settings and more than two cars are present on the track at the same time. 
The second consideration then arises from the fact that some false positives can still arrive from the \emph{Perception} module (e.g. caused by reflections on the floor, modified track boundaries, etc.), but since no static obstacle is present on the track during the \emph{Head-to-Head} phase if both types of obstacles are present, static classifications are deemed false positives and dynamic obstacles are preferred. If, instead, only static obstacles are present, the trailing controller will choose the closest (and eventually stop in case the object does not disappear).
The final consideration is then added to avoid rare cases where the trailing controller starts trailing very slow obstacles outside of \gls{los} (e.g. in a hairpin) which could lead even to a halt, in case the detected velocity is low enough. 
When a minimum velocity $v_{blind}$ is instead enforced, in the cases when there is no \gls{los} of an obstacle, a minimum velocity will still be kept, in order to move to a position with a better view of the potential obstacles. 
The final velocity sent to the low-level controller is then obtained as follows:
\begin{equation}
    v_{des,\,final} = \max(v_{blind},\,v_{des} ).
\end{equation}

%%%%%%%%%%%%%%%%%%%%%%%%%%%%%%%%%%%%%%%%%%%%%%%%%%%%%%%%%%%%%%%%%%%%%%%%%%%%%%%%%%%%%%%%%%%%%%%%
\subsection{Lateral Controller}
The lateral controller commands steering angles $\delta$ to the servo, with the goal of making the car follow a desired trajectory, as generated by the planner (\Cref{sec:planning}). To achieve this, the \gls{map} controller from \cite{map} was chosen, due to \revdel{its high tracking accuracy and little computational complexity}\rev{its low computational complexity and high tracking accuracy, the latter derived from the theoretical properties of \cite{l1}}. It outperforms \gls{sota} geometric controllers like \emph{Pure Pursuit} \cite{tunercar_icra_2020}, by incorporating tire-slip through a model of the car's steady state cornering behavior. 
The following sections lay out how this model is obtained experimentally and how the \gls{map} controller is set up and tuned.

%%%%%%%%%%%%%%%%%%%%%%%%%%%%%%%%%%%%%%%%%%%%%%%%%%%%%%%%%%%%%%%%%%%%%%%%%%%%%%%%%%%%%%%%%%%%%%%%
\subsubsection{System Identification} \label{sec:sysid}
%\assign{Jonny} here state how SysID is handled and how we estimate our model. Just describe a bit more in detail how SysID procedure works a bit more in detail than in MAP paper.\\
The system identification follows the procedure outlined in \cite{map}. Notably, the vehicle dynamics are modelled using a single-track model \cite{commonroad}. Tire slip is incorporated trough the \textit{Pacejka} tire model \cite{pacejka1992magic}. Most physical static model parameters can be measured, namely the mass $m$ with a scale and the distances of the front and rear axle from the \gls{cg} ($l_r$, $l_f$) by balancing the car on a string and measuring the distances from the front and rear axles. Furthermore, the height of the \gls{cg} above the axles $h_\text{cg}$ was obtained by first vertically balancing the car to identify the vertical position of the \gls{cg} and then, to take suspensions into account, the car was put on the ground and the final $h_\text{cg}$ was obtained. The yaw moment of inertia $I_{zz}$ is calculated based on the oscillation period in a bifilar pendulum experiment \cite{green1927measurement}.

The characterizing parameters of the tire model are obtained through steady-state cornering experiments, as described by \cite{sysidVoser}. The car is driven in circles at a constant speed and with slowly varying steering angles, resulting in a range of quasi-steady state measurements of the lateral acceleration for a sweep of tire slip angles. 
In particular, the steering angle is increased with \SI{0.02}{rad/s} from neutral to the negative and the positive maximum for 3, 4, and \SI{5}{m/s} respectively. 
Following equations (7 - 14) of \cite{map}, the lateral forces produced by the front and rear tires are related to the respective tire slip angles. 
The parameters of the tire model are obtained by solving a nonlinear least-squares problem, using the \texttt{least\_squares} function from the \texttt{Python} library \texttt{scipy}. 
Outliers are removed through three steps $k$ of expectation maximization, removing any measurements more than $\frac{10}{2^k}$ away from the previous estimate. 
To constrain the curves' shapes for high tire slip angles, the shape and curvature factor of the \textit{Pacejka} model are limited to 1.5 and 0.8 respectively for the rear axle and 4.0 and 1.1 for the front axle. 
The scripts to automatically run the experiments and analyze the recorded data are part of the published race stack.
In a racing scenario, such measurements would be conducted usually outside of the track, as a space of approximately \SI{10}{\metre} by \SI{5}{\metre} is needed.

%%%%%%%%%%%%%%%%%%%%%%%%%%%%%%%%%%%%%%%%%%%%%%%%%%%%%%%%%%%%%%%%%%%%%%%%%%%%%%%%%%%%%%%%%%%%%%%%

\subsubsection{MAP Controller}
% The racing line is given by waypoints consisting of coordinates and associated velocities.

\begin{figure}[ht]
    \centering
    \subfloat[Conceptual mechanism of how \gls{map} works\label{fig:map_overview}]{\includegraphics[width=0.6\textwidth]{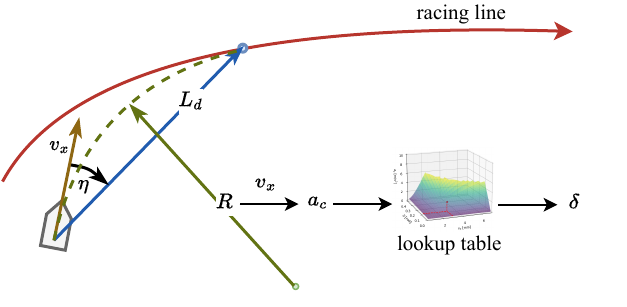}}
    \hfill
    \subfloat[Lookup table\label{fig:pacejka_lu}]{\includegraphics[width=0.4\textwidth]{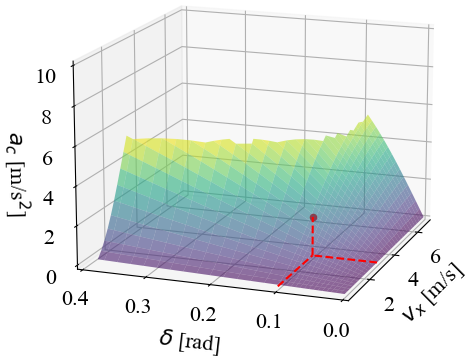}}
    \caption{(a) \gls{map} obtains required centripetal acceleration $a_c$ to reach a lookahead point $L_d$ using the $L_1$ guidance law. (b) The lookup table is used to retrieve the steering angle $\delta$ that results in the calculated $a_c$ for the speed $v_x$.}
    \label{fig:map-vs-pp2}
\end{figure}

Similar to traditional geometric controllers like \emph{Pure Pursuit}, \gls{map} calculates steering commands intending to reach a lookahead point. This is a point on the trajectory at tunable distance $L_d$ from the car, as depicted in \Cref{fig:map_overview}. The advantage of \gls{map} over traditional geometric control lies in the incorporation of tire slip into the calculation of steering inputs. \revdel{This has been shown to}\rev{By accounting for wheel-slip it} significantly improve\rev{s} its tracking performance in high-speed corners \cite{map}. The \gls{map} controller achieves this by computing the required centripetal acceleration $a_c$ to reach the lookahead point and retrieves the corresponding steering angle from a precomputed lookup table.

The calculation of lateral acceleration is based on the $L_1$ guidance \cite{l1}, originally designed for fixed-wing \gls{uav}s. It finds the lateral acceleration $a_c$ making the vehicle with longitudinal speed $v_x$ follow a circular motion with radius $R$ onto the lookahead point. This is obtained from the following relationship:
\begin{equation}
    a_{c} = 2 \frac{v_x^2}{L_d}sin(\eta) = \frac{v_x^2}{R},
    \label{eq:l1_guidance}
\end{equation}
where $\eta$ denotes the angle between the heading and the lookahead point as illustrated in \Cref{fig:map_overview}. 

As there is no explicit solution to map steering angles to lateral accelerations, a lookup table is used, which is generated offline to keep computation lightweight. 
It is obtained by simulating the identified single-track model dynamics of \Cref{sec:sysid} for a sweep of constant speeds and steering angles. 
These simulations, conducted for two seconds of simulated time, are repeated for velocities from 0.5 to \SI{7}{m/s} at \SI{0.1}{m/s} intervals and for a range of steering angles from 0 to \SI{0.4}{rad} at steps of \SI{0.01}{rad}. 
A finer step size of \SI{0.0033}{rad} is used for the steering angles between 0 and \SI{0.1}{rad}. 
The final steady-state lateral acceleration $a_c$ is stored. 
If no steady state is reached the pair of velocity and steering angle is considered unstable. This results in a stable steering region illustrated as a surface in \Cref{fig:pacejka_lu}.

% To account for the impact of actuation and computation delays, the controller projects the current vehicle position into the future by an estimated delay duration \SI{0.2}{s}. Subsequently, it selects the speed $v_x$ associated with the nearest waypoint to the projected position, as this speed corresponds to the expected vehicle velocity at the time the steering command is in effect.

As proposed in \cite{map}, the lookahead distance parameter $L_d$ is scaled with the vehicle's current speed $v$ using an affine mapping: $L_d = mv + q$. The tunable coefficients, $m$ and $q$, are used to adjust the scaling.

This tuning of $L_d$ and the scaling is critical to achieving stable and effective control. If the distance is set too short, especially at high speeds, it can result in undesirable vehicle oscillations. Conversely, setting the distance too large may tempt the vehicle to cut corners and potentially lead to collisions. Therefore, the goal is to keep $L_d$ as small as possible without it causing oscillations, striking a balance between stability and performance.

\subsection{Control - Results}
Here, the performance of the selected control algorithms is analyzed. Firstly the results of the longitudinal controller are examined, covering velocity tracking accuracy and the longitudinal distance-keeping outcomes of the trailing controller. Subsequently, the lateral control results, regarding the lateral tracking accuracy \rev{and how it changes with different tracks and velocity scalers} are explored.
\rev{For a comparison of the \gls{map} controller with the other baseline controllers implemented in the \emph{ForzaETH Race Stack}, we refer the reader to \Cref{chap:res_tt}.}
% Subsequently, we explore the lateral control results, emphasizing the superior trajectory tracking capabilities of \gls{map} with respect to \emph{Pure Pursuit}.

\subsubsection{Velocity Controller Results}
In this section, the results for the nominal and trailing velocity controllers on the physical platform are reported. 
In both cases, the car was run on the racing track shown in \Cref{fig:trailing_gap_comb}. 
In the nominal case, the sector scalers were set uniformly to 60\%. This scaler was chosen to drive fast lap times while also ensuring consistent driving behavior.
In the second case, the ego car was set in a permanent \emph{Trailing} state and a second car was run with a 40\% sector scaler setup on the same racing line, to ensure continuous operation of the trailing controller.
% In the second case, a second car was run with a 33\% slower setup on the same racing line, in order to put the ego car in a permanent \emph{Trailing} state.
In both cases, data was recorded over ten laps, and both ego and opponent velocities were recorded running the \emph{State Estimation} module respectively on the two cars. Data was then synchronized with the \gls{ros} timestamp. 

\begin{figure}[ht]%
\centering

\captionsetup[subfloat]{font=footnotesize,labelfont=bf, textfont = normalfont}
\subfloat[\emph{GBFree}\label{subfig:vel_ctr_gbfree}] {\includegraphics[width=0.45\textwidth]{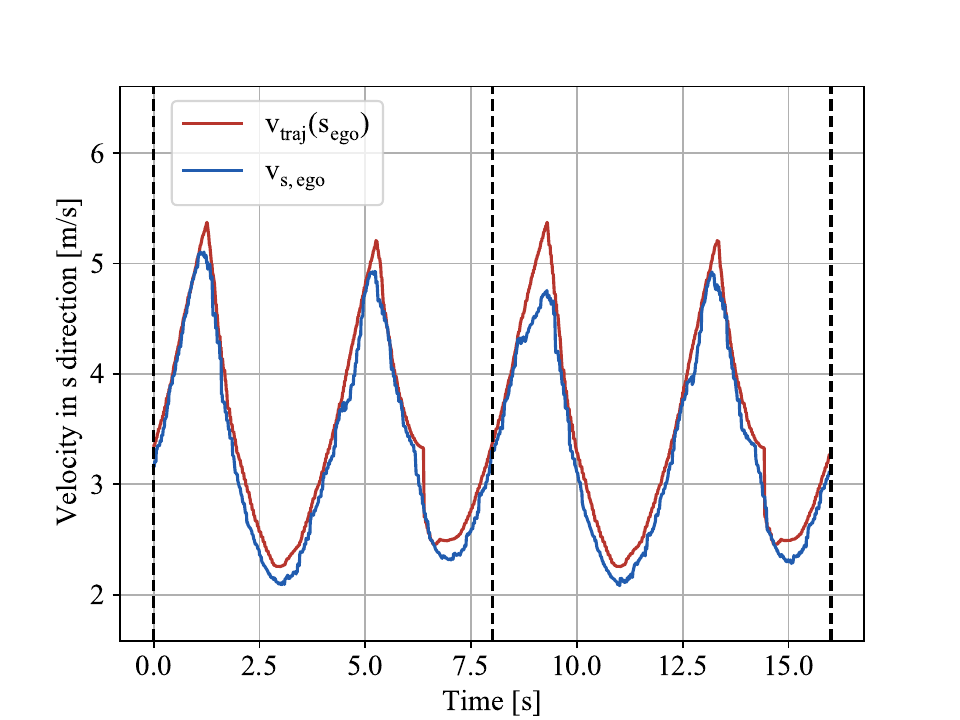}}\qquad
\subfloat[\emph{Trailing}\label{subfig:vel_ctr_trailing}] {\includegraphics[width=0.45\textwidth]{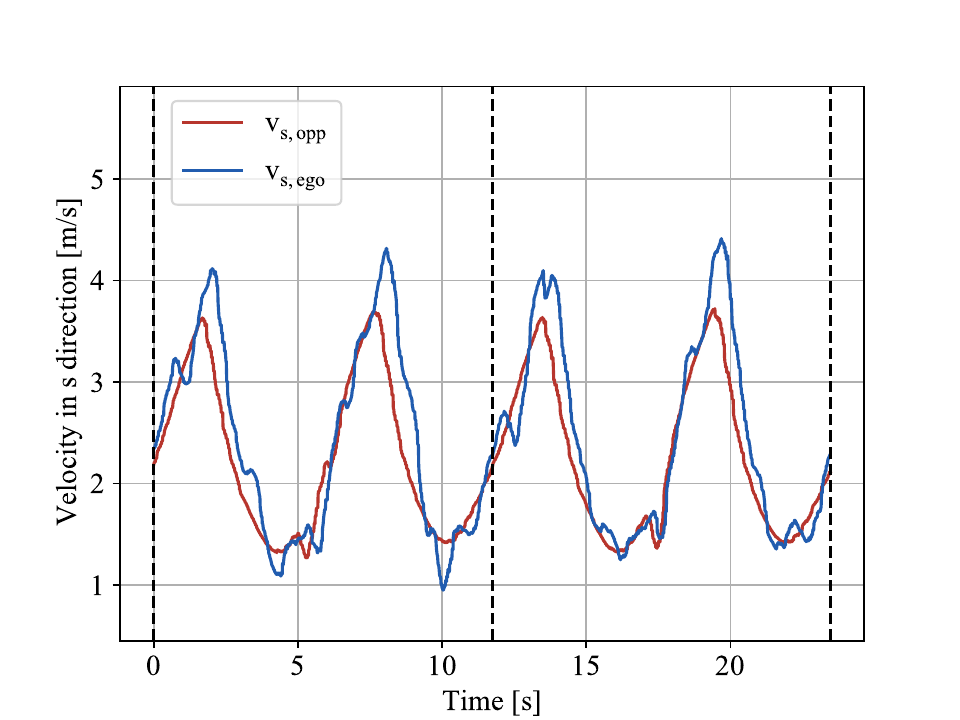}}\qquad
\caption{Velocity data over two consecutive laps, on the left the car is driving in the state \emph{GBFree} and on the right trailing an opponent. The to-be-tracked reference velocity is the velocity of the global waypoints in the state \emph{GBFree} and the opponent's velocity in the state \emph{Trailing}. The start of a new lap is marked by the black dashed line.}
\label{fig:vel_ctr}
\end{figure}

In \Cref{fig:vel_ctr}, the outcomes of the velocity control system are showcased. The longitudinal velocity directives are compared with the velocity estimates derived from state estimation. When driving in state \emph{GBFree}, the actual velocity of the racecar as estimated by the \emph{State Estimimation} is compared to the racing line velocity determined by the global planner. As can be seen in \Cref{fig:vel_ctr} the computation and activation delay is compensated by taking the velocity of the racing line at a point in front of the racecar as described in \Cref{sec:long_contr}. When driving in state \emph{Trailing}, the actual velocity of the racecar is compared to the velocity of the opponent's car. Although the ego-car velocity generally tracks that of the opponent, the trailing controller is designed to maintain a constant gap to the opponent. Consequently, velocity divergence occurs at certain points to rectify and uphold the desired gap to the opponent.

\subsubsection{Trailing Controller Results}
To test the trailing controller, the car was tuned to drive the fastest possible lap times without crashing, which was equivalent to a uniform 60\% scaler. Then the trailing controller was tested against three slower opponents with different driving styles by driving four consecutive laps behind the opponent.
The three different driving styles presented are a slower opponent with the same racing line and a 40\% scaler, a \gls{ftg} opponent with a constant speed, and a manually driven car.

\begin{figure}[ht]
    \centering
    \includegraphics[width=\columnwidth]{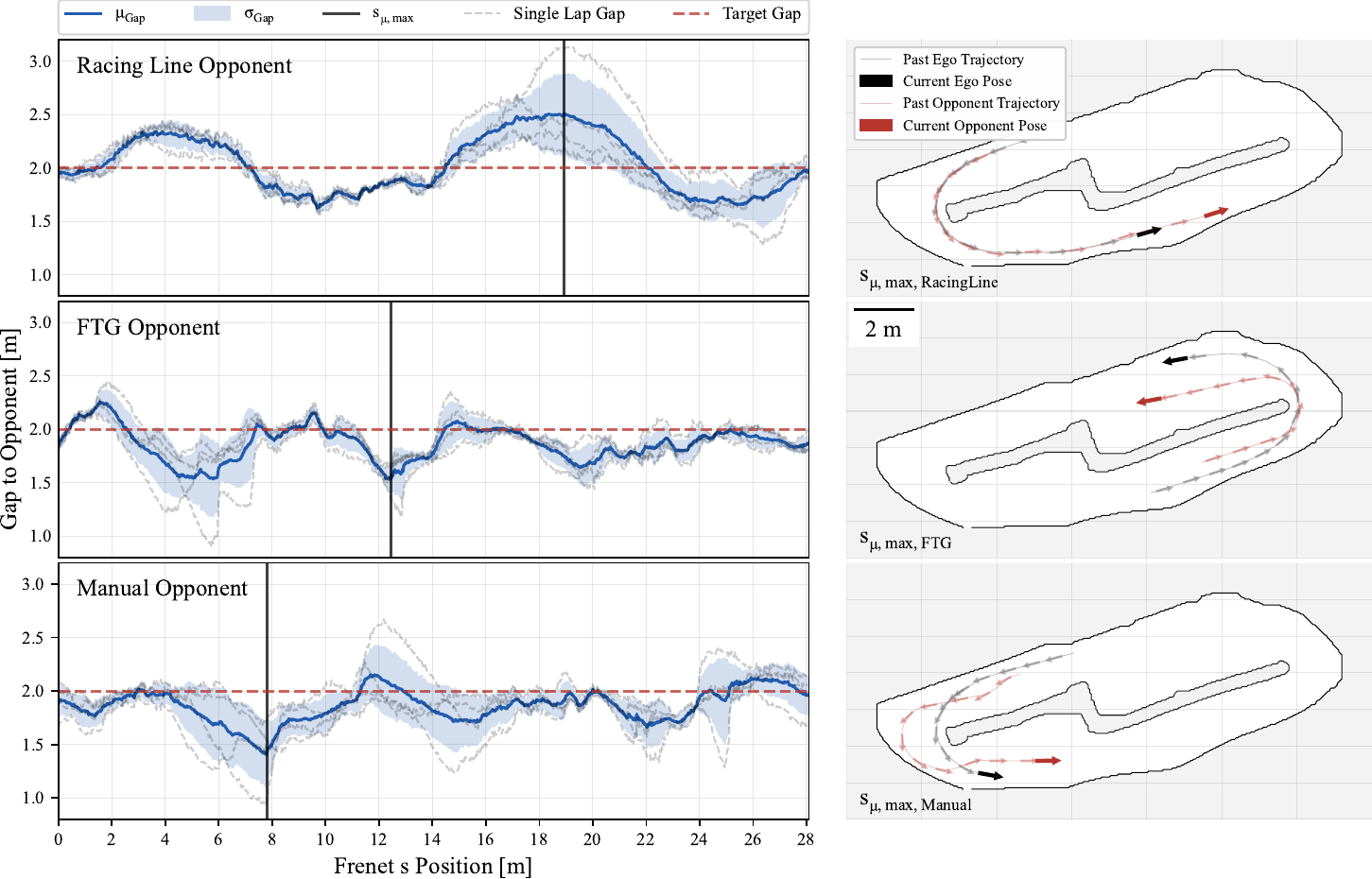}
    \caption{Trailing gap over four consecutive laps against three different opponents is shown. The reference gap is depicted in red, the mean and standard deviation from the reference gap in blue, and the measurements of the gap in each lap in grey.}
    \label{fig:trailing_gap_comb}
\end{figure}

The results presented in \Cref{fig:trailing_gap_comb} show a consistent behavior over four consecutive laps, even against the less predictable manually driven opponent. Against the racing line opponent it can be observed, that the gap to the opponent overshoots at two sections on the track.
These are the straights where the opponent is accelerating and the gap increases until the ego-car catches up. 
The mean deviation from the reference gap $\Delta \bar{\mu}_{Gap}$ and the maximal deviation from the reference gap $\Delta \mu_{Gap,\,max}$ are listed in \Cref{tab:trailing_eval}. $\Delta \mu_{Gap,\,max}$ never exceeds \SI{1.2}{\metre}, therefore a reference trailing gap of \SI{1.5}{\metre} - \SI{2}{\metre}, can be chosen to trail safely.
% and still optimize the chance of overtaking.

\begin{table}[ht]
    \centering
    \begin{tabular}{l|l|l}
        \textbf{Opponent Style} & \textbf{$\bm{\Delta\mu_{Gap,max}}$ [m]} & \textbf{$\bm{\Delta\bar{\mu}_{Gap}}$ [m]}  \\
        \hline
        \hline
        Racing Line Opponent & \SI{1.13}{} & \SI{0.23}{} \\
        \gls{ftg} Opponent & \SI{1.07}{} & \SI{0.18}{}   \\
        Manual Opponent & \textbf{1.06}  & \textbf{0.17}  \\
    \end{tabular}
    \caption{Evaluation of the trailing controller against three different opponents.}
    \label{tab:trailing_eval}
\end{table}

\subsubsection{Trajectory Tracking Results}
\label{chap:ctrl_tracking_results}
\begin{figure}[ht]
    \centering
    \includegraphics[width=\linewidth]{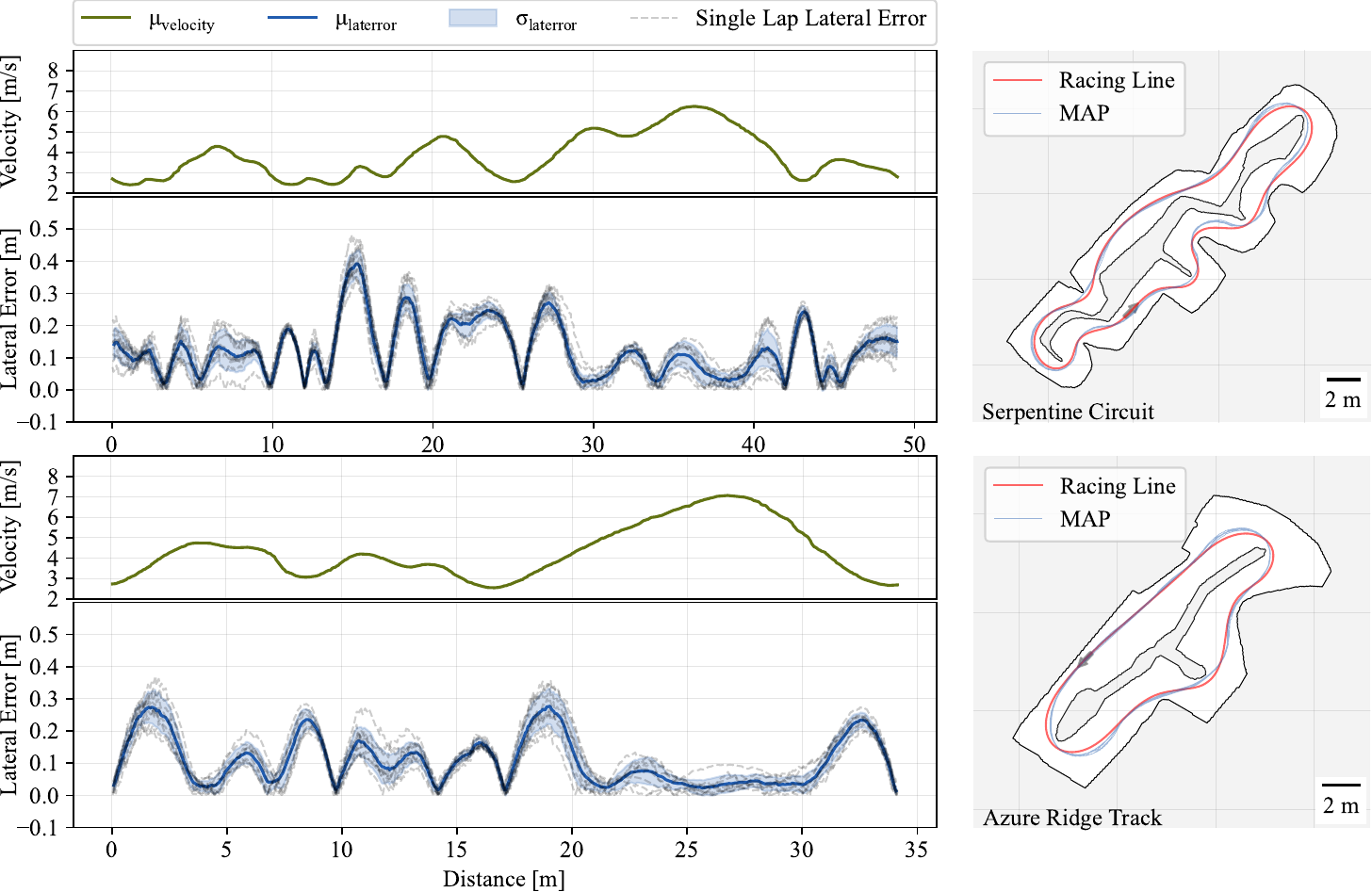}
    \caption{Ten laps of the \gls{map} controller on two different real-world tracks tracking a reference racing line. The left side shows velocity and average and per lap lateral tracking error. On the right, the map and trajectories of the laps are shown.}
    \label{fig:map-vs-pp}
\end{figure}

The tracking performance of the \gls{map} controller is illustrated in \Cref{fig:map-vs-pp}, depicting trajectories over ten laps on two different tracks, as well as the lateral tracking error and the velocity. The controller was tuned to maximize speed while accurately following the racing line (represented in red). The tuning process involved adjusting lookahead distance and scaling parameters ($L_d$, $m$ and $q$) on the first track, scaling the trajectory speed to 75\%. The speed was then incrementally increased until it reached the limits of the track bounds.

Trajectory tracking results, with lap times and lateral errors, can then be seen in \Cref{tab:tracking-results}.
On the \emph{Serpentine Circuit}, \gls{map} consistently reached 79\% of the maximum speed without crashing. This map was used to perform tuning. On \emph{Azure Ridge Track}, the tuning parameters, obtained from \emph{Serpentine Circuit} were then used to demonstrate the adaptability of the controllers across different tracks. This track reaches higher velocities than \emph{Serpentine Circuit} due to its long straight. On this track, \gls{map} was able to consistently reach 81\% of the maximum speed without crashing, due to the easier configuration of the circuit, presenting fewer corners.
\Cref{fig:map-vs-pp} shows the two tracks where the trajectory speed for both maps was scaled to 79\% for comparison. This indicates that \gls{map} can effectively adapt to different tracks when the tuning yields a sufficiently low lateral error ($\sim$\SI{10}{\centi \metre} in this case) and such error performance is maintained of a different track.
When pushing to the limits, instead, and lateral error is higher, behavior deviates more from the expected racing line, and performance can be transferred only partially.
% \todo{@Nadine New more curvy track(s)?}

\begin{table}[ht]
    \centering
    \begin{tabular}{l|c|c|c|c|c|c|c}
        \textbf{Track ($\sigma^{scaler}$)} & \multicolumn{4}{c|}{\textbf{Lap Time [s] $\downarrow$}} & \multicolumn{3}{c}{\textbf{Lateral error [cm] $\downarrow$}}\\
        & $\mu_{t}$ & $\sigma_{t}$ & $t_{min}$ & $t_{max}$ & $\mu_{err}$ & $\sigma_{err}$ & $err_{max}$\\
        \hline
        \hline
        \rowcolor{lightgray}
        Serpentine Circuit (75\%) & 14.80 & 0.11 & 14.62 & 14.99 & 9.35 & 0.97 & 42.08 \\ % results for 75
        \hline
        Serpentine Circuit (79\%)& 14.47 & 0.09 & 14.38 & 14.66 & 12.99 & 0.56 & 47.85 \\
        \hline
        \rowcolor{lightgray}
        Azure Ridge Track (75\%)& 9.04 & 0.09 & 8.89 & 9.21 & 7.46 & 0.70 & 21.31\\
        \hline
        Azure Ridge Track (79\%)& 8.77 & 0.12 & 8.52 & 8.98 & 11.67 & 1.01 & 37.06\\
    \end{tabular}
    \caption{The figure shows average, standard deviation, minimum, and maximum lap time, respectively as $\mu_t,\,\sigma_t,\,t_{min},\,t_{max}$.  Furthermore, the average, standard deviation, and maximum lateral error with respect to the racing line, are shown, respectively, with $\mu_{err},\,\sigma_{err},\,err_{max}$. The used uniform \emph{sector scalers} are signaled with $\sigma^{scaler}$.
    Data was gathered on the tracks shown in \Cref{fig:map-vs-pp}. The statistics have been computed from ten laps each.}
    \label{tab:tracking-results}
\end{table}
%%%%%%%%%%%%%%%%%%%%%%%%%%%%%%%%%%%%System%%%%%%%%%%%%%%%%%%%%%%%%%%%%%%%%%%%%
\FloatBarrier
\section{Full Stack Performance}
The previous sections detailed the performance of individual components within the \emph{See-Think-Act} cycle, specifically \emph{State Estimation}, \emph{Opponent Estimation}, \emph{Planning}, and \emph{Control}. These assessments aimed to quantify each module's performance in an isolated context --- as far as possible. This chapter presents the performance analysis of the full \emph{ForzaETH Race Stack} when all modules operate interconnected with the focus on evaluating the integrated system behavior during \emph{Time-Trials} and \emph{Head-to-Head} racing scenarios.

\rev{Additional results on the racing performance of the \emph{ForzaETH Race Stack} during official races are described in \Cref{sec:app_race_results}. However, as the setting of the official race does not allow for extensive data recordings, to minimize additional compute and to focus on performance during the official race, the current section presents the full stack performance based on a recreated racing scenario, consisting of both \emph{Time-Trials} and \emph{Head-to-Head}.}

\subsection{Time-Trials Performance}\label{chap:res_tt}
\begin{figure}[ht]
    \centering
    {\includegraphics[width=0.49\textwidth]{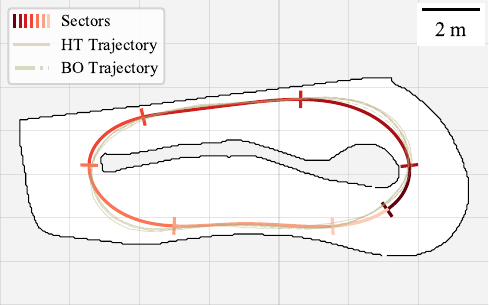}}
    \includegraphics[width=0.49\textwidth]{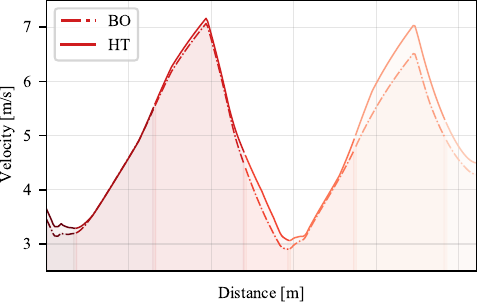}
    \caption{On the left, the image displays sectors as defined for the displayed circuit. Sectors are ordered from zero to six from the darkest hue to the lightest. A perpendicular segment is used to indicate the beginning of the sector. The trajectories following the hand-tuned (HT) sectors and the \gls{bo} tuned sectors are also shown, respectively in bronze and green. On the right, the final velocity profiles resulting from the tuned sector scalers are displayed. The final values resulting from the \gls{bo} setup were \texttt{[0.76778, 0.76083, 0.78929, 0.78776, 0.75155, 0.75019, 0.77131]}. The final values resulting from the hand-tuning setup were \texttt{[0.81, 0.76, 0.80, 0.86, 0.76, 0.81, 0.81]}.}
    \label{fig:tt_sector_tuning}
\end{figure}

The \emph{Time-Trials} stage in \emph{F1TENTH} racing emphasizes achieving the fastest lap time while maintaining a high number of consecutive crash-free laps. This necessitates the ability to execute laps that are rapid, safe, and consistent. To assess these aspects, the \emph{ForzaETH Race Stack} was configured as outlined and deployed on a test track (see \Cref{fig:tt_sector_tuning}). In this environment --- free from obstacles and opponents --- ten laps were recorded to evaluate racing performance, focusing on lap time, deviation from the optimal racing line, and consistency as indicated by the standard deviation of these metrics.
% For ease of notation, the average lap time is indicated with $\mu$, with a subscript to indicate which method was used for tuning the sector scalers. $\mu_u$ indicates a uniform setup for the scalers, a number e.g. $\mu_{75}$ indicates a uniform sector scaler at that value, $\mu_{HT}$ indicates hand-tuned values, and $\mu_{BO}$ indicates Bayesian optimization as a way of tuning.

\begin{figure}[ht]
    \centering
    \includegraphics[width=\textwidth]{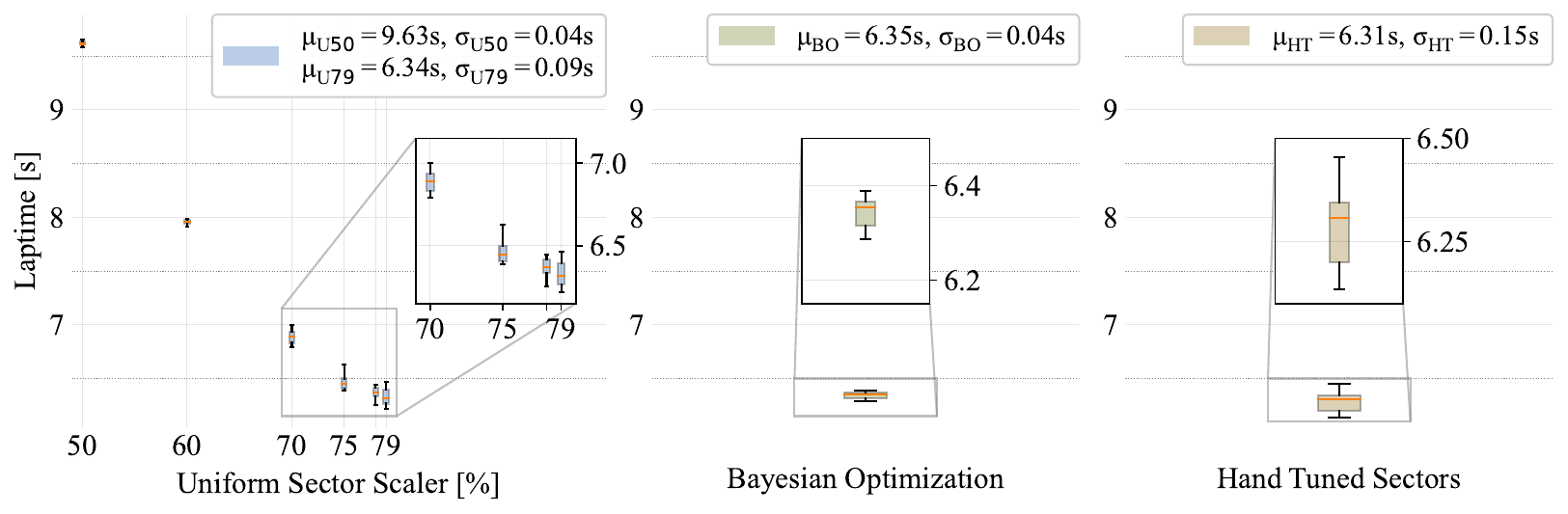}
    \caption{Lap times across 10 laps with the proposed \emph{ForzaETH Race Stack} using the \gls{map} controller. Uniform sector scalers in blue, \gls{bo} obtained sector scalers in green, and hand-tuned sector scalers in bronze. The mean and standard deviation of all corresponding lap times are depicted in the top right, zoom in to enhance visibility when necessary.}
    \label{fig:tt_laptime}
\end{figure}

\begin{figure}[ht]
    \centering
    \includegraphics[width=\textwidth]{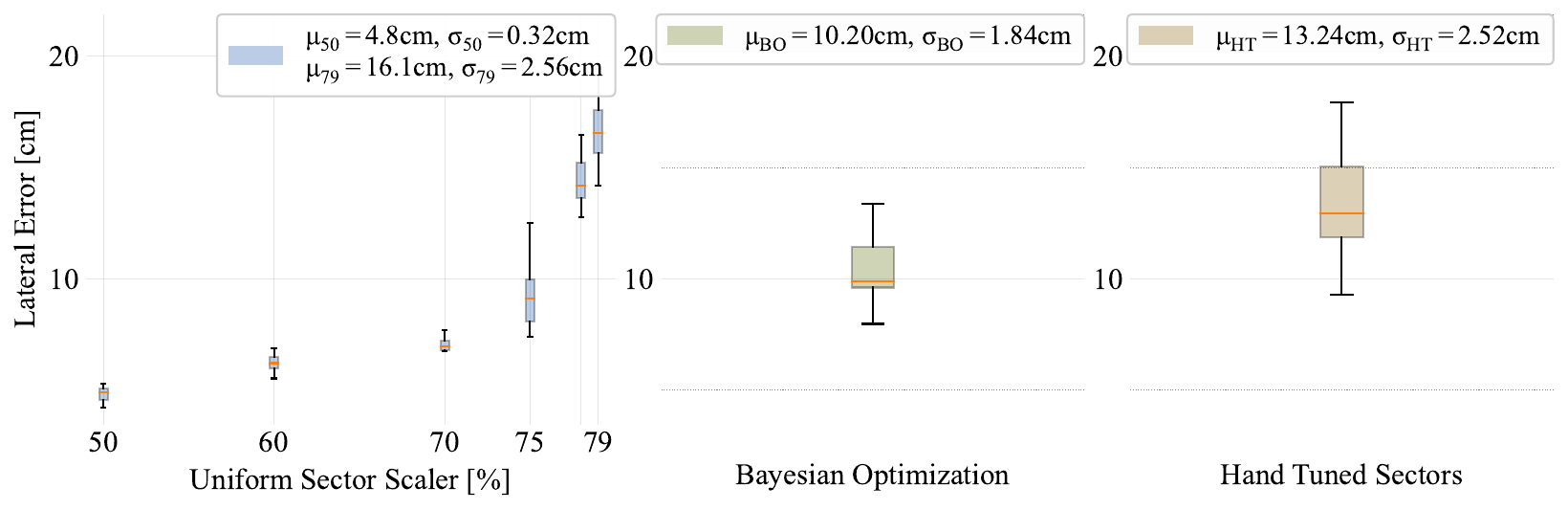}
    \caption{Lateral error across 10 laps, relative to the optimal racing line using the \emph{ForzaETH Race Stack} with the \gls{map} controller. Uniform sector scalers (blue), \gls{bo} scalers (green), and hand-tuned scalers (bronze). The mean and standard deviation are in the top right.}
    \label{fig:tt_latdev}
\end{figure}

In racing, tuning the racing line to match specific environmental conditions, such as varying track surface friction coefficients, is crucial for achieving performant behavior. This necessitates adjusting the velocity profile of the racing line to align with the track layout and grip levels. The following tuning strategies, as outlined in \Cref{subsec:gb_optimizer}, were employed to adjust the velocity scalers for each sector and refer to \Cref{fig:tt_laptime} and \Cref{fig:tt_latdev}:

\begin{enumerate}[I]
\item \textbf{Uniform Sector Scalers:} Sectors are uniformly scaled by a constant factor, a quick but often suboptimal method. Analysis shows a linear decrease in lap times up to a $78\%$ scaler, then a minimum is obtained with a $79\%$ scaler, yielding a lap time of $\mu_{u}=6.34s$. The lateral error increases slightly up to $70\%$ then sharply rises from \SI{5.0}{\centi\metre} to \SI{16.1}{\centi\metre} in the $70-79\%$ range, indicating competitive lap times but with increasing safety risks. Higher sector scalers, from $80\%$ onwards, led to collisions with the boundaries.
\item \textbf{Bayesian Optimized Scalers:} Individual sector tuning using \gls{bo} optimizes lap time and lateral deviation but requires multiple uninterrupted laps for convergence, which might not be feasible during a racing scenario. 
The cost function was acquired averaging three consecutive laps, and ten iterations were run, for a total of 30 laps. The optimum was obtained with the parameters from the fifth iteration.
This method yields an average lap time of $\mu_{BO}=6.35s$ with a significantly lower lateral deviation of \SI{10.2}{\centi\metre}, about 1.5 times lower than uniform tuning, enhancing safety and consistency.
The trajectories and speed profile resulting from such a tuning procedure can be seen in \Cref{fig:tt_sector_tuning}.
\item \textbf{Hand-Tuned Scalers:} Manual adjustment based on racing expertise offers a balance between uniform and \gls{bo} tuning, achieving $\mu_{HT}=6.31s$ and a lateral deviation of \SI{13.24}{\centi\metre}, a compromise between performance and safety.
The trajectories and speed profile resulting from such tuning procedure can also be seen in \Cref{fig:tt_sector_tuning}, compared to the result from \gls{bo} tuning.
\end{enumerate}

Refer to \Cref{fig:tt_laptime} and \Cref{fig:tt_latdev} for a depiction of lap times and lateral deviations for these methods. 
% Note that manual driving, performed by a non-expert human driver using a joystick, is included for comparison, with higher lateral deviation not necessarily indicative of skilled human performance. 
This combined approach offers a comprehensive overview of each tuning strategy's impact on lap time and lateral deviation, facilitating a nuanced understanding of their effectiveness and suitability in various racing scenarios. \rev{The global planner's estimated lap time, assuming perfect tracking for the depicted map, is \SI{4.66}{\second}. However, it does not consider accurate traction limits, which can be incorporated with velocity scaling, as described in \Cref{subsec:gb_optimizer}. Adjusting for reduced traction using a uniform velocity scaler of 79\% yields an estimated ideal lap time of \SI{5.89}{\second}. The physical racecar achieved \SI{6.34}{\second}, indicating the autonomy stack performs within 7.1\% of the theoretical optimum, accounting for sensor noise, estimation errors, tracking inaccuracies, and further real-world imperfections.}

The performance in \emph{Time-Trials} was compared against other prevalent \emph{F1TENTH} racing architectures to provide context for the proposed \emph{ForzaETH Race Stack}:

\begin{enumerate}[I]
\item \textbf{MAP:} Uses the \emph{ForzaETH Race Stack} with the \gls{map} controller.
\item \textbf{PP:} Employs the same stack but replaces the \gls{map} controller with the widely used \gls{pp} controller \cite{purepursuit}, maintaining the same \emph{State Estimation, Opponent Estimation,} and \emph{Planning} setup.
\item \textbf{FTG:} Utilizes the \gls{ftg} controller \cite{ftg}, a reactive method that bypasses the \emph{See-Think-Act} cycle, responding directly to \gls{lidar} data by navigating through the largest perceived gap. While simpler and autonomous, it lacks the performance and adaptability of systems that adhere to the \emph{See-Think-Act} paradigm.
\end{enumerate}

\begin{figure}[ht]
    \centering
    \includegraphics[width=\textwidth]{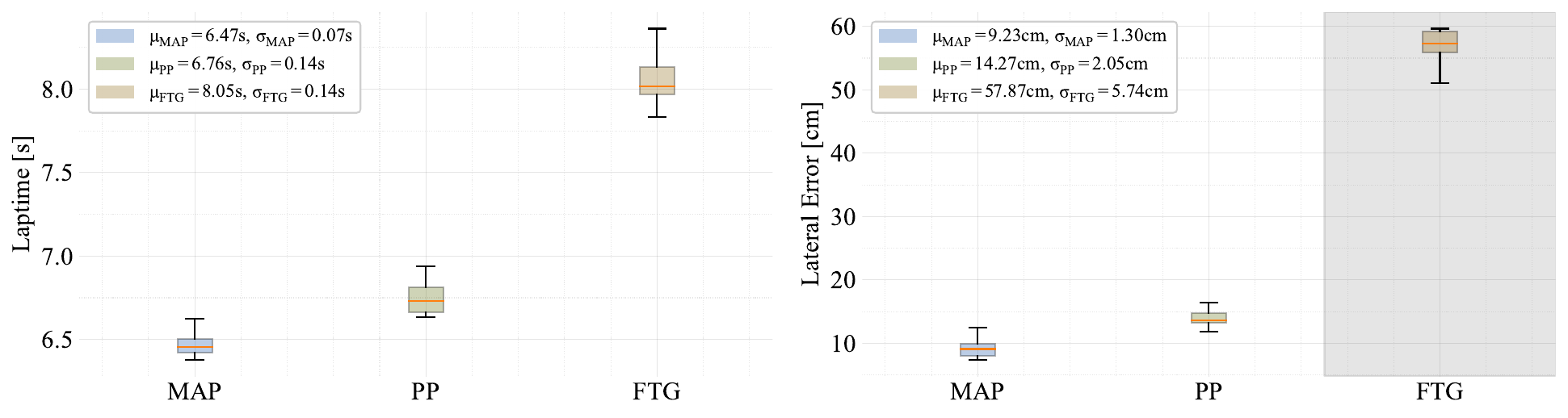}
    \includegraphics[width=\textwidth]{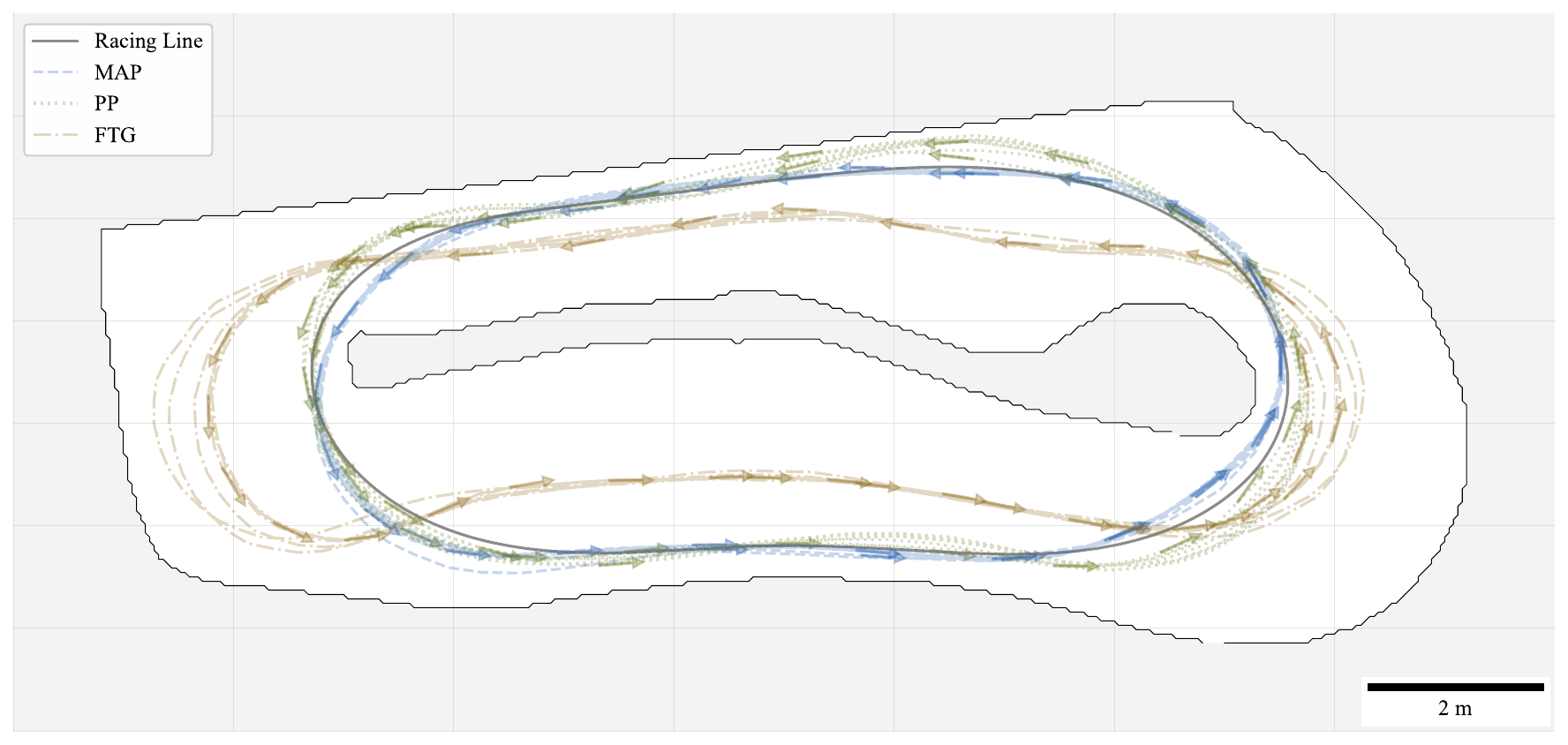}
    \caption{A lap time and lateral deviation comparison of the proposed \emph{ForzaETH Race Stack} configuration against a different controller configuration of the race stack and a completely reactive \gls{ftg} race stack. \gls{map} denotes the proposed configuration, \gls{pp} denotes a popular alternative for autonomous racing, and a fully reactive \gls{ftg} race stack that does not adhere to the \emph{See-Think-Act} cycle.}
    \label{fig:tt_laptimes_algo}
\end{figure}

\Cref{fig:tt_laptimes_algo} presents the lap times and lateral deviations relative to the racing line across ten laps for different race stack configurations. To ensure a fair comparison, the \gls{pp} and \gls{ftg} parameters were hand-tuned to achieve the lowest possible lap times. Both the \gls{map} and \gls{pp} versions of the \emph{ForzaETH Race Stack} employed a uniformly scaled racing line at $75\%$. The \gls{ftg}, being a reactive strategy, does not adhere to a predefined racing line. As a result, its lateral deviation is less relevant and is accordingly greyed out in the figure.

The \gls{map} controller within the proposed race stack demonstrates superior performance, achieving the most consistent and fastest lap time at $\mu_{MAP}=6.47s$ and a lateral deviation of \SI{9.23}{\centi\metre}. This represents a roughly $4.5\%$ speed increase and a $50\%$ reduction in lateral deviation compared to the \gls{pp} setup, showcasing that the \gls{map} configuration's enhanced precision and consistency in trajectory tracking at elevated speeds. In contrast to the \gls{ftg} system, the proposed stack offers a significant \SI{1.58}{\second} advantage in lap time, approximating a $24.4\%$ improvement. This, coupled with a $50\%$ lower standard deviation, demonstrates that it is a significantly faster and more consistent racing approach.

\subsubsection{Computation - Time-Trials} \label{ssubsec:tt_comp}
\begin{figure}[ht]
    \centering
    \subfloat[\emph{Time-Trials} Latency Histogram]{\includegraphics[width=0.46\textwidth]{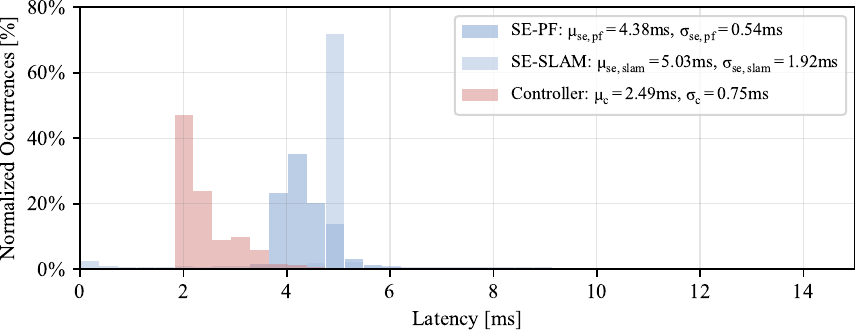}}
    \hfill
    \subfloat[\gls{cpu} Utilization Breakdown]{\includegraphics[width=0.52\textwidth]{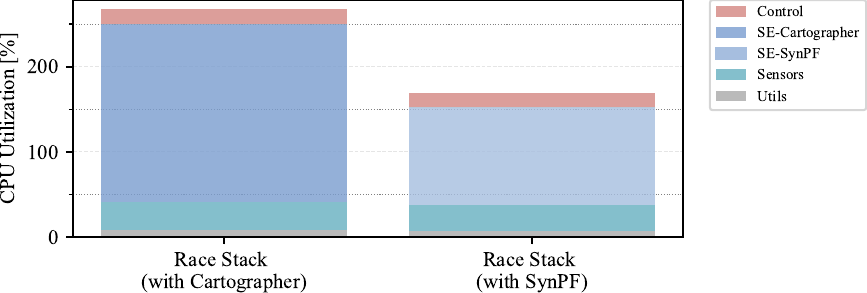}}
    \hfill
    \caption{Latency histogram and \gls{cpu} utilization breakdown of the \emph{ForzaETH Race Stack} during the \emph{Time-Trials} phase. Latencies of the \emph{Control} module are obtained from the run with \emph{Cartographer}-\gls{slam}.}
    \label{fig:tt_compute}
\end{figure}

\Cref{fig:tt_compute}, on the left-hand side, displays the latency histogram for the \emph{ForzaETH Race Stack} during the \emph{Time-Trials} phase. In this phase, \emph{Opponent Estimation} and \emph{Planning} modules are deactivated, under the presumption of an opponent-free track. Local planning is simplified to waypoint sampling from a pre-computed global racing line and can be neglected. Thus, \emph{State Estimation} and \emph{Control} are the primary contributors to the system's latency. The latency associated with state estimation is measured separately for \emph{Cartographer}-\gls{slam} and \emph{SynPF}, highlighting their distinct computational effects. The two different state estimation systems, however, are only used mutually exclusively. Under this setup, the race stack demonstrates a maximum average latency of \SI{7.5}{\milli\second}, when using \emph{Cartographer}-\gls{slam} and evaluated on the \texttt{Intel i5-10210U} \gls{obc}.

Further, \Cref{fig:tt_compute} on the right-hand side, showcases the \gls{cpu} usage of the race stack, categorized by autonomy modules, based on data captured via the \texttt{cpu\_monitor} library \cite{cpumonitor}, which internally uses the \texttt{psutil} process monitoring library. It encompasses various non-autonomy related \texttt{rosnodes} such as sensor drivers and dynamic reconfigure tools, categorized under \emph{Sensors} and \emph{Utils}. In this setup, the race stack's total \gls{cpu} utilization reaches 268.23\% or 169.41\% for either the \gls{slam} or the \gls{pf} based localization backbone, on an \texttt{Intel i5-10210U} \gls{obc} (which can reach a maximum of 800\% in terms of absolute \gls{cpu} load). The \emph{State Estimation} module is the primary computational load contributor, consuming 67.48\% or 55.34\% of the total racer stack \gls{cpu} utilization (relative to the race stack computation) with \emph{Cartographer}-\gls{slam} or \emph{SynPF}, respectively. In comparison, the \emph{Control} module's computational demand is relatively low, accounting for approximately 7\% of the total race stack utilization.

\FloatBarrier
\subsection{Head-to-Head Performance}\label{chap:res_h2h}
The \emph{Race Stack} was tested in a one-versus-one setup in the same racetrack as in \Cref{chap:res_tt}.
This assessment was used to understand under which circumstances the car can overtake an opponent driving a slower lap, and therefore understand when to enable the overtaking maneuvers. 
It is to be noticed, that in the setting of an \emph{F1TENTH} competition overtaking is not a necessary move to complete in order to win a \emph{Head-to-Head} race, as the rules state that the car that first completes a predefined number of laps (counted from the starting position) wins. 
Understanding the overtaking behavior of the car is however crucial in a future perspective, as multi-opponent racing --- a natural next step in the \emph{F1TENTH} competition --- will enforce overtakes for winning. 

In the experimental setup presented here, the ego-car used the nominal parameters described in the rest of this paper, exploiting the state machine described in \Cref{subsec:sw} and with only minor modifications to the single components, detailed as follows. 
As a low deviation from the reference line is needed to robustly track the overtaking trajectories, the selected sector scalers were uniformly set at 75\%, given the very close performance in terms of lap time (98\% average speed over a lap when compared to the fastest settings, from \Cref{fig:tt_laptime}) with the average lateral deviation below \SI{10}{\centi \metre} (from \Cref{fig:tt_latdev}).
Furthermore, the trailing distance was set to the minimum value of \SI{1.2}{\metre}, corresponding to the minimum recorded distance to the opponent of circa \SI{1}{\metre} from \Cref{sec:long_contr}, plus an additional \SI{20}{\centi \metre} to account for robustness.

\begin{figure}[ht]
    \centering
    \subfloat[Altered Racing Line \label{fig:rr_h2h_alt}]{\includegraphics[width=0.49\textwidth]{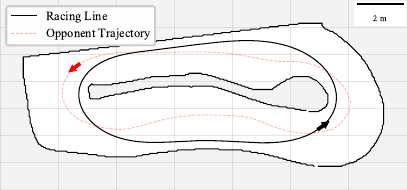}}
    \hfill
    \subfloat[\gls{ftg} racing line \label{fig:rr_h2h_ftg}]{\includegraphics[width=0.49\textwidth]{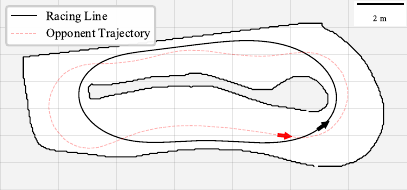}}
    \hfill
    \caption{Opponent Racing Lines for the \emph{Head-to-Head} analysis. The altered racing line was obtained by modifying the boundaries and then applying the minimum curvature racing line optimizer discussed in \Cref{subsec:gb_optimizer}, while the \gls{ftg} racing line is a result of the algorithm as described in \Cref{chap:res_tt}.}
    \label{fig:rr_h2h_trajectories}
\end{figure}

The opponent racecar was deployed with the same \emph{Head-to-Head} capable \emph{Race Stack} as the ego-car, with the only difference being the global scaler setup, which was set uniformly at lower values in order to match the lap time of the ego-car at lower percentages.
The chosen opponents were as follows:
\begin{enumerate}[I]
    \item \textbf{Racing Line Opponent}: The first opponent was chosen to be on the same racing line as ours. The velocity setups chosen corresponded to a speed over the lap 66\%, 73\%, and 80 \% lower when compared to the fastest lap time achieved in \Cref{chap:res_tt}.
    \item \textbf{Altered Racing Line Opponent}: The second opponent was on an altered racing line, which can be seen in \Cref{fig:rr_h2h_alt}.
    This different racing line was generated by manually altering the track boundaries and then running the global planner described in \Cref{subsec:gb_optimizer}.
    The velocity setups chosen corresponded to 66\%, 73\%, 80\%, and 86\% slower average velocity.
    \item \textbf{FTG Opponent}: The third and final opponent was set up to drive with an \gls{ftg} algorithm, and the trajectory can be seen in \Cref{fig:rr_h2h_ftg}. The chosen \gls{ftg} drove around at a speed comparable to circa 80\% that of the fastest speed achieved on the track.
\end{enumerate}

The two trajectories different from the nominal racing line can be seen in \Cref{fig:rr_h2h_trajectories}. The cars were set up in the same racing scenario as the official \emph{F1TENTH} \emph{Head-to-Head} format, where the vehicles are positioned at opposite sides of the track and the first car finishing ten laps is deemed the winner. 
The main analyzed parameter was the overtake success. We defined an overtake maneuver as successful if the faster car managed to overtake the slower car by the end of the ten laps. 
An overtake is not successful if all the tentatives fail by the end of the test laps, and this result was caused either by a collision with the opponent or by lack of a speed advantage. In case of collision, the cars were reset from a standstill at predetermined points, with the offending car (the one causing the collision) situated behind the other. 
A further result reported is the eventual race outcome of the one-versus-one; it is however to be noted that since we tested with slower opponents the ego-car ended up winning all the runs --- necessary with this algorithm, as without speed advantage it is not possible to overtake. The results of the overtaking experiments are shown in \Cref{tab:race_res_hh}.

\begin{table}[ht]
\centering
\begin{tabular}{c|c|c|c}
\textbf{Opponent Type} &  \textbf{{[}\%{]}} & \textbf{Overtake Completed} & \textbf{Win} \\ 
\hline \hline
                       & 66                 & \textbf{yes}                & \textbf{yes} \\ \cline{2-4} 
Same Racing Line       & 73                 & \textbf{yes}                & \textbf{yes} \\ \cline{2-4} 
                       & 80                 & no \textdaggerdbl                          & \textbf{yes} \\ \hline
                       & 66                 & \textbf{yes}                & \textbf{yes} \\ \cline{2-4} 
Altered Racing Line    & 73                 & \textbf{yes}                & \textbf{yes} \\ \cline{2-4} 
                       & 80                 & \textbf{yes}                & \textbf{yes} \\ \cline{2-4} 
                       & 86                 & no \textdagger                          & \textbf{yes} \\ \hline
\gls{ftg}              & 79                 & \textbf{yes}                & \textbf{yes} \\
\end{tabular}%
\caption{Overtaking results of the full \emph{Head-to-Head} system against opponents at slower velocity. \textdagger: the overtake was unsuccessful due to a collision. \textdaggerdbl: the overtake was unsuccessful due to a lack of speed advantage.}
\label{tab:race_res_hh}
\end{table}

While it is clear that for slower opponents (66\% and 73\% average velocity) the overtaking procedure is carried out successfully, for faster opponents the ability to overtake is dependent on the racing line. 
For the altered racing line and the \gls{ftg} opponent, a setup with higher velocity can still be overtaken, as the velocity advantage in the corner allows for the maneuver. For the opponent on the same racing line, instead, the overtake is carried out with a straight-line speed advantage, which in the case of this specific track is not enough to achieve enough advantage to safely complete an overtake when the opponent is set at a velocity approximately 80\% of the ego-car.

While disabling overtake and setting the trailing gap at a safe \SI{2}{\metre} length still remains a dominant strategy that would allow winning every \emph{Head-to-Head} confrontation with an opponent at a slower average velocity than the ego agent, assessing the overtake capacities remains important.
In a setting where overtaking is mandatory, the evaluation carried out here can help in deciding how to set the vehicle's parameters. In case the opponent keeps a velocity comparable to 70\% of ours, the overtaking strategy can be deployed effectively, while, in case of a faster opponent, a clear section with a speed advantage should be first identified to ensure that an overtake is feasible, and with any opponent above 80\% of our average velocity, overtake should be avoided. 

\subsubsection{Computation - Head-to-Head} \label{ssubsec:h2h_comp}
\begin{figure}[ht]
    \centering
    \subfloat[\emph{Head-to-Head} Latency Histogram]{\includegraphics[width=0.47\textwidth]{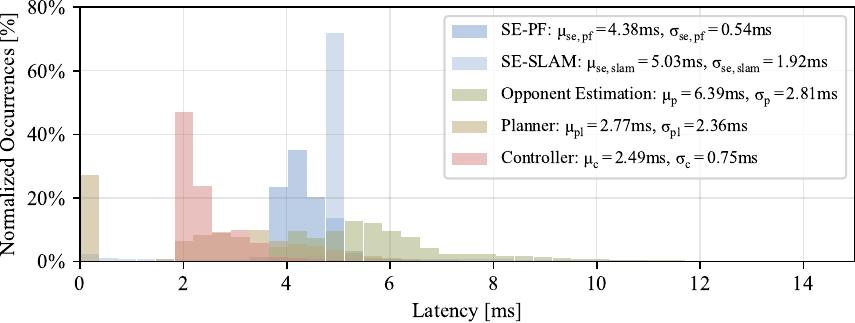}}
    \hfill
    \subfloat[\gls{cpu} Utilization Breakdown]{\includegraphics[width=0.51\textwidth]{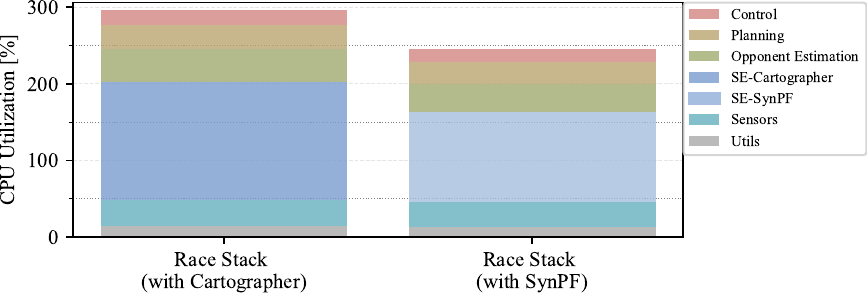}}
    \hfill
    \caption{Latency histogram and \gls{cpu} utilization breakdown of the \emph{ForzaETH Race Stack} during the \emph{Head-to-Head} phase.Latencies of the \emph{Opponent Estimation}, \emph{Planning}, and \emph{Control} module are obtained from the run with \emph{Cartographer}-\gls{slam}.}
    \label{fig:h2h_compute}
\end{figure}

\Cref{fig:h2h_compute} displays the latency histogram  of the \emph{Head-to-Head} normalized to the total occurrences. Here all autonomy modules within the \emph{See-Think-Act} cycle are active and the latency histogram gives insights into the robot's operation. The highest average latency can be seen from the \emph{Opponent Estimation} module with \SI{6.39}{m\second}. Interestingly to note, is the distribution shape of the \emph{Planning} latency, as there are two noticeable peaks, one around \SI{0}{\milli\second} and then one after \SI{3}{\milli\second}. This is due to the fact that the local planner is only active if an opponent is present in front of the car.

During the \emph{Head-to-Head} phase, all autonomy modules are involved and thus contribute towards the total computation, which amounts to 295.96\% or 245.48\% \texttt{psutil} \gls{cpu} utilization on the\texttt{Intel i5-10210U} \gls{obc}, respectively for a \gls{slam} or \gls{pf} based \emph{State-Estimation} backbone. For the \gls{slam} based localization system, relative to the total compute, \emph{State-Estimation} accounts for 51.97\%, \emph{Opponent Estimation} for 14.45\%, \emph{Planning} for 10.51\%, and \emph{Control} for 6.66\% of the entire race stack. The remainder is used for utility functionalities and sensor drivers. In the \gls{pf} based system, the relative compute utilization is nearly identical as in the aforementioned \gls{slam} scenario, albeit of course with a lower total consumption.

\FloatBarrier
\section{Lessons Learned}
Throughout the development of the \emph{ForzaETH Race Stack}, many learnings could be drawn that improve the overall performance of the race stack. In this chapter, we distill the key lessons learned throughout this process. 
These lessons span various aspects from the development process to technical decision-making and have played pivotal roles in shaping the stack's performance and reliability. By sharing these insights, we hope to provide valuable guidance for similar endeavors in the field of autonomous racing, aiding in the advancement of technology and operational strategies.

\begin{enumerate} [I]
    \item \textbf{Upstream Affects Downstream:}  The system architecture depicted in \Cref{fig:sys_architecture} demonstrates the interdependent nature of autonomy modules within the \emph{See-Think-Act} cycle. The performance of upstream modules, particularly \emph{State Estimation}, is critical as it lays the foundation for subsequent processing stages. For instance, inaccuracies in \emph{State Estimation} will propagate errors into the \emph{Control} module, regardless of the controller's capabilities. This interdependence necessitates a holistic approach to the development and iterative refinement of all autonomy modules to enhance the overall system performance.

    \item \textbf{Balance Development Simplicity with Latency:} While a lower code execution latency is always desired, it is essential to strike a balance between development simplicity and performance. Rapid prototyping with \texttt{Python} can identify where low latency is critical and if deemed necessary \texttt{C++} for performance optimization can be used. This approach helps in efficiently allocating development resources and ensures that efforts are directed towards modules where speed is crucial, without unnecessarily complicating the system and the code maintenance.
    
    \item \textbf{Operational Simplicity:} The dynamic and unpredictable nature of real-world racing events demands an autonomy stack designed for operational robustness. Unanticipated variables, such as variations in surface traction or environmental conditions affecting sensor performance, must be accounted for. The stack should facilitate straightforward parameter adjustments, providing operators with the ability to swiftly adapt to changes in the racing environment. 
    % A transparent system, as opposed to a black-box approach, is essential for rapid adaptation and informed decision-making through key parameters in autonomous racing scenarios.
    A transparent system with tunable knobs that produce predictable changes is essential for rapid adaptation and informed decision-making through key parameters in autonomous racing scenarios.
    
    \item \textbf{Importance of Testing:} To build upon the previous point, figuring out the key parameters that yield full operational control of the robot, is only possible through thorough testing of the system. Testing is not only vital for obtaining technical insights but also for training the operational handling of the pit staff. Therefore, fully rehearsing the race can allow the operators to spot technical, as well as operational flaws, which are invaluable information to increase performance during the development phase as well as the race itself. Leverage the simulation environment for as much as possible, but evaluate racing performance only on the physical platform, to not be misled by the \emph{Sim-to-Real} gap.  
    
    \item \textbf{Team Spirit:} 
    % A robot is only as good as the humans that created it. 
    As this project has built upon many different student theses throughout their academic studies, we believe that upholding team spirit and passion for autonomous racing is a key enabler to achieving strong performance.
\end{enumerate}

\section{Conclusion and Future Work}
This work presents the \emph{ForzaETH Race Stack}. The presented system is designed for \gls{cots} hardware as proposed by the official \emph{F1TENTH} competition. The system is fit for highly competitive autonomous racing in both a \emph{Time-Trials} setting and \emph{Head-to-Head} races. In this paper, we describe the essential \rev{autonomy modules}\revdel{ which are \emph{Perception}, \emph{Planning}, and \emph{Control}. 
We } \rev{and} detail the subsystems of each component, giving insights on how to configure those subsystems including information regarding the performance that can be expected from each system. 
The system has been extensively tested under various track conditions\rev{, specifically at speeds over \SI{10}{\metre \per \second} and on tracks longer than \SI{140}{\metre},} and has proven race-winning in official \emph{F1TENTH} competitions.
\rev{Additionally, the \emph{ForzaETH RaceStack} has proven a localization accuracy below \SI{10}{\centi \metre} and provides two different state estimation pipelines, to be used depending on the quality of the wheel odometry signal. 
It has proven to be able to detect opponent vehicles with an \gls{rmse} of \SI{17}{\centi \metre} spatially and \SI{0.49}{\metre \per \second} velocity-wise. 
This in turn enables the system to trail opponents at distances of \SI{1}{\metre} or larger and to overtake opponents at up to $80\%$ of the ego velocity.}

\begin{table}[th]
    \centering
    \begin{tabular}{p{0.26\textwidth - 2\tabcolsep}p{0.16\textwidth - 2\tabcolsep}p{0.18\textwidth - 2\tabcolsep}p{0.16\textwidth - 2\tabcolsep}p{0.18\textwidth - 2\tabcolsep}}
        \rowcolor{darkgray}
        \textbf{Source} & Open Source & \gls{cots} Hardware & \emph{Time-Trials} &  \emph{Head-to-Head} \\
        \cite{amz_fullsystem} & No & No & \textbf{Yes} & No\\
        \rowcolor{lightgray}
        \cite{raji2023erautopilot} & No & No & \textbf{Yes} & \textbf{Yes}\\
        \cite{kaist_fullsystem} & No & No & \textbf{Yes} & \textbf{Yes}\\
        \rowcolor{lightgray}
        \cite{tum_fullsystem} & No & No & \textbf{Yes} & \textbf{Yes}\\
        \;\textbf{Ours} & \textbf{Yes} & \textbf{Yes} & \textbf{Yes} & \textbf{Yes}\\
    \end{tabular}
    \caption{Comparison summary of the proposed \emph{ForzaETH Race Stack} compared to other full-stack implementations. The term open source refers specifically to the code availability of the complete stack.}
    \label{tab:concl}
\end{table}

\rev{As highlighted in \Cref{tab:concl},} the primary goal of this paper is to provide the first complete, robust, and accessible autonomy stack on \gls{cots} hardware for autonomous racing and research communities. By offering a race-proven system, we lower barriers to entry, enabling teams and researchers to focus on innovation and performance enhancement without the need for extensive resource investment. This approach enables inclusive access to advanced autonomous racing technologies, for growth and experimentation in the field.

\rev{When considering future work, the various components of the \emph{ForzaETH Race Stack} provide a foundation for the integration of new technologies, each presenting unique challenges. For instance, in the \emph{Perception} domain, future efforts could draw from \gls{sota}, large-scale \gls{ml} approaches like those in \cite{centerpoint, qd_track}. The limited hardware capacity of \emph{F1TENTH} would intensify these computational challenges, necessitating creative solutions. In \emph{Planning}, incorporating opponent behavior and modeling interactions between agents, as discussed in \cite{gametheory_liniger}, could be explored next; however, computational limitations remain a significant hurdle. Finally, in \emph{Control}, advanced methods such as \gls{mpc} from \cite{hierarchical_mpc} could be adopted, where ensuring minimal model mismatch would add an additional layer of complexity, especially given computational constraints.
}

\revdel{Moving forward,}\rev{It is clear, that not only the \emph{Time-Trials} racing domain, but also} the \emph{Head-to-Head}\revdel{racing} domain offer expansive prospects for innovation. The \emph{ForzaETH Race Stack} lays a robust foundation for delving into multi-opponent racing dynamics, a promising research field poised to yield significant advancements in autonomous racing strategies and technologies. Additionally, exploring the scalability of this stack for full-scale autonomous vehicles presents an interesting topic for future research, potentially broadening its applicability and impact in the domain of general autonomous systems.

\subsection*{Acknowledgements}
\rev{The authors would like to thank Prof. Dr. Frazzoli and the entire autonomous Go-Kart team of the \gls{idsc} laboratory at ETH Zurich for allowing the usage of their large testing facility and the fruitful discussion of various algorithms.}

%TODO: ADD BIO back in afterward
%\input{chapters/biography}

%Refs
\bibliographystyle{apalike}
\bibliography{main}

%Appendix
\appendix
\newpage
\section{Appendix} 
\subsection{Parameters}
\label{sec:app_params}
This section aims to supply the reader with relevant hyperparameters used in each autonomy module.

\subsubsection{State Estimation} \label{sec:app_SE}
\begin{enumerate}[I]
    \item \textbf{Localization Parameters:}
    
    Both the \emph{Cartographer} and \emph{SynPF} algorithms were tuned to minimize the lap time in a \emph{Time-Trials} scenario over multiple maps. The parameters are shown as \verb|[min, max, tuned]|:

    \begin{enumerate}
        \item \textbf{Cartographer:}
        \emph{Cartographer} parameters are identical for the mapping and race stages of competition. However, at race-time, \emph{Cartographer} is run in \emph{localization-only} mode, which runs with lower latency and compute usage. It then uses the map of the racetrack obtained from the mapping stage.

        \begin{itemize}
            \item \verb|POSE_GRAPH.optimization_problem.odometry_rotation_weight| \texttt{[0, $\infty$, 0]} - Weight of rotating incoming LiDAR scans away from the predicted pose given by odometry input in the scan-matching optimization problem.
            \item \verb|POSE_GRAPH.optimization_problem.odometry_translation_weight| \texttt{[0, $\infty$, 0]} - Weight of translating incoming LiDAR scans away from the predicted pose given by odometry input in the scan-matching optimization problem.
            \item \verb|TRAJECTORY_BUILDER_2D.ceres_scan_matcher.rotation_weight| \texttt{[0, $\infty$, 8]} - Cost of rotational deviation from the prior position in the scan matcher, with higher values increasing the penalty for rotational discrepancies.
            \item \verb|TRAJECTORY_BUILDER_2D.ceres_scan_matcher.translation_weight| \texttt{[0, $\infty$, 5]} - Cost of translational deviation from the prior position in the scan matcher, with higher values increasing the penalty for translational discrepancies.
        \end{itemize}
    
        \item \textbf{SynPF}
        These parameters affect the variance of noise applied to the \emph{SynPF} motion model and correspond to those in \cite{synpf}.
    
        \begin{itemize}
            \item \texttt{$\alpha_{1}$} \verb|[0.0, 1.0, 0.5]| - how much rotation affects rotation variance.
            \item \texttt{$\alpha_{2}$} \verb|[0.0, 0.05, 0.01]| - how much translation affects rotation variance.
            \item \texttt{$\alpha_{3}$} \verb|[0.0, 1.0, 0.1]| - how much translation affects translation variance.
            \item \texttt{$\alpha_{4}$} \verb|[0.0, 5.0, 1.0]| - how much rotation affects translation variance.
            \item \texttt{lam\_thresh} \verb|[0.01, 0.2, 0.1]| - minimum translation for the TUM model to become effective.
        \end{itemize}
    
    \end{enumerate}

    \item \textbf{Extended Kalman Filter Parameters:}
    The parameters relevant to the \gls{ekf}, responsible for generating the velocity estimates, can in this case be subdivided into two categories: the covariances $\sigma_{i}^2 \in [0, \infty)$ associated with observations (zero elements are handled by the \verb|robot_localization| package) and the configuration parameters which dictate what sources are to be considered by the \gls{ekf}. %The covariances have been chosen to be static, listed below, and the configuration parameters have been determined experimentally and can be easily modified. Any additional \gls{ekf} parameters, such as the update frequency, have been chosen to comply with the rest of the race-stack, and should therefore only be changed in specific circumstances. \\
    
    The \verb|robot_localization| sensor fusion \gls{ekf} configuration parameters are defined according to \cite{robotlocalization} as:
    
    $\verb|config| =
        \begin{bmatrix}
            x & y & z \\
            roll & pitch & yaw \\
            vx & vy & vz \\
            vroll & vpitch & vyaw \\
            ax & ay & az \\
        \end{bmatrix}$

    where each entry is of type boolean, indicating whether or not the associated data source should be considered by the filter.

    The \gls{ekf} parameters intended for tuning are summarized as follows:

    \begin{itemize}
        \item $[\sigma_{VESC,x}, \sigma_{VESC,y}, \sigma_{VESC,\psi}] = [0.25, 0.5, 0.4]$ - the covariances associated with \gls{vesc} control odometry x position, y position, and yaw angle, respectively.
        \item $[\sigma_{VESC,vx}, \sigma_{VESC,vy}, \sigma_{VESC,vaz}] = [0.02, 0.05, 0.0]$ - the covariances associated with \gls{vesc} control odometry longitudinal, lateral, and angular yaw velocities, respectively.
        \item $[\sigma_{IMU,al}, \sigma_{IMU,va}, \sigma_{IMU,q}] = [0.0,0.0,0.0]$ - the covariances associated with \gls{imu} linear acceleration, angular velocity, and orientation measurements, respectively.
        \item The \gls{ekf} configuration matrix for the \gls{vesc} control odometry is defined as: \\
            $\verb|config_odom| = 
                \begin{bmatrix}
                    \verb|false| & \verb|false| & \verb|false| \\
                    \verb|false| & \verb|false| & \verb|false| \\
                    \verb|true| & \verb|true| & \verb|false| \\
                    \verb|false| & \verb|false| & \verb|false| \\
                    \verb|false| & \verb|false| & \verb|false| \\
                \end{bmatrix}$
        \item The \gls{ekf} configuration matrix for the \gls{imu} is defined as: \\
            $\verb|config_imu| = 
                \begin{bmatrix}
                    \verb|false| & \verb|false| & \verb|false| \\
                    \verb|false| & \verb|false| & \verb|true| \\
                    \verb|false| & \verb|false| & \verb|false| \\
                    \verb|false| & \verb|false| & \verb|true| \\
                    \verb|false| & \verb|false| & \verb|false| \\
                \end{bmatrix}$
    \end{itemize}
    
\end{enumerate}

\subsubsection{Opponent Estimation} \label{sev:app_perception}
% \todo{@Tobias come up with the key perception params here.}
\begin{enumerate}[I]
    \item \textbf{Detection:}
    The parameters for the detection algorithm that were used to achieve the documented results are shown below as \verb|[used, minimum, maximum]|:
    \begin{itemize}
        \item \verb|min_obs_size [40, 5, 300]| - Minimum number of cloud points of an obstacle
        \item \verb|max_obs_size [0.5, 0.1, 1]| - Maximum size of an obstacle in meters
        \item \verb|max_viewing distance [9, 3, 10]| - Maximal reliable distance of LiDAR measurements in meters
    \end{itemize}
    \item \textbf{Tracking:}
    The parameters for the tracking algorithm that were used to achieve the documented results are shown below:
    \begin{itemize}
        \item The process Gaussian noise is defined as $Q =
            \begin{bmatrix}
                Q_1 & 0 \\
                0 & Q_2 \\
            \end{bmatrix}$ where $Q_1 =
            \begin{bmatrix}
                1.95 \cdot 10^{-7} & 1.56 \cdot 10^{-5}\\
                1.56 \cdot 10^{-5} & 1.25 \cdot 10^{-3}\\
            \end{bmatrix}$ and $Q_2 =
            \begin{bmatrix}
                7.81 \cdot 10^{-7} & 6.25 \cdot 10^{-5}\\
                6.25 \cdot 10^{-5} & 5 \cdot 10^{-3}\\
            \end{bmatrix}$
        \item The input Gaussian noise is defined as $R =
            \begin{bmatrix}
                0.002 & 0 & 0 & 0\\
                0 & 0.2 & 0 & 0  \\
                0 & 0 & 0.002 & 0\\
                0 & 0 & 0 & 0.2  \\
            \end{bmatrix}$
        \item Proportional gains for Input: $P_{v_s} = 0.2$, $P_d = 0.02$, $P_{v_d} = 0.2$
        \item Target speed $v_{s,target} = v_{s,ego}(s) \cdot r$, with $v_{s,ego}(s)$ being the car's speed at the opponent $s$ position and $r$ a configurable ratio of this speed, set to $0.6$
    \end{itemize}
\end{enumerate}

\subsubsection{Planning} \label{sec:app_planning}
\begin{enumerate}[I]
    \item \textbf{Global Planner:}
    The parameters for the global planner that were used to achieve the documented racing line are shown below as \verb|[minimum, maximum, tuned]|:
    \begin{itemize}
        \item \verb|curvlim [0, |$\infty$\verb|, 1.0]| - Maximum curvature of the vehicle in radians per meter
        \item \verb|iqp_curverror_allowed [0, |$\infty$\verb|, 0.1]| - Maximum curvature error in radians per meter allowed between the curvature of the optimized path and \verb|curvlim| 
        \item \verb|width_opt [0, |$\infty$\verb|, 0.8]| - Vehicle width in meters including a safety distance
        \item \verb|stepsize_reg [0, |$\infty$\verb|, 0.2]| - Distance in meters between two points on the reference line during optimization
    \end{itemize}

    \item \textbf{Local Planner:}
    \begin{itemize}
    \item $n^{spline} = 3$ - number of $preapex/postapex$ points before and after the opponent's position, selected on the racing line to construct the overtaking spline. 
    \item $(\Delta^{preapex}_1,\,\Delta^{preapex}_2,\,\Delta^{preapex}_3) = (2,\,3,\,4)$ - baseline distances of the $preapex$ points from the opponent's $d$ position.
    \item $(\Delta^{postapex}_1,\,\Delta^{postapex}_2,\,\Delta^{postapex}_3) = (4.5,\,5,\,5.5)$ - baseline distances of the $postapex$ points from the opponent's $d$ position.
    \item $\delta^{apex} = 0.4$ - extra lateral distance of the overtaking apex, to account for the controller's imperfections in tracking the reference spline.
    \end{itemize}
\end{enumerate}

\subsubsection{Control}\label{sec:app_controls}
The controller parameters that were used to achieve the documented results are shown below as \verb|[minimum, maximum, tuned]|:

\begin{enumerate}[I]
    \item \textbf{Longitudinal Controller:}
    
    \begin{itemize}
        \item $t_{la}$ \verb|[0, |$\infty$\verb|, 0.25]| - Lookahead time in seconds to account for actuation and computation delay.
        \item $\lambda_{lat}$ \verb|[0, 1, 1]| - How much of the lateral error is taken into account to smoothly rejoin the trajectory. Higher values increase the dependence of the lateral error on the speed reduction.
        \item $k_p$ \verb|[0, |$\infty$\verb|, 1]| - Proportional gain for the error term $e_{gap}$ in the calculation of $v_{des}$ in the \emph{Trailing} state.
        \item $k_d$ \verb|[0, |$\infty$\verb|, 0.2]| - Gain for the derivative term in the calculation of $v_{des}$ in the \emph{Trailing} state.
        \item $v_{blind}$ \verb|[0, |$\infty$\verb|, 1.5]| - Minimum velocity in \SI{}{\metre / \second} in the case when there is no \gls{los} of an obstacle.
        \item $g_{tar}$ \verb|[0, |$\infty$\verb|, 2]| - Target gap in \SI{}{\metre} to the to-be-trailed opponent.
    \end{itemize}

    \item \textbf{Lateral Controller:}
    \begin{itemize}
        % \item $t_{la}$ \verb|[0, |$\infty$\verb|, 0.25]| - Lookahead time in seconds to account for actuation and computation delay.
        \item $m$ \verb|[0, |$\infty$\verb|, 0.6]| - Proportional term for the affine mapping of the velocity to the lookahead distance for the \gls{map} controller.
        \item $q$ \verb|[|$-\infty$\verb|,| $\infty$\verb|, -0.18]| - Offset term for the affine mapping of the velocity to the lookahead distance.
    \end{itemize}
\end{enumerate}

\subsection{Extended Kalman Filter Definition of Odometry Filter}
\label{sec:ekf_filter_definition}
The \gls{ekf} model used for the odometry filter consists of an omnidirectional, three-dimensional, point-mass motion model. The state $X$ and the discrete-time transfer function $f(X)$ used in the library are defined in the following equations:
\allowdisplaybreaks
\begin{align*}
    X &= \left[
        x,\,
        y,\,
        z,\,
        \phi ,\,
        \theta ,\,
        \psi ,\,
        \dot{x} ,\,
        \dot{y} ,\,
        \dot{z} ,\,
        \dot{\phi} ,\,
        \dot{\theta} ,\,
        \dot{\psi} ,\,
        \ddot{x} ,\,
        \ddot{y} ,\,
        \ddot{z}
        \right]^\top
        \\
    X^+ &= f(X) = \left[
        x^+,\,
        y^+,\,
        \hdots,\,
        \ddot{z}^+
        \right]^\top
        \\
    x^+ &= x + \left(\dot{x}\cos{\psi}\cos{\theta}+ \dot{y}\left(\cos{\psi}\sin{\theta}\sin{\phi} - \sin{\psi}\cos{\phi}\right)+ \dot{z}\left(\cos{\psi}\sin{\theta}\cos{\phi} + \sin{\psi}\sin{\phi}\right)\right)\Delta_t \\ 
    &\qquad+ \frac{1}{2}\left(\ddot{x}\cos{\psi}\cos{\theta} + \ddot{y}\left(\cos{\psi}\sin{\theta}\sin{\phi} - \sin{\psi}\cos{\phi}\right) + \ddot{z}\left(\cos{\psi}\sin{\theta}\cos{\phi} + \sin{\psi}\sin{\phi}\right)\right)\Delta_t^2 \\
    y^+ &= y + \left(\dot{x}\sin{\psi}\cos{\theta} + \dot{y}\left(\sin{\psi}\sin{\theta}\sin{\phi} + \cos{\psi}\cos{\phi}\right) + \dot{z}\left(\sin{\psi}\sin{\theta}\cos{\phi} - \cos{\psi}\sin{\phi}\right)\right)\Delta_t\\
    &\qquad +\frac{1}{2}\left(\ddot{x}\sin{\psi}\cos{\theta} + \frac{1}{2}\ddot{y}\left(\sin{\psi}\sin{\theta}\sin{\phi} + \cos{\psi}\cos{\phi}\right) + \frac{1}{2}\ddot{z}\left(\sin{\psi}\sin{\theta}\cos{\phi} - \cos{\psi}\sin{\phi}\right)\right)\Delta_t^2 \\
    z^+ &= z + \left(-\dot{x}\sin{\theta} + \dot{y}\cos{\theta}\sin{\phi} + \dot{z}\cos{\theta}\cos{\phi}\right)\Delta_t + \frac{1}{2}\left(-\ddot{x}\sin{\theta} + \ddot{y}\cos{\theta}\sin{\phi} + \ddot{z}\cos{\theta}\cos{\phi}\right)\Delta_t^2 \\
    \phi^+ &= \phi + \left(\dot{\phi} + \dot{\theta}\sin{\phi}\tan{\theta} + \dot{\psi}\cos{\phi}\tan{\theta}\right)\Delta_t \\
    \theta^+ &= \theta + \left(\dot{\theta}\cos{\phi} - \dot{\psi}\sin{\phi}\right)\Delta_t \\
    \psi^+ &= \psi + \left(\dot{\theta}\frac{\sin{\phi}}{\cos{\theta}} + \dot{\psi}\frac{\cos{\phi}}{\cos{\theta}}\right)\Delta_t \\
    \dot{x}^+ &= \dot{x} + \ddot{x}\Delta_t \\
    \dot{y}^+ &= \dot{y} + \ddot{y}\Delta_t \\
    \dot{z}^+ &= \dot{z} + \ddot{z}\Delta_t \\
\end{align*}
where $(x,\,y,\,z)$ represent the position, $(\phi,\,\theta,\,\psi)$ represent the orientation as roll, pitch, yaw respectively, and, for ease of notation, the superscript $^+$ represents the next-timestep state, e.g. $X^+ \coloneqq X[k+1] = f(X[k])$. Furthermore, the next-states not present in the equation are assumed to be constant.

\subsection{Race Results}
\label{sec:app_race_results}
By the end of 2023, the \emph{ForzaETH} team did participate in three of the official \emph{F1TENTH} competitions. The first participation was at the \emph{ICRA Grand-Prix 2022} in Philadelphia where the team was able to reach fourth place. With the learnings of the first competition, the team was able to further improve the \emph{Race Stack} and reach first place at the \emph{German Grand-Prix 2022} which was held next to the Lausitzring (full-scale) race track. At the \emph{ICRA Grand-Prix 2023} in London, the \emph{ForzaETH} team was able to confirm its first place against a broader and more international range of opponents. \Cref{tab:grand_prix_facts_figures} shows the official \emph{F1TENTH} race rankings that the proposed race stack has been able to achieve --- keeping in mind, that the stack improved throughout time.

The \emph{Race Stack} presented in this paper is for most parts the stack that was used in the \emph{ICRA Grand-Prix 2023}. The subsequent subsections describe in greater detail how the \emph{Race Stack} configurations perform for both the \emph{Time-Trials} and \emph{Head-to-Head} phases of the competitions with performance evaluations for various test tracks, as well as the achieved race results of the \emph{ForzaETH} team.

\begin{table}[ht]
    \centering
    \begin{tabular}{l|l|l|c|c}
        \textbf{Competition} & \textbf{Year} & \textbf{Venue} & \textbf{\# Teams} & \textbf{Ranking}  \\
        \hline
        \hline
        ICRA Grand-Prix & 2022 & Philadelphia, PA, USA & 20 & 4  \\
        German Grand-Prix & 2022 & Lausitzring, Germany & 6 & \textbf{1}  \\
        ICRA Grand-Prix & 2023 & London, UK & 22 & \textbf{1} \\
    \end{tabular}
    \caption{Overview of the \emph{F1TENTH} competitions where the \emph{ForzaETH} team did compete. Note that at \emph{ICRA Grand-Prix 2022} the team did start under the name \emph{ForzaPBL}.}
    \label{tab:grand_prix_facts_figures}
\end{table}

\subsubsection*{Competition Results}
\begin{comment}
\begin{table}[h]
    \centering
    \begin{tabular}{l|c|c|c|c|c|c}
        Team Name & Heat & \# Cons. & Fastest & \multicolumn{3}{c}{Score} \\
         & Nr. & Laps $\uparrow$ & Lap [s] $\downarrow$ & Cons. Laps $\uparrow$ & Laptime $\uparrow$ & Total $\uparrow$ \\
        \hline
        ForzaETH & 2 & 41 & 7.00 & 12 & 12 & 24 \\
        ForzaETH & 1 & 41 & 7.05 & 12 & 11 & 23 \\
        HiPeRT Modena & 2 & 38 & 7.62 & 10 & 8 & 18 \\
        HiPeRT Thundershot & 2 & 36 & 8.15 & 9 & 7 & 16 \\
        HiPeRT Modena & 1 & 12 & 7.50 & 5 & 9 & 14 \\
        HiPeRT Thundershot & 1 & 10 & 7.46 & 4 & 10 & 14 \\
        Krizaly & 1 & 26 & 10.90 & 8 & 3 & 11 \\
        PUT-PPI & 1 & 18 & 9.95 & 7 & 4 & 11 \\
        PUT-PPI & 2 & 16 & 9.29 & 6 & 5 & 11 \\
        Krizaly & 2 & 3 & 8.58 & 3 & 6 & 9 \\
        Dzik Team & 1 & - & - & 1 & 1 & 2 \\
        Dzik Team & 2 & - & - & 1 & 1 & 2 \\
    \end{tabular}
    \caption{Time trials results of the \emph{Germany Grand-Prix 2022}.}
    \label{tab:results_germany22_time_trials}
\end{table}

\begin{table}[h]
    \centering
    \begin{tabular}{c|c||c|c}
        \multicolumn{2}{c||}{Semi Final} & \multicolumn{2}{c}{Final} \\
        \hline
        ForzaETH & 2 & ForzaETH & 2 \\
        PUT-PPI  & 0 & Dzik Team & 1
    \end{tabular}
    \caption{Caption}
    \label{tab:results_germany22_h2h}
\end{table}
\end{comment}

In the following, we are presenting the results of the most recent competition \emph{ForzaETH} competed in, the \emph{ICRA Grand-Prix 2023}. As previously mentioned the \emph{Grand-Prix} style competitions consist of two phases, a \emph{Time-Trials} phase and a \emph{Head-to-Head} phase. 

In the \emph{Time-Trials} phase the teams have a predefined time window (typically 5 minutes) to achieve two goals. Firstly the teams need to achieve the fastest lap time and secondly, they need to reach the highest possible number of consecutive (uninterrupted) laps. The teams get two attempts (heats) to reach these goals. Only the better heat where both the lap time and number of consecutive laps are considered is used for the ranking. The ranking of the top 10 teams of the \emph{Time-Trials} phase at \emph{ICRA Grand-Prix 2023} is given in \Cref{tab:results_icra23_time_trials}. The results show that both a fast and robust system are required to reach first place. As can be seen from \Cref{tab:results_icra23_time_trials} most teams perform well in one of the two criteria, e.g. \emph{HiPeRT Modena} reached second place in the number of consecutive laps, but only fourth place in the lap time and even more strikingly \emph{Scuderia Segfault} reached second place in the lap time, but only eleventh place regarding the consecutive laps. Hence to score well in \emph{F1TENTH}, one's car needs to both be fast and follow a trajectory consistently.

\begin{table}[ht]
    \centering
    \begin{tabular}{l|c|c|c|cc|cc|c|c}
        Team Name & Heat & \# Cons. & Fastest & \multicolumn{5}{c|}{Score} & Qualifying \\
         & Nr. & Laps $\uparrow$ & Lap [s] $\downarrow$ & \multicolumn{2}{c}{Cons. Laps $\uparrow$} & \multicolumn{2}{c}{Laptime $\uparrow$} & Total $\uparrow$ & Rank \\
        \hline
        \hline
        ForzaETH & 1 & 25 & 11.54 & 15 & (1) & 20 & (1) & 35 & 1\\
        HiPeRT Modena & 2 & 22 & 13.37 & 14 & (2) & 17 & (4) & 31 & 2 \\
        AUTh Dependables & 1 & 21 & 13.11 & 13 & (3) & 18 & (3) & 31 & 2 \\
        PUT-PPI & 2 & 21 & 13.87 & 13 & (3) & 16 & (5) & 29 & 4 \\
        VAUL & 1 & 20 & 14.84 & 12 & (4) & 14 & (7) & 26 & 5 \\
        Suzlab & 1 & 19 & 14.36 & 11 & (5) & 15 & (6) & 26 & 5 \\
        Scuderia Segfault & 1 & 8 & 12.83 & 5 & (11) & 19 & (2) & 24 & 7 \\
        UT AUTOmata & 1 & 17 & 16.61 & 10 & (6) & 11 & (10) & 21 & 8 \\
        HMCar & 1 & 15 & 16.60 & 9 & (7) & 12 & (9) & 21 & 9 \\
        HUMDA-SZE 2. & 2 & 15 & 18.37 & 9 & (7) & 9 & (12) & 18 & 9 \\
        $\cdots$ & $\cdots$ & $\cdots$ & $\cdots$ & $\cdots$ & $\cdots$ & $\cdots$ & $\cdots$ & $\cdots$ & $\cdots$ \\
    \end{tabular}
    \caption{\emph{Time-Trials} results of the top 10 teams at the \emph{ICRA Grand-Prix 2023}. The table only shows the better of the two heats for each team. Both the fastest achieved lap time and the number of uninterrupted laps are given. In separate columns, the achieved score for both the number of consecutive laps and the lap times are given, and the relative rankings for both categories are given in brackets.}
    \label{tab:results_icra23_time_trials}
\end{table}

During the \emph{Head-to-Head} phase, two teams race in a 1v1 mode where the goal is to finish a given number of laps (typically 10) before the other team. To check for consistent driving of the teams the \emph{Head-to-Head} race is done in a best-of-three mode for each pairing. In the current mode, the teams start on opposite sides of the race track, so overtaking can be avoided. In \Cref{tab:results_icra23_h2h} the results of the \emph{ICRA Grand-Prix 2023} are shown, as one can see \emph{ForzaETH} was able to win each of the brackets in the first two races and never had to rely on a tie-breaker round.

\begin{table}[ht]
    \centering
    \begin{tabular}{c|c|c|c|c|c}
        \textbf{Round} & \multicolumn{2}{c|}{\textbf{Team 1}} & \multicolumn{2}{c|}{\textbf{Team 2}} & \textbf{Score} \\
        \hline
        \hline
        Round of 16 & 1 & ForzaETH & 16 & Technion F1Tenth Team 1 & 2 - 0 \\
        Quarter Final & 1 & ForzaETH & 9 & HMCar & 2 - 0 \\
        Semi Final & 1 & ForzaETH & 4 & PUT-PPI & 2 - 0 \\
        Final & 1 & ForzaETH & 7 & Scuderia Segfault & 2 - 0 \\
    \end{tabular}
    \caption{Results of the \emph{Head-to-Head} phase at the \emph{ICRA Grand-Prix 2023}. For the teams, the Qualifying Rank and the team names are given. One can see that \emph{ForzaETH} managed to win in all \emph{Head-to-Head} pairings within the first two races of the best-of-three setting, never relying on a tie-breaker race.}
    \label{tab:results_icra23_h2h}
\end{table}

% ICRA 2023 Time Trials https://docs.google.com/spreadsheets/d/1ndb5vRNDgYAZtKtV7k-oRrLjCa-48ZjaheCXMYnWL0c/edit#gid=746122213
% ICRA 2023 KO Roster: https://challonge.com/rqhgpovs

% Lausitz 2022 Time Trials: https://drive.google.com/file/d/1NYVlShWQxYOolt1sxOWoJsqYyvJ3iQBC/view
% Lausitz 2022 KO Roster: https://drive.google.com/file/d/1PQpNvzHEU3PFg0TTVsZx8gxpyREwQg29/view

% ICRA 2022 Time Trials: https://docs.google.com/spreadsheets/d/1J0I2ZOywqtDBQ26y8lXAycjIhsyxg_iJi2hk3WOJW1g/edit#gid=0
% ICRA 2022 KO Roster: https://challonge.com/jf88sqbk

\subsection{Qualitative Large Track Run}
\label{app:long_track}
\begin{figure}[ht]
    \centering
    \includegraphics[width=\textwidth]{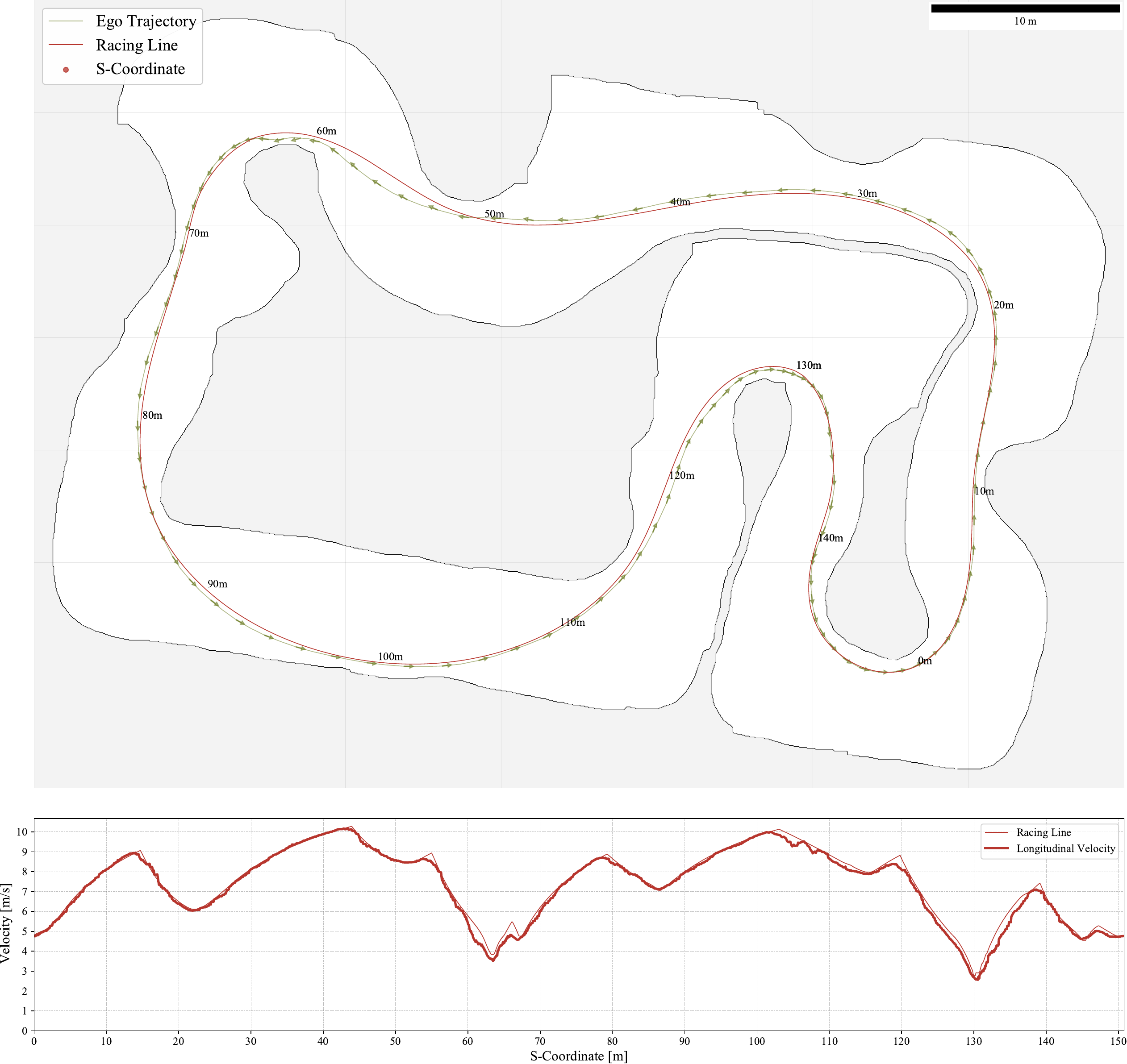}
    \caption{The \emph{ForzaETH Race Stack} deployed on a \SI{150}{\metre} length track, reaching velocities up to \SI{10.21}{\metre\per\second} and completing a single lap within \SI{21.15}{\second}.}
    \label{fig:winti_run}
\end{figure}

\rev{To qualitatively demonstrate the capabilities of the \emph{ForzaETH Race Stack} on larger and complex track layouts, \Cref{fig:winti_run} shows the stack running on a \SI{150}{\metre} long \gls{idsc} autonomous Gokart track reaching a top velocity of \SI{10.21}{\metre\per\second}, reached with a constant scaler of $78 \%$.
The fastest lap time to complete this track with such a scaler was of \SI{21.15}{\second}. The stack was running in \emph{Time-Trials} mode, with default parameters, resulting in a lateral error of \SI{22}{\centi \metre}. The discrepancy from the numbers presented in \ref{chap:ctrl_tracking_results} could be resolved with additional tuning of the control parameters, such as the lookahead distance parameters $m$ and $q$, and additional robustness could be accounted for in the global planning module, by increasing the lateral safety distance.}

\FloatBarrier

\subsection{Frenet Conversion}
\label{app:frenet}
\rev{
This section introduces the three components of the \texttt{frenet\_conversion} library of the \emph{ForzETH Race Stack}, namely the initialization, the \textit{Cartesian-to-Frenet} conversion, and the \textit{Frenet-to-Cartesian} conversion.
It is to be noted, that in both conversions the outcomes are actual approximations, when compared to a continuous \textit{Frenet} representation of the racing line, however, given the used discretization step (\SI{0.1}{\metre}) and minimum radius of curvature (\SI{1}{\metre}) of the racing line, the error becomes negligible for practical purposes.
Indeed, if we consider the overestimated error using a fixed curvature circle with the minimum radius $R=1$, and a maximum delta $\Delta x=0.1$, it is evident that the error along the $s$ coordinate is bounded by the following:
\begin{align*}
    err_{max} &= | \Delta s^* - \Delta x| , \\
    &= \left| R \arctan \left( \frac{\Delta x}{R} \right) - \Delta x \right| , \\
    &= 3.3 \cdot 10^{-4} \;\text{m},
\end{align*}
where $err_{max}$ represent the overestimated $s$ error, and $\Delta s^*$ represents the length of the actual $\Delta s$ segment coming from the approximation.
}
\subsubsection{Initialization}
\rev{
The waypoints obtained from the minimum curvature optimizer are stored in arrays, containing the following information: $x[k],\,y[k]$ for the \textit{Cartesian} coordinates of the waypoints, $\psi$ for the yaw of the waypoint, defined as the angel with the positive global $x$ axis, $s[k],\,d[k]$ for the progression and orthogonal distance from the given minimum curvature line, respectively. It is to be said that the discretization level automatically defines the $s[k]$ array, as it consists of uniformly spaced points, and that the $d[k]$ coordinate is everywhere equal to zero, being it the racing line.
The discrete symbols are further defined in this section as such:
\begin{equation*}
    x[k],\,y[k],\,s[k],\,d[k] \quad \text{for} \quad k \in [0,\,1,\,\hdots N_{wp}],
\end{equation*}
where $N_{wp}$ indicates the total number of waypoints. 
All such vectors are already provided by the optimizer from \cite{Heilmeier2020MinimumCar}.
}

\subsubsection{Frenet-to-Cartesian Conversion}
\rev{
Consider a couple of \textit{Frenet} coordinates $\Bar{s},\,\Bar{d}$.
First, the index of the closest stored waypoint $\Bar{k}$ is defined as follows:
\begin{equation*}
    \Bar{k} \vcentcolon = \arg \min_k |\Bar{s} - s[k]|
\end{equation*}
Then, the requested \textit{Cartesian} point $\Bar{x},\,\Bar{y}$ is obtained as:
\begin{align*}
    \Delta \Bar{s} &= \Bar{s} - s[\Bar{k}], \\
    \Bar{x} &= x[\Bar{k}] + \Delta \Bar{s} \cos (\psi [\Bar{k}]) - \Bar{d} \sin (\psi [\Bar{k}]), \\
    \Bar{y} &= x[\Bar{k}] + \Delta \Bar{s} \sin (\psi [\Bar{k}]) + \Bar{d} \cos (\psi [\Bar{k}]),
\end{align*}
where the second term, in both coordinates' summations, is added to account for linear interpolation between points, and the third term is used to account for the lateral $\Bar{d}$ deviation. 
}

\subsubsection{Cartesian-to-Frenet Conversion}
\rev{
Consider a couple of \textit{Cartesian} coordinates $\Bar{x},\,\Bar{y}$.
First, the index of the closest stored waypoint $\Bar{k}$ is defined as follows:
\begin{equation*}
    \Bar{k} \stackrel{!}{=} \arg \min_k (\Bar{x} - x[k])^2 + (\Bar{y} - y[k])^2
\end{equation*}
Then, the requested Frenet point $\Bar{s},\,\Bar{d}$ is obtained as:
\begin{align*}
    \Delta \Bar{x} &= \Bar{x} - x[\Bar{k}], \\
    \Delta \Bar{y} &= \Bar{y} - y[\Bar{k}], \\
    \Bar{s} &= s[\Bar{k}] + \Delta \Bar{x} \cos(\psi [\Bar{k}]) + \Delta \Bar{y} \sin(\psi [\Bar{k}]) \\
    \Bar{d} &= - \Delta \Bar{x} \sin (\psi [\Bar{k}]) + \Delta \Bar{y} \cos (\psi [\Bar{k}])
\end{align*}
where the $\Delta$ terms in the the resulting \textit{Frenet} coordinates $\Bar{s},\,\Bar{d}$ are obtainable as local projection of the \textit{Cartesian} $\Delta \Bar{x},\,\Delta \Bar{y}$ at the rotated point corresponding to $\Bar{s}$. 
}

\end{document}